\documentclass[10pt,journal,compsoc]{IEEEtran}
\usepackage{cite}
\usepackage{amsmath}
\usepackage{graphicx}
\usepackage{multirow}
\usepackage{multicol}
\usepackage{amssymb}
\usepackage{rotating}
\usepackage{mathtools}
\usepackage[table]{xcolor}
\usepackage{ragged2e}
\usepackage{hhline}
\usepackage{booktabs}
\usepackage{bbding}
\usepackage{wasysym}
\usepackage{threeparttable}
\usepackage{dcolumn}
\usepackage{float}
\usepackage{widetext}
\usepackage{wrapfig}
\usepackage{color}
\usepackage{pifont}

\usepackage[pagebackref=false,breaklinks=true,colorlinks,bookmarks=false]{hyperref}
\makeatletter
\newcommand*\bigcdot{\mathpalette\bigcdot@{0.8}}
\newcommand*\bigcdot@[2]{\mathbin{\vcenter{\hbox{\scalebox{#2}{$\m@th#1\bullet$}}}}}
\makeatother

\begin{document}
\title{Weakly Supervised Visual-Auditory Fixation Prediction with Multigranularity Perception}
\author{Guotao~Wang,~
        Chenglizhao~Chen*,~
        Deng-Ping~Fan,~
        Aimin~Hao,~
        and~Hong~Qin\\
        \vspace{-0.1cm}

\IEEEcompsocitemizethanks{
\IEEEcompsocthanksitem
G. Wang is with the State Key Laboratory of Virtual Reality Technology and Systems, Beihang University, China.
(E-mail: qduwgt@163.com)
\IEEEcompsocthanksitem
C. Chen is with the College of Computer Science and Technology, China University of Petroleum, China.
(e-mail: cclz123@163.com).
\IEEEcompsocthanksitem
D.-P. Fan is with ETH Zurich, Zurich, Switzerland
(Email: dengpingfan@mail.nankai.edu.cn)
\IEEEcompsocthanksitem
A. Hao is with the State Key Laboratory of Virtual Reality Technology and Systems, Beihang University,
Research Unit of Virtual Human and Virtual Surgery, Chinese Academy of Medical Sciences, and
with Pengcheng Laboratory, China.
(Email: ham@buaa.edu.cn)
\IEEEcompsocthanksitem
H. Qin is with the Computer Science Department, Stony Brook University, USA.
(e-mail: qin@cs.stonybrook.edu).
\IEEEcompsocthanksitem
A preliminary version of this work appeared at CVPR 2021 ~\cite{Wang_2021_CVPR}.
\IEEEcompsocthanksitem
Corresponding author: Chenglizhao Chen.}}

\markboth{IEEE TRANSACTIONS ON PATTERN ANALYSIS AND MACHINE INTELLIGENCE}%
{Shell \MakeLowercase{\textit{\textit{et al.}}}: Bare Advanced Demo of IEEEtran.cls for IEEE Computer Society Journals}

\IEEEtitleabstractindextext{%
\begin{abstract}
\justifying
Thanks to the rapid advances in deep learning techniques and the wide availability of large-scale training sets,
the performance of video saliency detection models has been improving steadily and significantly.
However, deep learning-based visual-audio fixation prediction is still in its infancy.
At present, only a few visual-audio sequences have been furnished, with real fixations being recorded in real visual-audio environments.
Hence, it would be neither efficient nor necessary to recollect real fixations under the same visual-audio circumstances.
To address this problem,
this paper promotes a novel approach in a weakly supervised manner to alleviate the demand of large-scale training sets for visual-audio model training.
By using only the video category tags,
we propose the \emph{\textbf{s}elective} \textbf{c}lass \textbf{a}ctivation \textbf{m}apping (SCAM) and its upgrade (SCAM+).
In the spatial-temporal-audio circumstance, the former follows a coarse-to-fine strategy to select the most discriminative regions, and these regions are usually capable of exhibiting high consistency with the real human-eye fixations.
The latter equips the SCAM with an additional multi-granularity perception mechanism, making the whole process more consistent with that of the real human visual system.
Moreover, we distill knowledge from these regions to obtain complete new \textbf{s}patial-\textbf{t}emporal-\textbf{a}udio (STA) \textbf{f}ixation \textbf{p}rediction (FP) networks, enabling broad applications in cases where video tags are not available.
Without resorting to any real human-eye fixation,
the performances of these STA FP networks are comparable to those of fully supervised networks. The code and results are publicly available at \url{https://github.com/guotaowang/STANet}.
\end{abstract}

}

\maketitle
\vspace{-0.4cm}
\IEEEdisplaynontitleabstractindextext
\IEEEpeerreviewmaketitle
\ifCLASSOPTIONcompsoc

\vspace{-0.4cm}
\section{Introduction and Motivation}
\label{sec:introduction}
\IEEEPARstart{I}{n} the deep learning era,
we have witnessed a growing development in video saliency detection techniques
~\cite{wang2019revisiting,linardos2019simple,lai2019video,fang2018deep3dsaliency,newvideosal20aaai,xu2018personalized},
where the primary task is to locate the most distinctive regions in a series of video sequences.
At present, this research field consists of two parallel research directions, \textit{i.e.},
the \textbf{v}ideo \textbf{s}alient \textbf{o}bject \textbf{d}etection (VSOD) and the \textbf{v}ideo \textbf{f}ixation \textbf{p}rediction (VFP).
In practice, the former~\cite{gu2020pyramid,wang2020learning,ren2020tenet,MGA,SSAV,CC2017TIP,CC2020TIPWGT,cong2019video,li2019supervae}
aims to \emph{\textbf{segment}} the most salient objects with clear object boundaries (\emph{e.g.}, Fig.~\ref{fig:Brief}-B).
The latter~\cite{min2019tased,wu2020salsac,droste2020unified,bak2017spatio,sun2018sg,gorji2018going}, the main topic of this paper,
attempts to predict human-eye-like fixations --- scattered coordinates spreading over the entire scene without any clear boundaries (\emph{e.g.}, Fig.~\ref{fig:Brief}-A).
In fact, this topic has been investigated extensively in recent decades.

Different from the previous works~\cite{jiang2015salicon,pan2016shallow,lai2019video,wang2017deep},
the interests in this paper lie in exploiting deep learning techniques to predict fixations under visual and audio circumstances,
also known as visual-audio fixation prediction,
and this topic is still in its early exploration stage.

Human visual-audio fixation prediction is applied to a wide scope of \textbf{\emph{visual-audio}} applications in various areas.
The representative applications include Kinematics~\cite{shewchenko2005heading}, Criminal Psychology~\cite{shigeoka2013tumor}, Airplane Pilots' Skill Training~\cite{mandal2016towards}, Lie Detection~\cite{podlesny1977physiological}, Operative Risk Assessment~\cite{kok2017before}, 360 Video Surveillance~\cite{zhu2018prediction, zhu2021viewing}.
For example, when people lie, they produce a series of physiological reactions, such as pupil dilation, increased blinking times, voice trembling, \textit{etc.}., and such ``external appearances'' have already been used in lie detection. However, in some cases, we shall observe the subject from the inside~\cite{podlesny1977physiological}. By analyzing the subtle changes in the subject's eye fixations, we can improve the lie detection accuracy. In practice, \textbf{\emph{the subject is usually exposed in a visual-audio environment}}, where the conventional visual-based fixation methods cannot handle well.

At present,
almost all \textbf{s}tate-\textbf{o}f-\textbf{t}he-\textbf{a}rt (SOTA) visual-audio fixation prediction approaches~\cite{tsiami2020stavis,tavakoli2019dave} are developed with
the help of deep learning techniques,
using the vanilla encoder-decoder structure,
facilitated with various attention mechanisms,
and trained in a fully  supervised manner.
Despite achieving progress,
these fully supervised approaches are plagued by one critical limitation.

\begin{figure}[!t]
\centering
\includegraphics[width=0.9\linewidth]{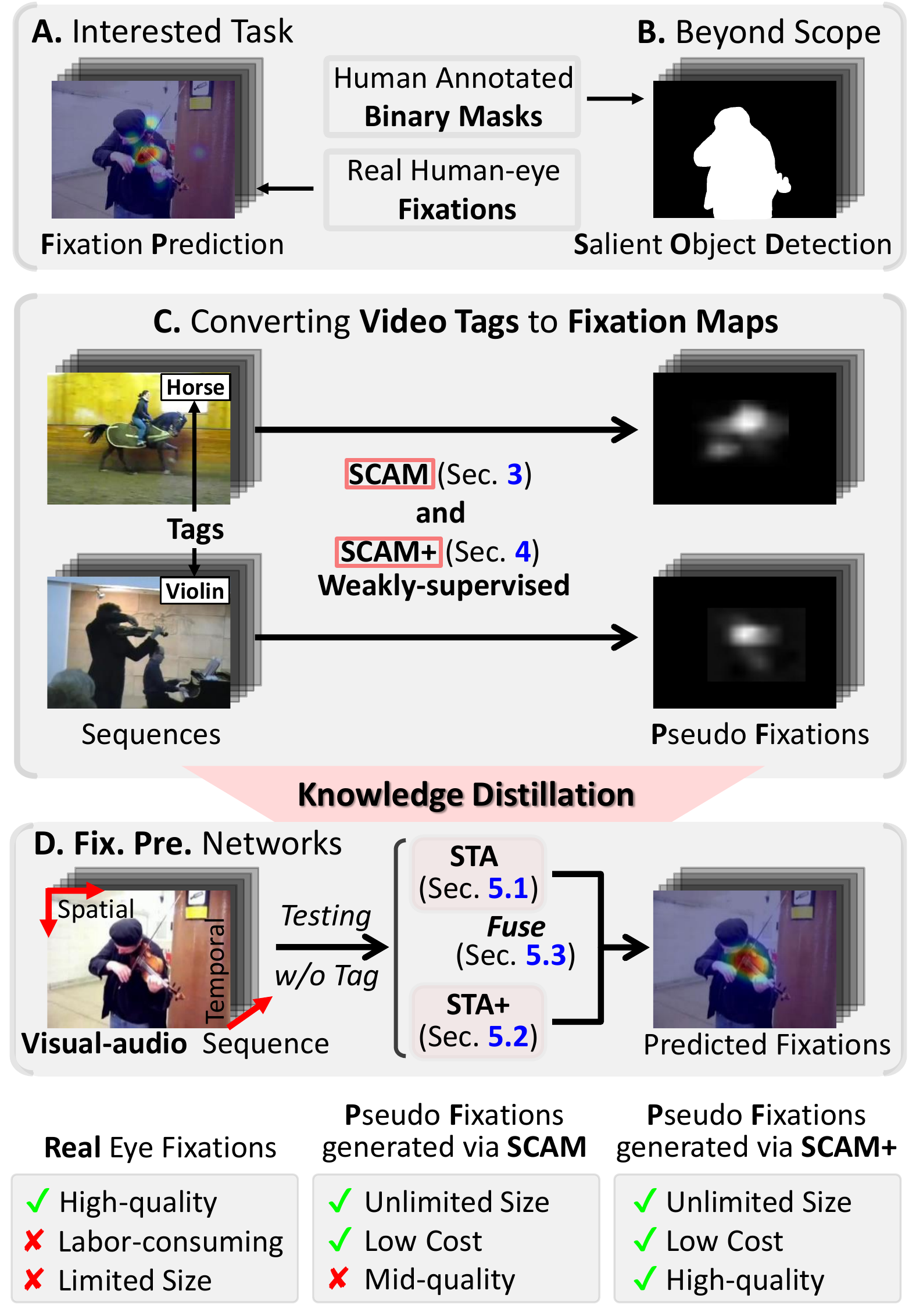}
\vspace{-0.4cm}
\caption{This paper mainly focuses on devising
a \textbf{\emph{weakly supervised}} approach for the
\textbf{s}patial-\textbf{t}emporal-\textbf{a}udio (STA) fixation prediction task,
where the key innovation is that, as one of the first attempts, we
automatically convert semantic category tags to pseudofixations via
the newly proposed \emph{\textbf{s}elective} \textbf{c}lass
\textbf{a}ctivation \textbf{m}apping (SCAM, Sec.~\ref{sec:SCAM})
and the upgraded version SCAM+ (Sec.~\ref{sec:SCAM+}) that has been additionally equipped with the multi-granularity perception ability.
The obtained pseudofixations can be used as the learning objective to guide \textbf{\emph{knowledge distillation}}
to teach two individual fixation prediction networks (\textit{i.e.}, STA and STA+, Sec.~\ref{sec:FPN}),
which jointly enable \textbf{\emph{generic}} video fixation prediction without requiring any video tags.
On the right side, we use {\color{green}\ding{51}} and {\color{red}\XSolidBrush} to denote advantages and disadvantages of each label, respectively.
\vspace{-0.4cm}}
\label{fig:Brief}
\end{figure}

It is well known that
a deep model's performance is heavily dependent on the adopted training set, and currently, large-scale visual-related training sets are already accessible in our research community.
However, the visual-audio circumstance-related training data are rather short since collecting real human-eye fixations in such multimodality circumstances is a time-consuming and laborious process.
To the best of our knowledge, only a few visual-audio sequences equipped with real fixation data are available for the visual-audio fixation prediction task, where only a small part of them are recommended for network training,
making the data shortage dilemma even worse.
As a result,
according to the extensive quantitative evaluation that we have performed,
most of the existing deep learning-based visual-audio fixaiton prediction models~\cite{tsiami2020stavis,tavakoli2019dave},
although reluctantly admitted, might be overfitted in essence.

To solve this problem,
we seek to realize visual-audio fixation prediction using a weakly supervised strategy.
Instead of using the labor-intensive framewise visual-audio \textbf{g}round \textbf{t}ruth\textbf{s} (GTs),
we devise a novel scheme to produce GT-like visual-audio pseudofixations by using only video category tags as supervision.
Actually, a plethora of visual-audio sequences with well-labeled semantic category tags already exist (\emph{e.g.},
the AVE set~\cite{tian2018audio}), where most of them were originally collected for the visual-audio classification task.
Note that, from a cost perspective, manually assigning semantic tags to videos is clearly more favorable than collecting real human-eye fixations.
Thus, the key is how to convert video tags to the real-fixation-like data.

Our approach is inspired by the \textbf{c}lass \textbf{a}ctivation \textbf{m}apping (CAM,~\cite{Zhou_2016_CVPR})
that has been used in various object localization and object detection tasks~\cite{zeng2019multi,wang2017learning,saito2019strong,bargal2018excitation,chattopadhay2018grad,Li_2021_WACV}.
Our rationale relies on the fact that image regions with the strongest discriminative power regarding the classification task should be the most salient ones,
where these regions usually tend to have larger classification confidences than others.

As seen in Fig.~\ref{fig:Brief}-C, considering that we aim at fixation prediction in visual-audio circumstances, we present two practical innovations in formulating high-quality fixation maps: 1) \emph{\textbf{s}elective} \textbf{c}lass \textbf{a}ctivation \textbf{m}apping (SCAM) and 2) SCAM+ --- an upgraded version of the SCAM.
The key rationale of SCAM relies on the `\emph{\textbf{data source}}' aspect, which performs a coarse-to-fine strategy to reveal the most discriminative regions from multiple sources (\textit{i.e.}, spatial, temporal, audio, and their combinations), where these regions exhibit high consistency with the real human-eye fixations.
This coarse-to-fine methodology ensures that the aforementioned less discriminative scattered regions are filtered completely,
and the selection operation between different sources helps reveal the most discriminative regions,
enabling the pseudofixations to be closer to real fixations.

Furthermore, since the human visual system is clearly not independent of our brain, real human-eye fixations are usually influenced by multiple physiological activities, \emph{e.g.}, short-term memory, long-term memory, associative memory, and semantic reasoning, making our attention mechanism a `\emph{\textbf{global}}' process in essence.
However, the proposed SCAM follows a typical `\emph{\textbf{local}}' mechanism that only considers 3 frames and a 1 second audio signal at most each time.
Thus, we devise the upgraded version of SCAM, named SCAM+, which equips SCAM with an additional multi-granularity perception mechanism to further boost the consistency degree between the derived pseudofixations and real fixations.
By using both SCAM and SCAM+, we can automatically convert video tags to high-quality pseudofixation maps, and thus, theoretically, we can easily expand the existing video fixation training sets to unlimited sizes.
To better highlight the novelty, we pictorially clarify the rationale of our approach in Fig.~\ref{fig:idea}.

\begin{figure}[!h]
  \centering
  \vspace{-0.2cm}
  \includegraphics[width=0.8\linewidth]{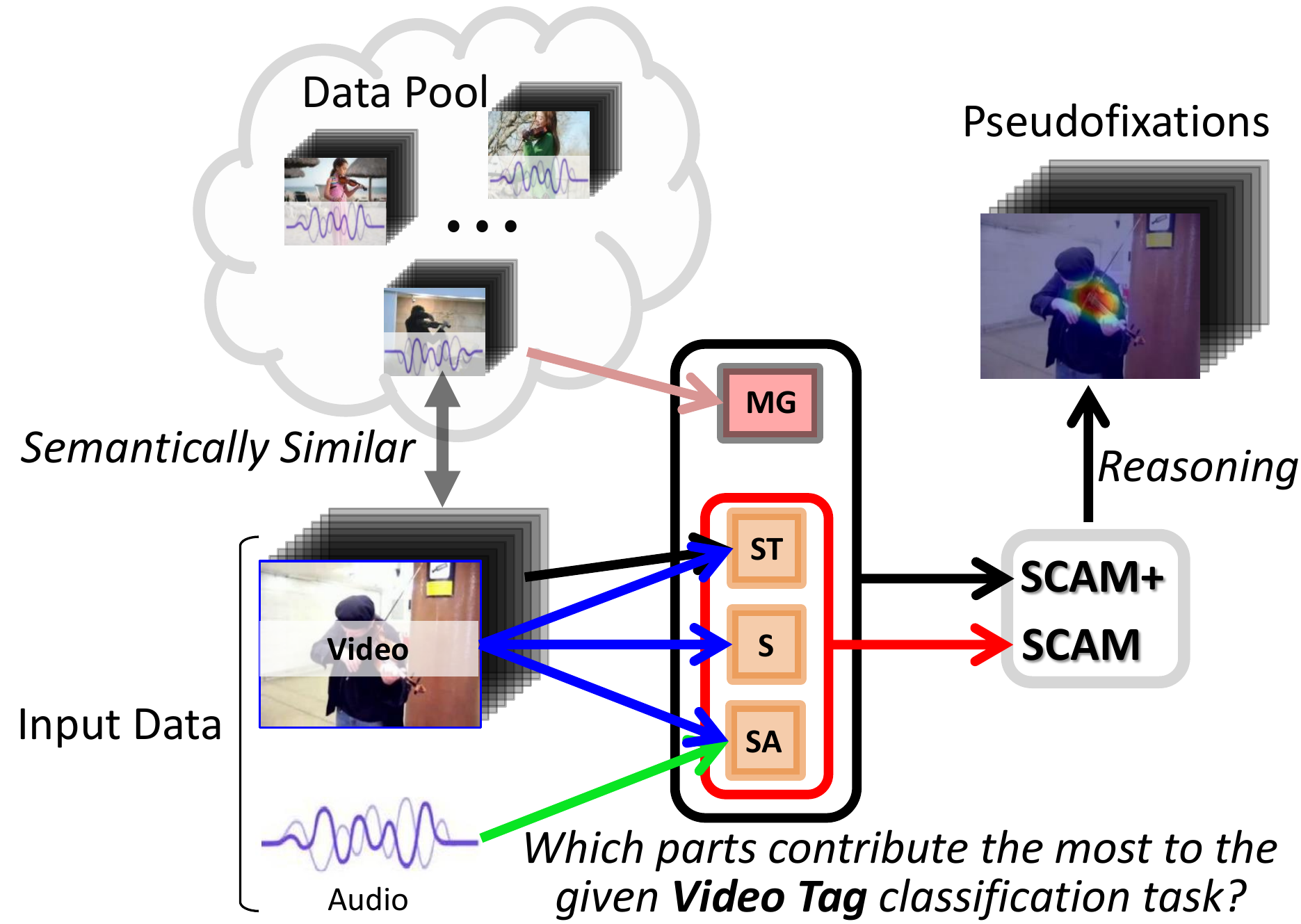}
  \vspace{-0.3cm}
  \caption{As the first attempt, our approach targets mimicking the human attention mechanism in a multimodality environment. The key technical highlight is the proposed feasible way to localize the most discriminative regions over different modalities (S, SA, and ST) with multi-granularity (MG) perception. S: spatial, SA: spatial + audio, ST: spatial + temporal.
  \vspace{-0.1cm}}
  \label{fig:idea}
\end{figure}

In addition, to ensure a broad application, we shall use the pseudofixations derived by both SCAM and SCAM+ to train end-to-end fixation prediction networks, and this process could be regarded as a variant of knowledge distillation, where the end-to-end fixation prediction networks are clearly the students.
Specifically, we devised 2 \textbf{s}patial-\textbf{t}emporal-\textbf{a}udio fixation prediction networks, which correspondingly correlate to the SCAM- and SCAM+-based pseudo GTs.
During the testing phase, the outputs of these 2 fixation prediction networks will be complementarily fused as high-quality fixation maps.
As seen in Fig.~\ref{fig:Brief}-D, the whole testing process does not require assigning semantic video tags to the given input testing sequence, and this attribute ensures a broad application of the proposed approach in real works.
To facilitate a better reading, we summarize the key contributions as follows:

1) From the sourcewise perspective, we have presented a novel way to convert video tags to fixation maps, \textit{i.e.}, SCAM, which, in a course-to-fine manner, localizes the discriminative image regions sourcewise, and these localized regions could be very consistent with real human-eye fixations.

2) Inspired by the physiological mechanism of human attention, we have devised an upgraded version of SCAM, named SCAM+, which equips the SCAM with multi-granularity perception ability. The proposed SCAM+ is able to compress those less discriminative image regions and focus on the most discriminative regions, further boosting the quality of the derived fixation maps.

3) We provide one of the first attempts to predict deep learning-based visual-audio fixation in a weakly supervised manner, which is expected to contribute to visual-audio information integration and relevant applications in computer vision. Our work could bridge the relationship between the machine-oriented classification and the real human visual-auditory system, paving a new way to mimic human attention in a multimodality environment.

This paper builds upon our conference version~\cite{Wang_2021_CVPR}, which has been extended in two distinct aspects.
First, on the basis of multisource and multiscale perspectives which have been adopted by the CVPR version,
we have provided a deep insight into the relationship between multi-granularity perception and real human attention behaved in visual-auditory environment.
Second, without using any complex networks, we have provided an elegant framework to complementary integrate multisource, multiscale, and multigranular information to formulate pseudofixations which are very consistent with the real ones.
Apart from achieving significant performance gain, this work also provides a comprehensive solution for mimicking multimodality attention.

\vspace{-0.2cm}
\section{Related Works}
\label{sec:Realated}
%

\subsection{Weakly-supervised Approaches}
Based on the pregiven image-level labels~\cite{wang2017learning,Shipami,Sanginetopami, lu2016learning},
points~\cite{qian2019weakly}, scribbles~\cite{zhang2020weakly} and bounding boxes~\cite{dai2015boxsup},
weakly supervised methods can usually outperform unsupervised approaches.
Here, we simply review both weakly supervised object detection and segmentation methods.

Generally, weakly supervised object detection methods that only use image-level tags as supervision could achieve plausible performance; however, they have one critical limitation, \textit{i.e.}, their detections tend to cover the most discriminative parts rather than the entire object. To facilitate better understanding, we shall briefly review several of the most representative methods.
Aiming to evaluate the object proposal's contribution, Li \textit{et al.}~\cite{li2019progressive} proposed occluding each proposal via a gray mask and then feeding the occluded image to an image-level classifier. By observing the difference in classification confidence before and after the occlusion operation, only those proposals with large differences will be retained, since these proposals have contributed most to the classification task.
Wan \textit{et al.}~\cite{Wanpami} focused on the object detection task, which attempts to convert image-level object labels to high-quality object proposals. Based on the initial proposals generated by selective search, this work proposed the min-entropy function to weakly supervise the learning process of an additional classifier, where those less trustworthy object category predictions of the selective search could exhibit large conflict with those of the newly trained classifier, only those high-quality object proposals shall be retained.
Following the knowledge distillation framework, Zeng \textit{et al.}~\cite{zeng2019wsod2} linearly combined image-level classification confidence with superpixel-level straddling --- a metric to measure the confidence of a superpixel belonging to an object --- to filter some inconsistent initial object proposals. Then, these retained proposals, as the teacher, were used to train a student object detector. This process was repeated multiple times to obtain high-quality object proposals.
In view of the weakly supervised segmentation approaches, the key rationale is to resort the image-level classification task to generate pseudomasks, and these masks will be used to facilitate the segmentation task.
Jiang \textit{et al.}~\cite{jiang2021online} proposed an online attention accumulation strategy
that utilizes the attention maps at different epochs of training phases to obtain more integral object regions.
Then, the accumulated attention map, serving as pseudoground truths, will be used to train an individual segmentation model.

\begin{figure}[!b]
  \centering
  \vspace{-0.7cm}
  \includegraphics[width=1\linewidth]{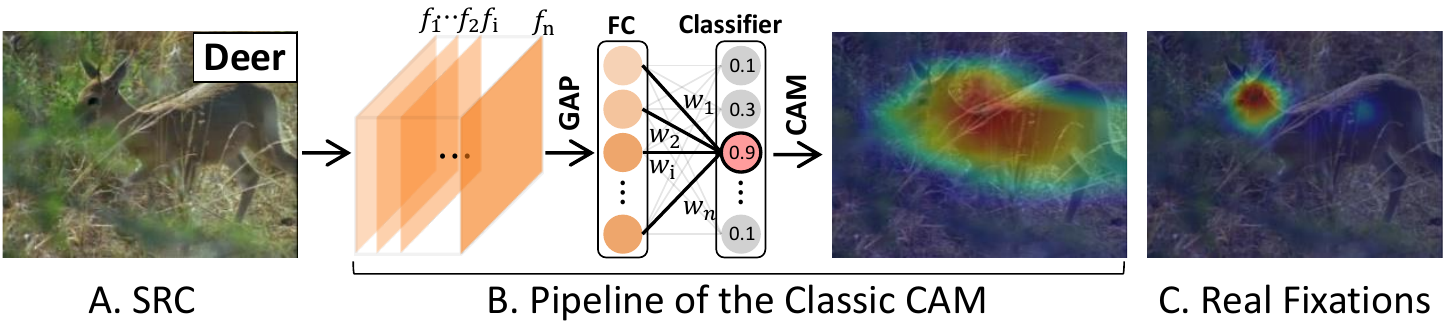}
  \vspace{-0.7cm}
  \caption{Illustration of the \textbf{c}lass \textbf{a}ctivation \textbf{m}apping (CAM) details.
  FC: fully connected layer;
  GAP: global average pooling;
  numbers in the classifier represent the classification confidences.
  \vspace{-0.2cm}}
\label{fig:ProblemState}
\end{figure}

Targeting the segmentation task, Choe \textit{et al.}~\cite{choe2020attention} proposed a recursive erasing strategy to achieve complete segmentation results. The key rationale is to partially erase the current most discriminative regions and then feed the erased image back to the image-level classifier to reveal the most discriminative regions. By repeating this process multiple times, all object parts, which usually exhibit different discriminative degrees, could eventually be revealed.
Similarly, methodologies can be found in~\cite{li2019guided}, while the major differences could be the usage of additional supervised training data, and thus, the weakly supervised training process could benefit from these additional supervised ones.
Zhang \textit{et al.}~\cite{2020Reliability} proposed a bistream structure, where one stream aims at image-level classification and another targets the pixelwise regression task, and the pixelwise regression objectives were generated by applying CRF over pseudomasks converted from image-level labels in advance (more details regarding pseudomask generation can be found in~\cite{Salehpami}).
The key rationale is to perform interactive cross-supervision between these two streams, where the primary objective is to ensure that pixels with similar contributions towards the classification task have similar regression values.
To correlate object binary masks with semantic labels, Lu \textit{et al.}~\cite{lu2016learning} proposed a joint L1 optimization to ensure that masks sharing similar appearances should belong to an identical category.
Lu \textit{et al.}~\cite{lu2016learning} address the problem of learning with both weak and noisy labels as a label noise reduction problem,
segmenting each image into a set of superpixels; then
a novel L1-optimization-based sparse learning model is formulated.
To solve the L1-optimisation problem,
they developed an efficient learning algorithm by introducing an intermediate labeling variable.

\vspace{-0.3cm}
\subsection{Class Activation Mapping}
\label{sec:CAM}
In the typical video classification field,
each training sequence is usually assigned a semantic tag
which associates this sequence with a specific video category.
In general, these semantic tags are assigned by performing majority voting between multiple persons,
aiming to represent the most meaningful objects or events in the given video sequence.
Similar to the process of tag assignment,
real human visual fixations tend to focus on the most meaningful and representative regions when watching a video sequence.
Thus, formulating pseudofixations from video category tags is theoretically feasible.

The fundamental idea of \textbf{c}lass \textbf{a}ctivation \textbf{m}apping (CAM)
is to use the weighted summation of feature maps
in the last convolutional layer
to coarsely locate the most representative image regions
regarding the current classification task.
In practice, as can be seen in Fig.~\ref{fig:ProblemState}-B,
those weights ($w_i$) correlated with the relatively highest classification confidences in the last fully connected layer
are selected to weigh the feature maps ($f_i$).
Thus, the CAM, a 2-dimensional matrix,
can be obtained by:
\begin{equation}
{\rm CAM} = \mathcal{Z}\Big(\sum_{i}^d w_i\times f_i\Big),
\label{eq:RCAM}
\end{equation}
where $d$ represents the feature channel number and $\mathcal{Z}(\cdot)$ is the \emph{min-max} normalization function.

From a qualitative perspective,
the CAM,
which has been visualized in the right part of Fig.~\ref{fig:ProblemState}-B,
usually shows a large feature response to frame regions (\textit{i.e.},
the `\emph{Deer}') that has contributed most regarding the classification task,
and these regions usually correlate to the most salient regions.

More commonly, the CAM could be quite different from the real human-eye fixations, \textit{i.e.}, Fig.~\ref{fig:ProblemState}-B vs. Fig.~\ref{fig:ProblemState}-C.
Actually, when performing the video classification task, the image regions with the most substantial contribution to the given category are capable of highlighting the salient object (\textit{i.e.}, the deer).
Following this rationale,
several previous works~\cite{CAMYangyi,gao2019c,wan2019c,arun2019dissimilarity,chen2020slv}
resorted to the CAM for the salient object localization task.
However, the CAM results derived by these methods are different from real human-eye fixations,
and the main reasons comprise the following 3 aspects.

First, since both local and nonlocal deep features contribute to the classification task,
the CAMs tend to be large scattered regions.
For example, as shown in Fig.~\ref{fig:ProblemState},
the main body of the deer can help the classifier to separate this image from other nonanimal cases,
while only the `\emph{deer head}' can tell the classifier that the animal in this scene is a `\emph{deer}'.
Instead of gazing at the `\emph{main body}',
our human visual system tends to focus more attention on the most discriminative image regions
(\emph{e.g.}, the `\emph{deer head}', Fig.~\ref{fig:ProblemState}-C).

Second, most of the existing works~\cite{zeng2019multi,wang2017learning,ye2019cap2det,yang2019activity,inoue2018cross}
have only considered the spatial information when computing CAM.
However, real human-eye fixations are affected by multiple sources,
including spatial, temporal, and audio sources.
In fact, this multisource nature has long been omitted by our research community
because, compared with the spatial information --- a stable source,
the other 2 sources (temporal and audio) are still considered to be rather unstable sources thus far,
and this unstable attribute makes them difficult to use for computing CAM.
However, in many practical scenarios,
it is exactly these 2 sources that could be the most beneficial to the classification task.

\begin{figure}[!t]
  \centering
  \includegraphics[width=1\linewidth]{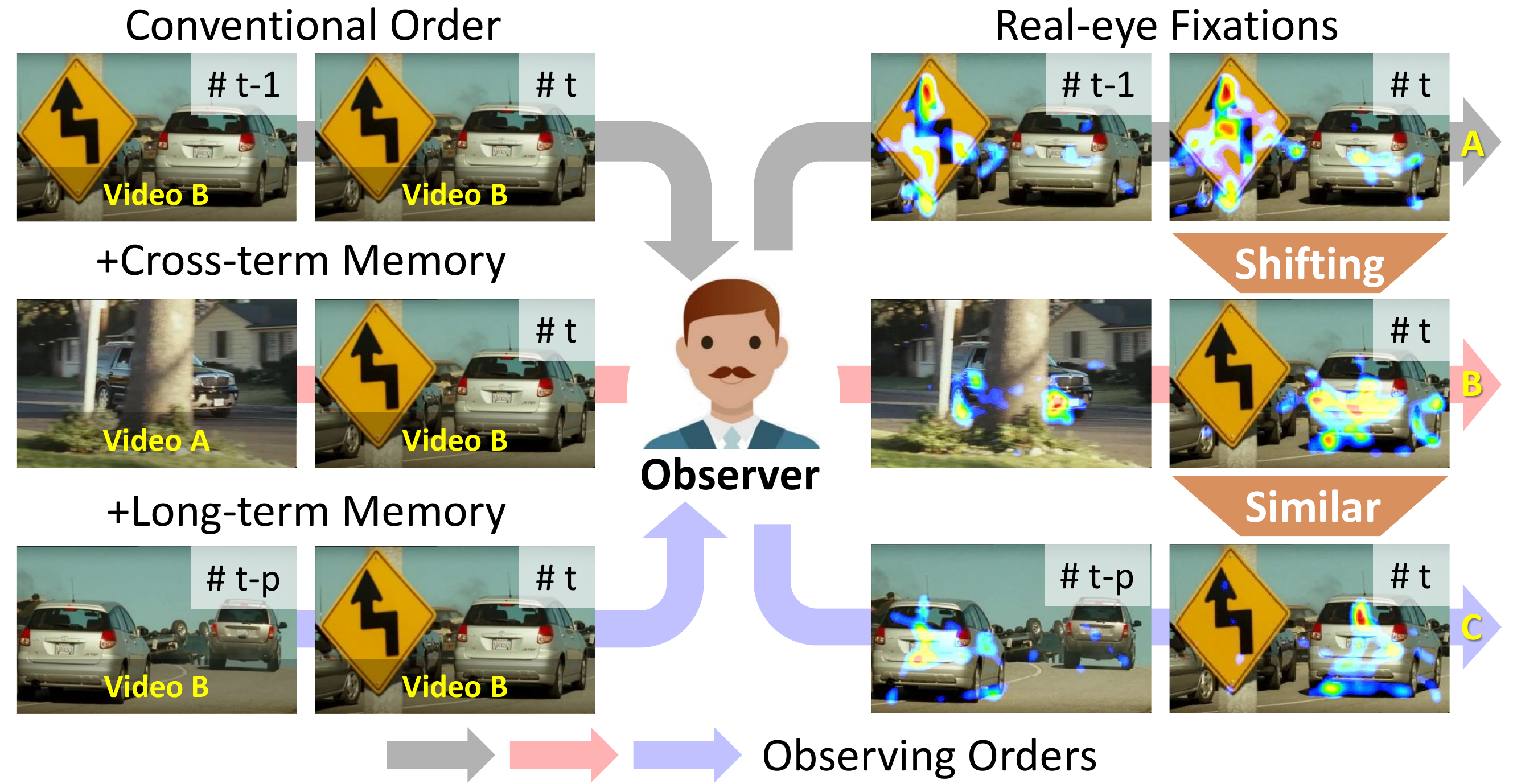}
  \vspace{-0.7cm}
  \caption{Illustration of the importance of multi-granularity perception ($t-p\ll t$). We have conducted additional experiment to support our claim, which can be found in the submitted `\textbf{Supplementary Material}'.
  \vspace{-0.4cm}}
  \label{fig:MGRationale}
\end{figure}

Third, real human-eye fixation is not completely derived via the physiological reflex, but it is also subtly influenced by our brain, which includes multiple physiological activities, implying that all short-term memory, long-term memory, associative memory, and semantic reasoning should be considered when performing CAM.
For example, as seen in Fig.~\ref{fig:MGRationale}-A, the `\emph{road sign}' attracts massive eye fixations if only the SRC image has been shown. However, the `\emph{silver car}' could replace the `\emph{road sign}' as the most salient region if either long-term information Fig.~\ref{fig:MGRationale}-B or associative information Fig.~\ref{fig:MGRationale}-C has been given in advance, where the associative information can be obtained by aligning a video frame from other video sequences sharing an identical video tag with the target sequence.
Therefore, multi-granularity perception is called for when resorting to the CAM to produce high-quality fixations.


\vspace{-0.2cm}
\section{Selective Class Activation Mapping}
\label{sec:SCAM}
Compared with the single image case,
the problem domain of our visual-audio case is much more complicated,
where we need to consider multiple sources simultaneously,
including spatial, temporal, and audio sources.

As mentioned above,
the conventional CAMs derived from using spatial information solely tend to be large scattered regions,
which are quite different from the real fixations;
even worse,
it has completely omitted the complementary status between different sources.
Actually, in spatial-temporal-audio circumstances,
the feature maps embedded by the encoder tend to be multiscale, multilevel, and multisource,
where all of them will jointly contribute to the classification task.
However, there would be massive false alarms and redundant responses if we natively combine all these sources, without considering the repulsiveness of complementary information.

\begin{figure}[!t]
\centering
\includegraphics[width=1\linewidth]{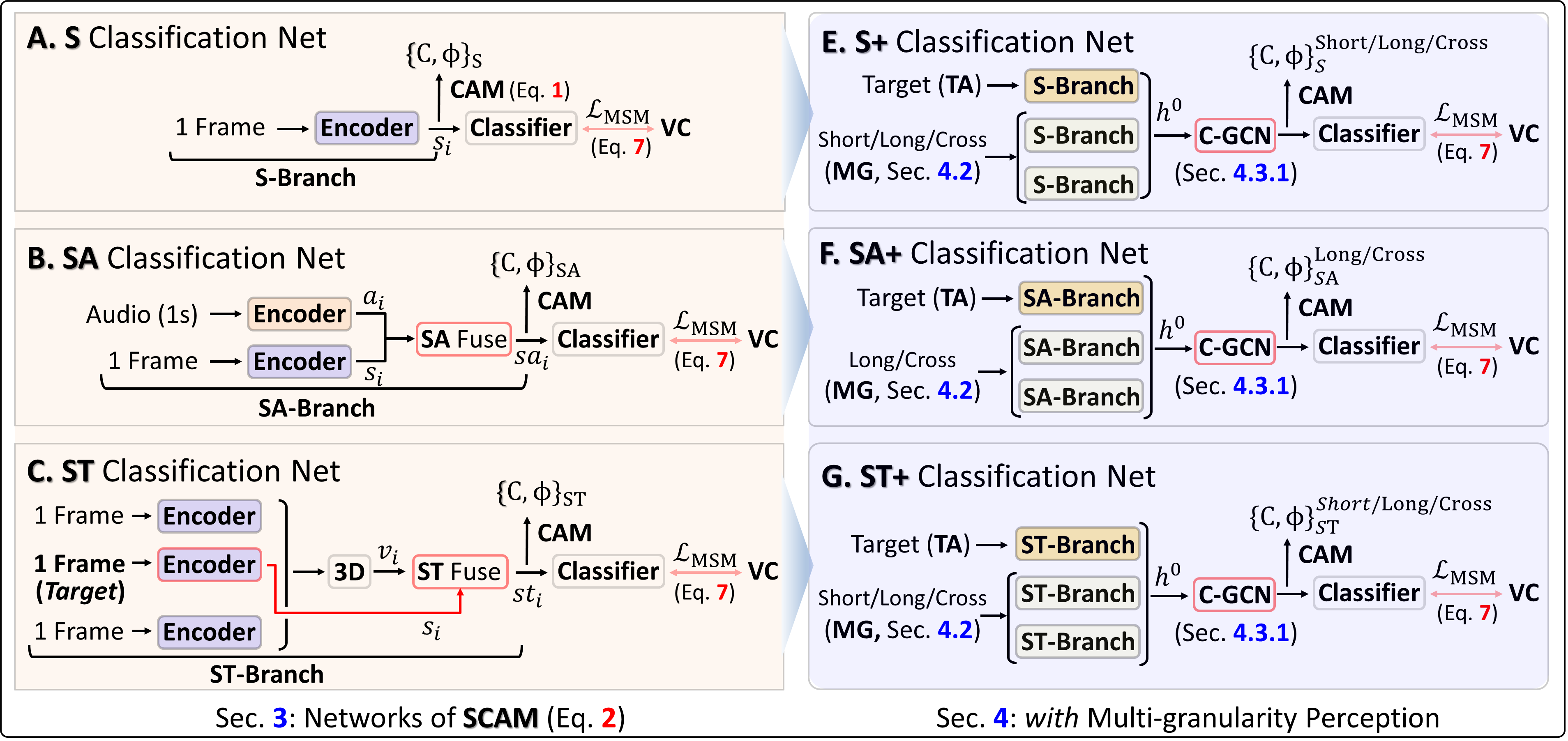}
\vspace{-0.6cm}
\caption{The detailed network architectures used in this paper.
        Subfigures A-C are the classification nets used in the proposed SCAM;
        subfigures E-G are the upgraded classification nets, which couple the raw versions together to serve the proposed SCAM+.
        3D: 3D convolution; VC: video category;
        both SCAM and SCAM+ can convert video tags to pseudofixations (pGTs), and SCAM+ can consistently outperform SCAM. 
        Details of `SA/ST Fuse' can be found in Fig.~\ref{fig:FusionM}.
        C-GCN is the proposed $m$-step reasoning, whose details can be found in Fig.~\ref{fig:CGCNpipe} and Sec.~\ref{sub:NMP}.}
        \vspace{-0.4cm}
\label{fig:Networks}
\end{figure}

To solve this problem,
we propose to decouple the spatial-temporal-audio circumstance into 3 independent sources, \textit{i.e.},
spatial (S), temporal (T), and audio (A), and these sources will be individually fed into 3 classification nets (\textit{i.e.}, S, ST, and SA, see subfigures A-C in Fig.~\ref{fig:Networks}, and the rationale towards this combination will be provided in the next subsection).
By performing CAM over these classification nets, we can eventually obtain the sourcewise CAMs.
Clearly, this divide-and-conquer strategy can alleviate the redundancy problem effectively; however, to mimic the real human attention mechanism in spatiotemporal-audio circumstances, we shall `selectively fuse' them to localize the most discriminative regions.
Thus, the key issue here is how to simultaneously achieve a sourcewise complementary status and avoid accumulating redundant feature responses.
Therefore, we present the \emph{\textbf{s}elective} \textbf{c}lass \textbf{a}ctivation \textbf{m}apping (SCAM),
whose technical details can be formulated by Eq.~\ref{eq:FU}.
\begin{equation}
\label{eq:FU}
\begin{aligned}
{\rm SCAM}& = \mathcal{Z}\Big(\frac{\big\lVert{\rm UC} \odot {\rm UR}\big\rVert_1 + \lambda}{\big\lVert{\rm UC}\big\rVert_1 + \lambda}\Big),\\[-0.5ex]
&{\rm UR}: \Big[{\rm \Phi}_{\rm S}\{i\}, {\rm \Phi}_{\rm ST}\{i\}, {\rm \Phi}_{\rm SA}\{i\}\Big],\\[-0.5ex]
&{\rm UC}: \bigg[\oint\big({\rm C}_{\rm S}\{i\}\big), \oint\big({\rm C}_{\rm ST}\{i\}\big), \oint\big({\rm C}_{\rm SA}\{i\}\big)\bigg],
\end{aligned}
\end{equation}
where $\odot$ is the elementwise multiplicative operation; $\lVert\cdot\rVert_1$ denotes the L1-norm; $\lambda$ is a small constant to avoid any division by zero;
$\rm\Phi_{S}$, $\rm\Phi_{ST}$, and $\rm\Phi_{SA}$ represent the CAM result (Eq.~\ref{eq:RCAM}) derived from S, ST, and SA classification nets, respectively; $\mathcal{Z}(\cdot)$ is the \emph{min-max} normalization operation.
In addition, suppose the pregiven category tag of the S classification net is the $i$-th category in total $c$ (28) classes,
then we use ${\rm C}_{\rm S}{\{i\}}$ to represent this confidence, where ${\rm C}_{\rm S}\in(0,1)^{1\times c}$.
$\oint(\cdot)$ is a soft filter (Eq.~\ref{eq:BI}) aiming to compress those features of low classification confidences to be considered when computing the SCAM.
\begin{equation}
\label{eq:BI}
\begin{aligned}
\oint\big({\rm C}_{\rm S}\{i\}\big)=\left\{\begin{array}{ll}{\rm C}_{\rm S}\{i\}\ \ &{\large{if}}\ \ \ {\rm C}_{\rm S}\{i\}>{\rm C}_{\rm S}\{u\}|_{i\ne u, 1\le u\le c}\\
\ 0\ \  &otherwise\\
\end{array}\right..
\end{aligned}
\end{equation}

In the following subsections, we provide the underlying rationales for the SCAM. Furthermore,
to make the SCAM more consistent with the real fixations, we will introduce an effective stagewise manner, \textit{i.e.}, the stagewise SCAM.

\begin{figure}[!t]
  \centering
  \includegraphics[width=1\linewidth]{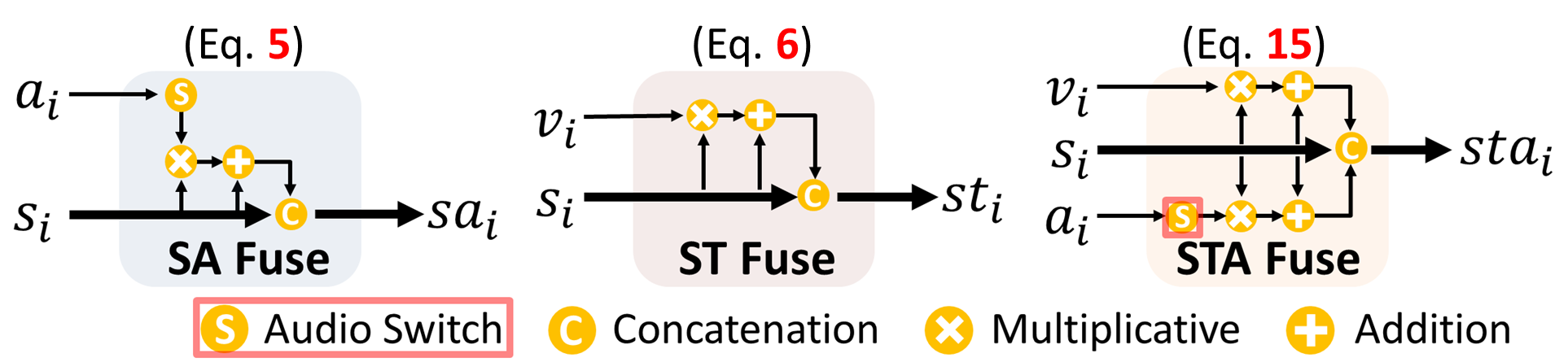}
  \vspace{-0.6cm}
          \caption{Architectures of fusion modules adopted in Fig.~\ref{fig:Networks} and~\ref{fig:VFPNetworks}.}
          \vspace{-0.3cm}
  \label{fig:FusionM}
\end{figure}

\subsection{SCAM Rationales}
Generally,
either spatial, temporal, or audio sources could influence our visual attention;
however, compared with the last two,
the spatial source is usually more important and stable in practice.
For example, when a video object remains static for a long period of time, its spatial appearance could be completely unchanged, while its temporal information becomes completely absent.
Similar cases also exist for the audio source, where the audio information might be completely irrelevant to its spatial counterpart.
Thus, in our classification nets, the spatial information should be treated as the main force,
while the other two can only be its subordinates.
This is the reason why we recombine S, T and A sources to S (no change), ST, and SA, respectively.

Considering that all S, ST, and SA classification nets have already been trained on training instances labeled with video category tags only,
most of these training instances usually perform very well if we feed them into these nets for testing.
However, the CAMs derived from these nets are still rather different in essence
because their inputs are different,
and we have demonstrated some of the most representative qualitative results in Fig.~\ref{fig:NetworkQ}.

\begin{figure}[!b]
  \centering
  \vspace{-0.6cm}
  \includegraphics[width=0.7\linewidth]{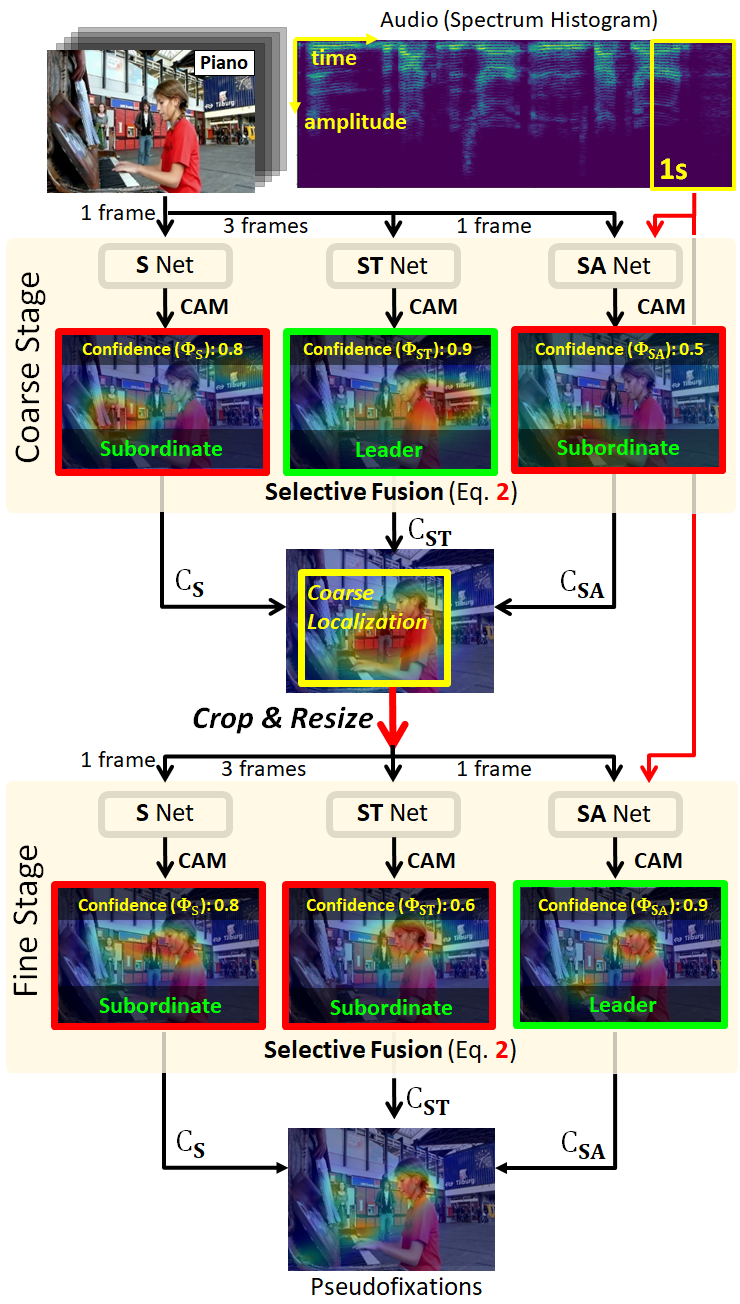}
  \vspace{-0.2cm}
  \caption{Our \textbf{s}elective \textbf{c}lass \textbf{a}ctivation \textbf{m}apping (SCAM) follows the coarse-to-fine methodology, where the coarse stage localizes the region of interest, and then the  fine stage reveals those image regions with the strongest local responses. S: spatial; ST: spatiotemporal; SA: Spatial-audio. The structures of S/ST/SA nets can be found in (A-C) of Fig.~\ref{fig:Networks}.}
  \label{fig:SCAM}
  \vspace{-0.4cm}
\end{figure}

\begin{figure*}[!t]
  \centering
  \includegraphics[width=0.85\linewidth]{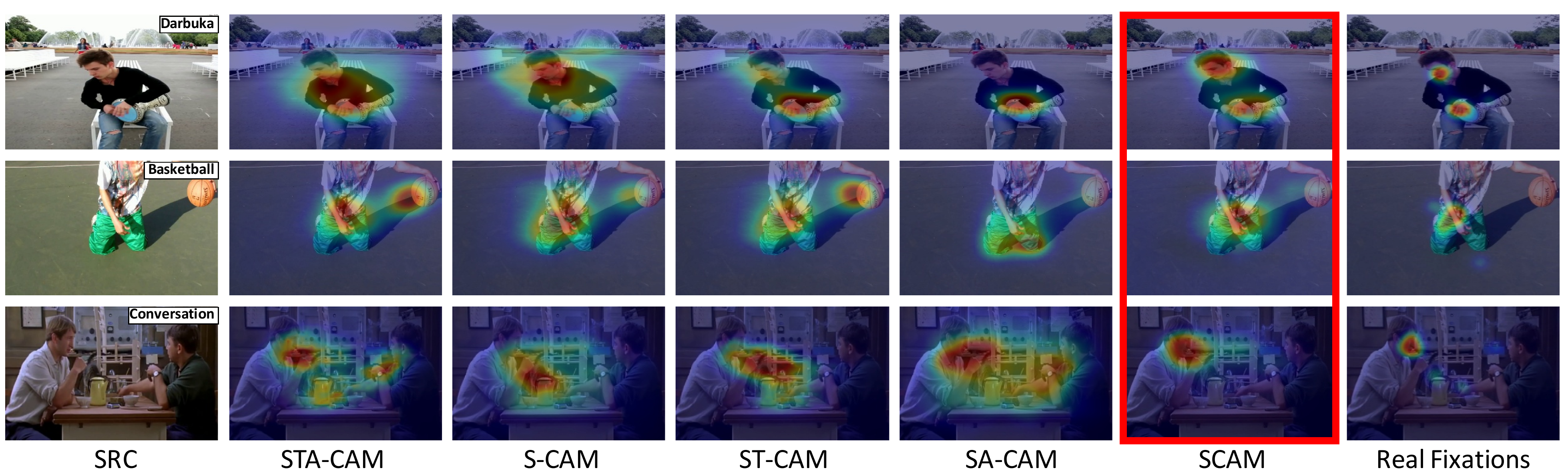}
  \vspace{-0.3cm}
  \caption{Qualitative illustration of CAMs derived from different sources.
  `STA(S/SA/ST)-CAM': CAM obtained from the spatial-temporal-audio (spatial/spatial-audio/spatiotemporal) circumstances;
  the `SCAM' represents the pseudofixations obtained by Eq.~\ref{eq:FU},
  where we can easily observe that the results in this column can be very consistent with the GTs.}
  \label{fig:NetworkQ}
  \vspace{-0.4cm}
\end{figure*}

Normally, the consistency level between multisource CAMs and real fixations is often positively related to the classification confidence level.
By using the classification confidences as the fusion weights to selectively compress those less trustworthy CAMs,
the pseudofixations obtained by selectively fusing all these multisource CAMs (Eq.~\ref{eq:FU}) can be much closer to the real fixations.

\subsection{Multistage SCAM}
\label{sub:MStage}
Benefiting from selective fusion over multiple sources,
the proposed SCAM (Eq.~\ref{eq:FU}) is able to outperform the conventional CAM in revealing pseudofixations.
However, the pseudofixations produced by SCAM may still differ from the real fixations occasionally,
especially for scenes with complex backgrounds,
where the pseudofixations tend to introduce errors.
There are two main reasons:\\
$\bigcdot$ A complex video scene usually contains more content,
yet it has been assigned with only one category tag;
thus, more content belonging to the out-of-scope categories could also contribute to the current classification task.\\
$\bigcdot$ The aforementioned SCAM has followed the single-scale procedure,
while, in sharp contrast,
real human visual system is a multiscale system,
where we tend to quickly locate the region of interest unconsciously before assigning our real fixations to the local regions \cite{lu2018revealing}.


To further improve,
we follow the coarse-to-fine methodology to sequentially perform SCAM twice.
The coarse stage decreases the given problem domain;
therefore, the pseudofixations revealed in the fine stage (\textit{i.e.}, the 2nd stage) are more likely to correlate to the most discriminative regions, and thus
improve the overall performance significantly.

As seen in Fig.~\ref{fig:SCAM}, the \textbf{\emph{coarse}} stage uses a rectangular box ($\rm Coarse\ Location$)
to tightly warp the pseudofixations that have been binarized by a hard threshold ($2\times average$),
and the given input video sequences will be cropped into video patches via these rectangular boxes.
In the \textbf{\emph{fine}} stage,
the video sequences are replaced by these video patches to be the classification nets' input,
and we perform SCAM again to obtain the sourcewise pseudofixations.

\vspace{-0.3cm}
\subsection{Fusion Modules in SCAM}
\label{sec:FMAIS}
All networks adopted in this paper have followed the simplest
encoder-decoder architecture. Following the previous
work~\cite{tavakoli2019dave}, we have converted audio signals to 2D
spectrum histograms (${\rm U}_i$) in advance, which will latterly be fed to the off-the-shelf VggSound ($\mathcal{F}_{\text {VggSound}}$)~\cite{chen2020vggsound} to
obtain the corresponding deep features ($a_i$), and similarly, the spatial deep features can be obtained by feeding individual video frame (${\rm I}_i$) to the off-the-shelf VggNet ($\mathcal{F}_{\text {VggNet}}$)~\cite{Karen_2015_iclr},
and these 2 processes can be formulated as follows:
\begin{equation}
\label{eq:spatialfea}
{a}_{i}=\mathcal{F}_{\text {VggSound}}\left({\rm U}_{i}\right),\ \ \ {s}_{i}=\mathcal{F}_{\text {VggNet}}\left({\rm I}_{i}\right).
\end{equation}

We believe all these implementations are quite
simple and straightforward, and almost all network details have been
clearly represented in Fig.~\ref{fig:Networks}.
Enhanced alternatives, of course, could result in additional performance gains.
Next, we shall provide the detailed architectures towards the SA/ST fusion
modules adopted in the SA/ST classification nets, \textit{i.e.}, the first 2 of Fig.~\ref{fig:FusionM}.

\vspace{0.2cm}
\noindent \textbf{SA Fuse Module.}
The primary target of the SA fuse module is to integrate spatial information with audio signals.
Suppose the input size of the adopted feature backbone is $256\times256$, the exact size of the spatial feature ${\rm s}_{i}$ should be $\{8\times8\times2048\}$, and the size of the audio feature ${\rm a}_{i}$ is $1\times 8192$.
We use a series of deconvolutions to convert ${\rm a}_{i}$'s size to be identical to that of ${\rm s}_{i}$.
The detailed SA fusion process can be detailed as follows:
\begin{equation}
\label{eq:SAFlow}
{sa_i}= Relu\bigg(\sigma\Big(DeConv\big(\phi({a_i})\big)\Big)\odot {s_i} + {s_i}\bigg)\otimes s_i,
\end{equation}
where $\otimes$ is the feature concatenation operation; $DeConv(\cdot)$ is the deconvolution operation; $\odot$ represents the elementwise multiplicative operation; $\sigma(\cdot)$ is the sigmoid function; and $Relu$ is the ReLU function.
$\phi(\cdot)$ is a `binary switch' (\textit{i.e.}, the proposed \textbf{audio switch}, which will be detailed later) to eliminate those irrelevant audio signals.
The rationale of the proposed SA fuse module is quite clear, where the audio signal serves its spatial counterpart as semantic attention to compress those less discriminative frame regions.

\vspace{0.2cm}
\noindent \textbf{ST Fuse Module.}
The methodology of the ST fuse module is quite similar to that of the SA fuse module, and the major difference is that the ST fuse module omits the audio switch.
The main reason is that temporal information could be more likely to benefit its spatial counterpart, yet the audio signal could completely conflict with the spatial clue, resulting in learning ambiguity.
The technical details of ST fusion can be seen in the middle column of Fig.~\ref{fig:FusionM}, and its dataflow can be formulated as:
\begin{equation}
\label{eq:STFlow}
{st_i}= Relu\big(\sigma({v_i})\odot {s_i} + {s_i}\big)\otimes s_i,
\end{equation}
where $v_i$ represents the temporal information after using the 3D convolution,
other symbols and operations are completely identical to those of Eq.~\ref{eq:SAFlow}.

\vspace{0.2cm}
\noindent \textbf{Audio Switch.} The main function of this module is to alleviate the potential side effects from the audio signal when performing visual-audio fusion.
Compared with the temporal source,
the audio source is usually less informative yet associated with strong semantic information,
making it easier to influence its spatial counterpart.
However, the audio source itself has a critical drawback,
where video sequences may usually couple with meaningless background music or noise.
In such cases, fusing audio sources with spatial sources may make the classification task more difficult.
In fact, the nature of the proposed `audio switch' is a plug-in tool,
which can be achieved via an individual network with identical structure to the `SA' classification net.
Instead of aiming at the video classification task,
this plug-in is trained on visual-audio data with binary labels as the learning objective, which indicates whether the current audio signal truly benefits the spatial source.
To obtain these binary labels automatically,
we resort to an off-the-shelf audio classification tool (VggSound~\cite{chen2020vggsound}),
which was trained on a large-scale audio classification set including almost 300 categories.
Our rationale is that the audio source could be able to benefit the spatial source only
if it has been synchronized with its spatial counterpart,
sharing an identical semantical information.

Therefore, for a visual-audio fragment (1 frame and 1 second audio),
we assign its binary label to `1' if the audio category predicted by the audio classification tool is identical to the pregiven video category;
otherwise, we assign its binary label to `0'.

\vspace{0.2cm}
\noindent \textbf{Classifiers.}
All classifiers adopted in the abovementioned networks are plain classifiers, which consist of only 2 steps.
First, we use a $1\times1$ convolution to refine the channel size of the input feature tensor (\textit{i.e.}, $s_i$, $sa_i$, and $st_i$) to $c$.
Next, we perform the \textbf{g}lobal \textbf{a}verage \textbf{p}oolig (GAP) operation to transform the tensor to a vector, whose dimension is identical to the video category number $c$.

\vspace{0.2cm}
\noindent \textbf{Classification Loss.}
All classification networks in SCAM have adopted an identical loss function, \textit{i.e.}, the standard \textbf{M}ultilabel \textbf{S}oft \textbf{M}argin (MSM) loss~\cite{wang2019zero, li2020group}.
For better clarification, we take the `ST net' (Fig.~\ref{fig:Networks}-C) as an example here, and its classification loss can be seen as follows:
\begin{equation}
\label{eq:classloss}
\begin{aligned}
\begin{split}
\mathcal{L}_{\mathrm{MSM}}(st_i,{\rm VC})=
&-\frac{1}{c}\sum_{i=1}^{c}{\rm VC}_{i}\times \log \Bigg( \frac{1}{1 + e^{-st_{i}}} \Bigg)\\[-0.5ex]
&+\left(1-{\rm VC}_{i}\right)\times \log \Bigg(\frac{e^{-st_{i}}}{1+e^{-st_{i}}}\Bigg),
\end{split}
\end{aligned}
\end{equation}
where `$st_i$' is the input of the `classifier', and it can be obtained via Eq.~\ref{eq:STFlow},
$\rm VC$ represents the exact class towards the given input, $c$ is the total video category number.
For all classifiers adopted in this paper, we use the same loss function as Eq.~\ref{eq:classloss}.


\vspace{-0.3cm}
\subsection{SCAM vs. Real Fixation}
\label{sec:SCAMRF}
The proposed SCAM is mainly based on performing selective fusion over discriminative regions revealed by the sourcewise classification nets, \textit{i.e.}, S, ST, and SA nets.
Although the pseudofixation maps generated by SCAM basically correlate to the most discriminative regions in the given visual-audio fragment, they might occasionally be different from real human-eye fixations (GTs), as demonstrated in Fig.~\ref{fig:SCAMandSCAM+}, \textit{i.e.}, SCAM vs. GTs.

\begin{figure}[!h]
  \centering
  \includegraphics[width=0.9\linewidth]{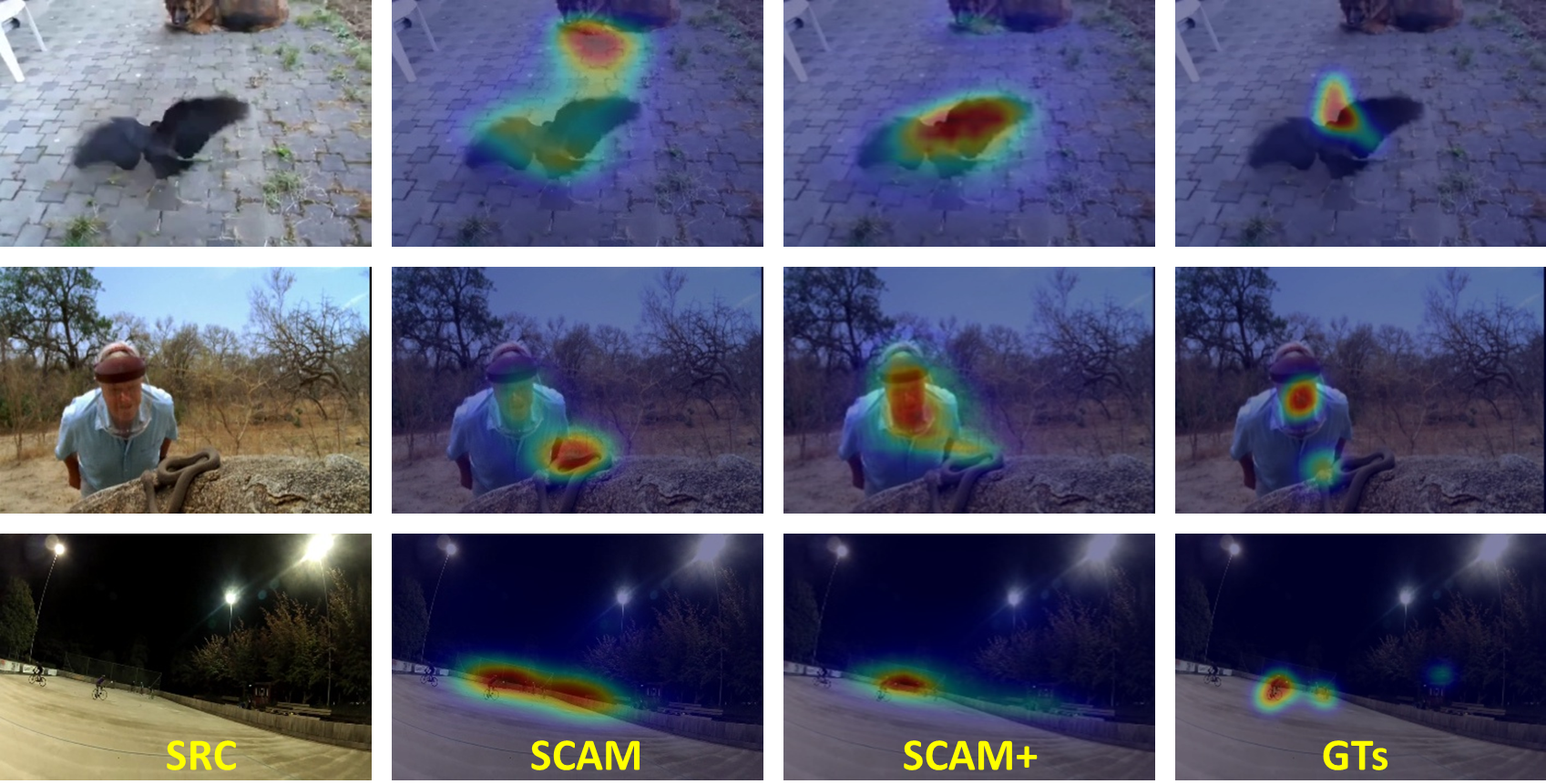}
  \vspace{-0.3cm}
  \caption{Visual comparison of local information- and global information-based fusion.
  It can be observed that benefiting from the relational constraints between frames of global info,
  global info-based method is capable of handling various challenging factors,
  such as location bias, color bias and multiple objects.
  \vspace{-0.4cm}}
  \label{fig:SCAMandSCAM+}
\end{figure}

From the data perspective, the total input of SCAM computation only comprises 3 neighboring video frames and a 1 second audio signal at most, which makes the proposed SCAM a typical \textbf{\emph{local}} method.
However, the real human visual system follows a \textbf{\emph{global}} manner, where the real fixations are jointly influenced by multiple factors (\emph{e.g.}, short/long-term memory).
Thus, the challenge for using SCAM to produce high-quality pseudofixations relies on including the sourcewise selection process (Eq.~\ref{eq:FU}) with \textbf{\emph{global}} information, \textit{i.e.}, multi-granularity information. We shall provide an in-depth explanation here.

First, the human visual system tends to omit those image regions that are less salient from the sourcewise perspective, \emph{e.g.}, that occurred repeatedly in the given video sequence, even the most discriminative regions in the current visual-audio fragment.
For example, as seen in the 1st row of Fig.~\ref{fig:SCAMandSCAM+}, the real fixations have avoided the `\emph{dog}' since this `\emph{dog}' has occurred multiple times before, making it less salient than the `\emph{hawk}'.
However, the SCAM has only considered the current local visual-audio fragment, and thus its pseudofixation map still treats the `\emph{dog}' as the salient one because, from the local perspective, the `\emph{dog}' could also contribute greatly to the current classification task.
Therefore, we incorporate both short-term and long-term information into the SCAM.

Second, the real fixation is influenced by the high-level semantic information that has been embedded in our brain~\cite{sun2020mining,li2020group,wang2019zero}.
In fact, the human visual system never works alone, where the human brain plays a critical role, \textit{i.e.}, the real human fixation tends to vary between subjects with different ages, agendas, careers, \textit{etc.}.
The main reason is that the high-level semantic knowledge that we have learned in our daily life can directly determine where we really look at.
Despite the existence of this phenomenon, the principle regards the positive relationship between being discriminative and being salient still holds, and to further refine the pseudofixations derived by SCAM, we shall introduce the high-level semantic information into the SCAM.
Therefore, we propose to consider the cross-term information, which could be obtained by introducing multiple visual-audio fragments cropped from other video sequences sharing an identical tag with the given sequence.
Thus, taking such additional cross-term information during the classification task can help us to compress those relatively less discriminative regions, which could eventually improve the pseudofixations.

In summary, perfect pseudofixations generation framework should consider not only the multisource and multiscale aspects but also the multi-granularity perception, where all \emph{short-term}, \emph{long-term}, and \emph{cross-term} information should be considered when performing SCAM.
Thus, we present the upgraded version of the SCAM, named SCAM+, which will be detailed in the next section.


\vspace{-0.2cm}
\section{Selective Class Activation Mapping +}
\label{sec:SCAM+}
\subsection{SCAM+ Overview and Technical Pipeline}
As seen in Fig.~\ref{fig:Networks}, the major difference between SCAM and SCAM+ relies on the adopted classification nets, where only 3 classification nets, \textit{i.e.}, S, ST, and SA (subfigures A-C), have been adopted by SCAM, while based on SCAM, the SCAM+ has additionally adopted 3 `\textbf{\emph{types}}' of classification net, \textit{i.e.}, S+, SA+, and ST+ (subfigures E-G).
Note that the combinations of short/long/cross-term information are used as the input of these new classification nets, leading to 8 additional classification nets with 8 classification confidences ($\rm C$) and CAM maps ($\Phi$), including $\rm \{C,\Phi\}^{Short}_{S+}$, $\rm \{C,\Phi\}^{Long}_{S+}$, $\rm \{C,\Phi\}^{Cross}_{S+}$, $\rm \{C,\Phi\}^{Short}_{ST+}$, $\rm \{C,\Phi\}^{Long}_{ST+}$, $\rm \{C,\Phi\}^{Cross}_{ST+}$, $\rm \{C,\Phi\}^{Long}_{SA+}$, and $\rm \{C,\Phi\}^{Cross}_{SA+}$.
The first 3 of them can be derived from the S+ net, the middle 3 can be obtained from the ST+ net, and the last 2 are formulated via the SA+ net.
Specifically, we have omitted $\rm \{C,\Phi\}^{Short}_{SA+}$ because the audio signal itself has already covered the short-term information.
The SCAM+-based pseudofixations can be obtained via the following equation.

\begin{equation}
\label{eq:SCAM+}
\begin{aligned}
{\rm SC}&{\rm AM}+ =\\
&\frac{\big\lVert{{\rm \textcolor{red}{UC}} \odot {\rm \textcolor{red}{UR}} + \rm SC} \odot {\rm SR} + {\rm LC} \odot {\rm LR} + {\rm CC} \odot {\rm CR}\big\rVert_1 + \lambda}{\big\lVert{\rm  \textcolor{red}{UC}+SC+LC+CC}\big\rVert_1 + \lambda},\ \ \ \ \ \ \ \ \ \ \\[-0.5ex]
{\rm SR}&: \Big[{\rm \Phi}_{\rm S+}^{\rm Short}\{i\}, {\rm \Phi}_{\rm ST+}^{\rm Short}\{i\}\Big],\\[-0.5ex]
{\rm LR}&: \Big[{\rm \Phi}_{\rm S+}^{\rm Long}\{i\}, {\rm \Phi}_{\rm ST+}^{\rm Long}\{i\}, {\rm \Phi}_{\rm SA+}^{\rm Long}\{i\}\Big],\\[-0.5ex]
{\rm CR}&: \Big[{\rm \Phi}_{\rm S+}^{\rm Cross}\{i\}, {\rm \Phi}_{\rm ST+}^{\rm Cross}\{i\}, {\rm \Phi}_{\rm SA+}^{\rm Cross}\{i\}\Big],\\[-0.5ex]
{\rm SC}&: \bigg[\oint\big({\rm C}_{\rm S+}^{\rm Short}\{i\}\big), \oint\big({\rm C}_{\rm ST+}^{\rm Short}\{i\}\big)\bigg],\\[-0.5ex]
{\rm LC}&: \bigg[\oint\big({\rm C}_{\rm S+}^{\rm Long}\{i\}\big), \oint\big({\rm C}_{\rm ST+}^{\rm Long}\{i\}\big), \oint\big({\rm C}_{\rm SA+}^{\rm Long}\{i\}\big)\bigg],\\[-0.5ex]
{\rm CC}&: \bigg[\oint\big({\rm C}_{\rm S+}^{\rm Cross}\{i\}\big), \oint\big({\rm C}_{\rm ST+}^{\rm Cross}\{i\}\big), \oint\big({\rm C}_{\rm SA+}^{\rm Cross}\{i\}\big)\bigg],
\end{aligned}
\end{equation}
where UC and UR can be found in Eq.~\ref{eq:FU}, and the meanings of $\lVert\cdot\rVert_1$, $\lambda$, $\rm\Phi$, $\oint$, and $\odot$ are identical to that of Eq.~\ref{eq:FU}.

Clearly, the proposed SCAM+ has included more granularity when producing pseudofixations.
To facilitate a better reading, 2 technical details should still be provided: 1) the exact definition and implementation of the abovementioned \emph{\textbf{multi-granularity information}} (Sec.~\ref{sec:muid}) and 2) the exact network architectures, especially the \emph{\textbf{fusion parts}}, of adopted classification nets (Sec.~\ref{sec:mulp}). These 2 aspects will be detailed in the following subsections.

\begin{figure}[!t]
  \centering
  \includegraphics[width=1\linewidth]{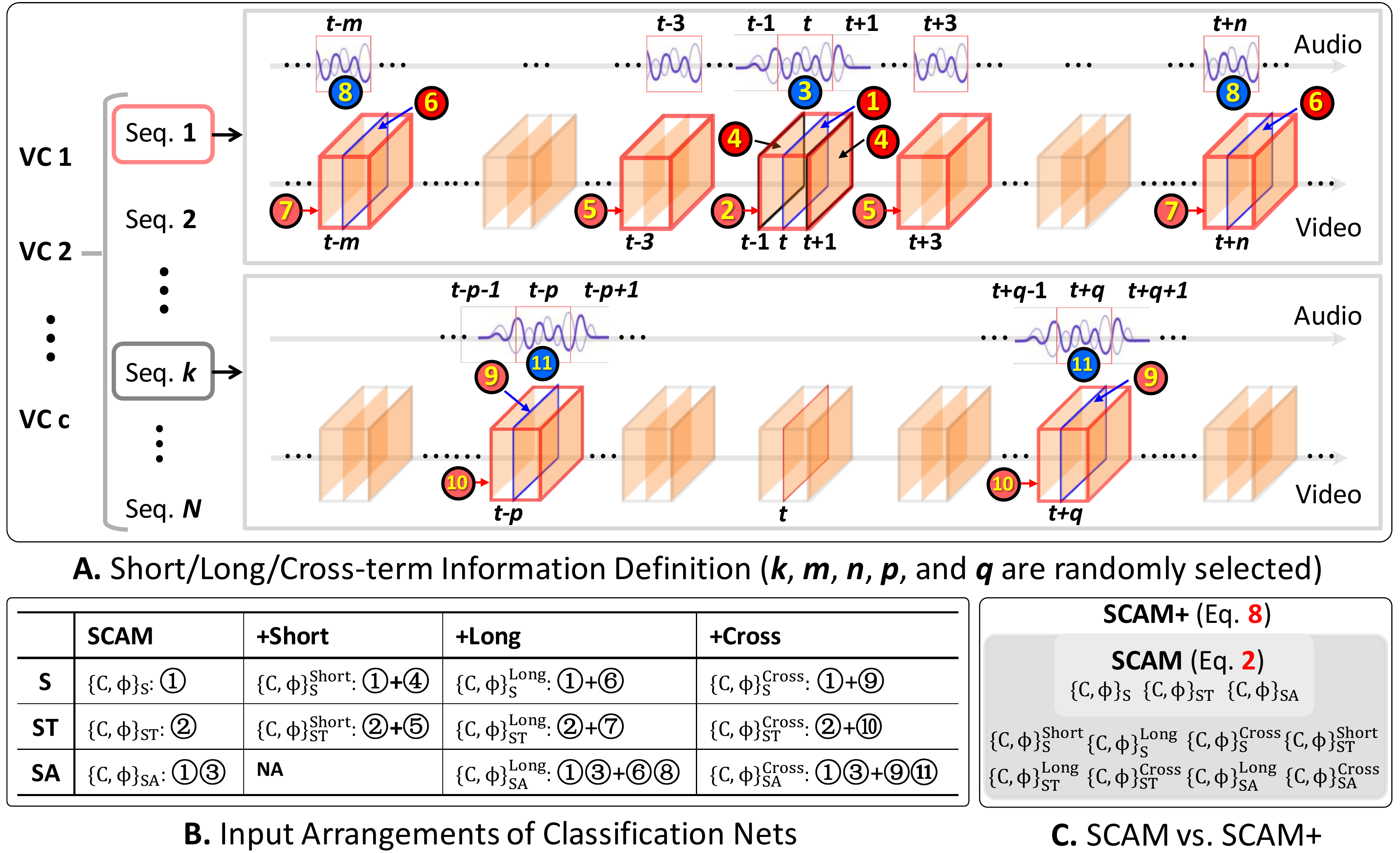}
  \vspace{-0.8cm}
  \caption{Input data formulations/definitions of SCAM and SCAM+.
  VC\ c: the $c$-th video category classes; Seq.k: the $k$-th video sequence.}
  \vspace{-0.4cm}
  \label{fig:pipe}
\end{figure}

\vspace{-0.2cm}
\subsection{Multigranularity Information Definition}
\label{sec:muid}
Different from the \textbf{\emph{local}} information (\textit{i.e.}, 3 consecutive frames and a 1 second audio signal, which we call \textbf{\emph{target}} data) that the SCAM has solely considered, the \emph{\textbf{multi-granularity information}} of SCAM+ mentioned in this paper includes 3 types of data formulation, and we shall give their definitions and implementations.

\vspace{0.2cm}
\noindent \textbf{Short-term Information.}
In our approach, the short-term information acts as the \emph{\textbf{human short-term memory}} to automatically suppress those less important scene contents and focus our attention on the remaining conspicuous ones.
As seen in Fig.~\ref{fig:pipe}-A, suppose the $t$-th video frame is the \textbf{\emph{target}} frame marked by {\large{\textcircled{\small{1}}}}; its short-term spatial-temporal information can be detailed as the neighboring 8 video frames marked by {\large{\textcircled{\small{2}}}} + {\large{\textcircled{\small{5}}}}, and all 9 frames together can be divided into 3 groups to be fed into a three-stream classification net, \textit{i.e.}, the ST+ net detailed in Fig.~\ref{fig:Networks}-G.
Specifically, from the sole spatial source perspective (\textit{i.e.}, the S+ net, Fig.~\ref{fig:Networks}-E), the short-term information is the 2 neighboring frames of the target frame, \textit{i.e.}, the mark {\large{\textcircled{\small{4}}}}, which could be completely identical to the input of the ST net (Fig.~\ref{fig:Networks}-C) adopted in the SCAM, yet the underlying rationale is different in essence.
The ST net simply resorts to a 3D convolution to acquire spatial-temporal information, yet in sharp contrast, the S+ net has mainly aimed to learn the frame-level similarity relationship measurement to mimic the perceptual mechanism of the real human attention mechanism.
Thus, both S+ net and ST net are essentially complementary to each other.

\vspace{0.2cm}
\noindent \textbf{Long-Term Information.}
As a complement part to the short-term information, the long-term information also serves the human visual system to focus our attention on the most conspicuous scene regions in the entire video sequence.
Similar to the short-term information defined above, we treat those video frames and audio signals that are \textbf{\emph{far}} away from the current target frame over the time scale as the long-term information.
Suppose the $t$-th frame is the current target frame, the scope of long-term information could be any frame in the same video sequence with the target frame's neighbor frames excluded (\textit{i.e.}, 8 frames).
Since the exact formulation of long-term information could vary with different sources, we shall provide their details here. From the spatial perspective, the short-term information of the $t$-th frame is 2 other frames that are temporally far away from the $t$-th frame and can be randomly selected (e.g., the mark {\large{\textcircled{\small{6}}}} in Fig.~\ref{fig:pipe}-A), and these 3 frames (1 target frame and 2 short-term frames) will be fed into the S+ net, as shown in Fig.~\ref{fig:Networks}-E.
Similarly, from the spatial-temporal perspective, the short-term information becomes 2 groups of 3 consecutive frames (\emph{e.g.}, the mark {\large{\textcircled{\small{7}}}} in Fig.~\ref{fig:pipe}-A), which are also randomly selected. Thus, a total of 9 frames (3 of them are the target consecutive frames, and the other 6 frames are the long-term frames) will be fed into the ST+ net, as shown in Fig.~\ref{fig:Networks}-G.
With regard to the visual-audio perspective, the only difference is that the audio signals have been additionally considered (\emph{e.g.}, the mark {\large{\textcircled{\small{8}}}} in Fig.~\ref{fig:pipe}-A); thus, a total of 3 frames and 3 s audio signals are fed into the SA+ net, as demonstrated in Fig.~\ref{fig:Networks}-F.

\vspace{0.2cm}
\noindent \textbf{Cross-term Information.}
As has been mentioned before, the real human visual system is influenced by the associative memory --- the accumulated basic impression towards different object categories.
For example, we have already known what a car looks like, given video sequences containing multiple cars, we might pay more attention to that one with the most distinctive appearance, this procedure requires more information towards the current video category as the subordinate.
This phenomenon motivates us to consider the cross-term information, we can regard those frames in other video sequences with an identical video tag to the current sequence as the cross-term information.
As seen in Fig.~\ref{fig:pipe}-A, from the sole spatial domain perspective, the cross-term information of the $t$-th frame in sequence 1 (with video category 2) can be any 2 frames randomly selected from other video sequences $k$ that are also assigned with tag 2, \emph{e.g.}, the mark {\large{\textcircled{\small{9}}}}.
Similarly, from either the spatial-temporal or visual-audio perspective, the cross-term information can be formulated accordingly as either mark {\large{\textcircled{\small{10}}}} or mark {\large{\textcircled{\small{11}}}}.

\vspace{-0.3cm}
\subsection{Multigranularity Perception}
\label{sec:mulp}
Compared with the SCAM, the upgraded version SCAM+ has adopted several new classification nets, \textit{i.e.}, S+, SA+, and ST+.
As seen in subfigures E-G of Fig.~\ref{fig:Networks}, all these classification networks have directly adopted partial of the SCAM's classification nets (\textit{i.e.}, S, SA, and ST) as their backbones, \textit{i.e.}, S/SA/ST-Branches.
Since these feature backbones are almost unaltered, we focus on the subsequent fusion parts.


{At present, there exist multiple ways to model cross-data relationships, \emph{e.g.}, \textbf{r}ecurrent \textbf{n}eural \textbf{n}etwork (RNN)~\cite{Xia2021evalu}, \textbf{l}ong \textbf{s}hort-\textbf{t}erm \textbf{m}emory (LSTM)~\cite{yang2019visual, sun2019visual}, and \textbf{g}ate \textbf{r}ecurrent \textbf{u}nit (GRU)~\cite{zhang2020neur}. Although these existing tools can sense subtle changes over time, they are clearly not suitable in our case because these tools require their input data to be ``spatially'' well aligned (two consecutive video frames as spatially well-aligned). However, in our case, it is almost infeasible to ensure such spatial alignment two frames belonging to different video sequences (e.g., our cross-term information).}
Although these existing tools can sense subtle changes over time, they are clearly not suitable in our case because these tools require their input data to be spatially well aligned.
However, in our case, it is almost infeasible to ensure such spatial alignment.
Moreover, in view of the human attention mechanism, the most valuable short-/long-/cross-term information could be the high-level semantic part, which motivates us to model the semantic relationship between target data and its short-/long-/cross-term company.
To achieve this, we devise a variant of \textbf{g}raph \textbf{c}onvolution \textbf{n}etwork (GCN)~\cite{Kipf_2017_iclr} to combine the multi-granularity information with sourcewise information, named conservative-GCN (C-GCN), whose major highlights include 1) a completely new methodology for performing $m$-step \textbf{\emph{semantic reasoning}} over multisource and multi-granularity nodes and 2) a completely new \textbf{\emph{fixation refine layer}} to ensure that the whole process is conservative enough to eliminate redundant feature responses.

\vspace{-0.2cm}
\subsubsection{Semantic Reasoning}
\label{sub:NMP}
The proposed C-GCN (Fig.~\ref{fig:CGCNpipe}) consists of multiple nodes and edges
$\mathcal{G}=(\mathcal{V}, \mathcal{E})$, in which the $\mathcal{V}$ represents the node set
$\left\{\rm TA, MG_1, \ldots, MG_{\emph n}\right\}$, where TA denotes the target node (\textit{i.e.}, the current input visual-audio fragment), and
MGs include $n$ nodes correlating to short/long/cross information.
We use $\mathcal{E}$ to represent the edge set
$\left\{e_{i,j}\right\}$, $i\ne j$; and $e_{{\rm TA}, {\rm MG}_{\emph{i}}} \in \mathcal{E}$ is the edge between $\rm TA$ and $\rm MG_{\emph{i}}$.

\begin{figure}[!h]
  \centering
  \includegraphics[width=1\linewidth]{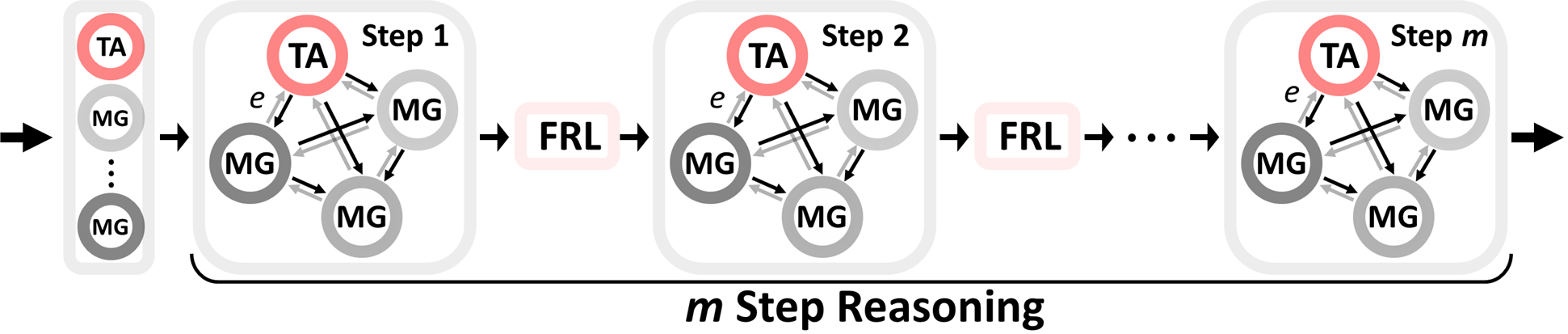}
  \vspace{-0.8cm}
  \caption{Network architecture of the proposed C-GCN, which mainly consists of 2 parts, \textit{i.e.}, 1) semantic reasoning (Sec.~\ref{sub:NMP}), and 2) \textbf{f}ixation \textbf{r}efine \textbf{l}ayer (FRL, Sec.~\ref{sub:frl}).
  \vspace{-0.2cm}}
  \label{fig:CGCNpipe}
\end{figure}

As seen in Fig.~\ref{fig:Networks} (E-G), the input of the C-GCN consists of 3 parts, \textit{i.e.} the output of the S/SA/ST branch ($s_i,sa_i,st_i$) containing rich semantic information, and for simplicity, we uniformly denote them as $h_i^0$, where the subscript $i$ denotes that this feature belongs to the $i$-th node, the superscript 0 denotes the beginning of semantic reasoning.
In the $t$-th reasoning stage, we use the \emph{interattention} ($F_{iatt}$, Eq.~\ref{eq:fiatt}) to model the semantic relation of each 2 of the input nodes, \textit{i.e.}, the edge $e_{i, j}$, which can be formulated as follows:
\begin{equation}
\begin{aligned}
\boldsymbol{e}_{{\rm TA}, {\rm MG_{\emph{i}}}}^{t}= Softmax\big(F_{iatt}(h_{\rm TA}^t, h_{\rm MG_{\emph{i}}}^t)^\top\big),\\
\boldsymbol{e}_{{\rm MG_{\emph{i}}},{\rm TA}}^{t}= Softmax\big(F_{iatt}(h_{\rm MG}^t,h_{\rm TA_{\emph{i}}}^t)^\top\big),
\end{aligned}
\end{equation}
where $\boldsymbol{e}_{{\rm TA}, {\rm MG_{\emph{i}}}}^{t}\in\mathbb{R}^{1024\times 1024}$,  $\top$ is
the matrix transpose operation, $F_{iatt}(\cdot,\cdot)$ measures the consistency between its input and can be detailed as follows:
\begin{equation}
\begin{aligned}
\label{eq:fiatt}
F_{iatt}(h_{\rm TA}^t,h_{\rm MG_{\emph{i}}}^t) &= \Big[\mathcal{R}_{28\times1024}\big(Conv_{1\times1}(h_{\rm TA}^t)\big)\Big]^{\rm \top}\\
& \circledast \Big[\mathcal{R}_{28\times1024}\big(Conv_{1\times1}(h_{\rm MG_{\emph{i}}}^t)\big)\Big],
\end{aligned}
\end{equation}
where $\circledast$ denotes matrix multiplication, $h\in\mathbb{R}^{28\times32\times32}$,
32 is the size of feature map, 28 is the total video category number ($c$),
$\mathcal{R}_{28\times1024}(\cdot)$ reshapes its input to a matrix with size $28\times1024$,
and $Conv_{1\times1}$ is a typical $1\times 1$ convolution.

Actually, the rationale of edge $\boldsymbol{e}_{{\rm TA}, {\rm MG_{\emph{i}}}}^{t}$ is the feature similarity between two neighboring nodes.
Since the primary functional of the reasoning process is to exchange/share information between nodes and formulate a series of implicit principles towards the given learning objective, we update all nodes' statuses after each reasoning stage.
We formulate the updating process towards the target node (TA) using the following equation:
\begin{equation}
\label{eq:updateTA}
\begin{aligned}
&h_{\rm TA}^{t+1}\gets\\
&\mathcal{R}_{28\times32^2}\bigg(h_{\rm TA}^{t}\odot \sigma\big(\sum_{i\in\mathcal{N}_{\rm TA}} \mathcal{R}_{28\times1024}(h_{\rm MG_{\emph{i}}}^t)\circledast \boldsymbol{e}_{{\rm TA}, {\rm MG_{\emph{i}}}}^{t}\big)\bigg),
\end{aligned}
\end{equation}
where $\sigma(\cdot)$ is a typical sigmoid function;
$\circledast$ denotes matrix multiplication; $\odot$ represents elementwise multiplicative operation;
$\times$ is the standard multiplication;
$\mathcal{N}_{\rm TA}$ includes all nodes neighboring the target node (TA),
and $\mathcal{R}_{28\times32^2}(\cdot)$ reshapes its input to a tensor with size $28\times32^2$.
Similarly, the updating process towards the multi-granularity nodes (\emph{e.g.}, ${\rm MG}_{\emph{i}}$) can be formulated as follows:
\begin{equation}
\begin{aligned}
\label{eq:updateMG}
&h_{\rm MG_{\emph{i}}}^{t+1}\gets
\mathcal{R}_{28\times32^2}\Big(h_{\rm MG_{\emph{i}}}^{t}\odot \sigma\big(\sum_{j\in\mathcal{N}_{\rm MG_{\emph{i}}}} \mathcal{R}_{28\times1024}(h_{\rm MG_{\emph{j}}}^t)\\[-1ex]
&\ \ \ \ \ \ \ \ \ \ \circledast \boldsymbol{e}_{{\rm MG_{\emph{i}}}, {\rm MG_{\emph{j}}}}^{t}+\mathcal{R}_{28\times1024}(h_{\rm TA}^t)\circledast \boldsymbol{e}_{{\rm MG_{\emph{i}}},{\rm TA}}^{t}\big)\Big).
\end{aligned}
\end{equation}
\indent By using the nodewise reasoning process, the high-level semantic information between the target input and its multi-granularity information can
be obtained.
The final status of the target node $h_{\rm TA}^{m}$($m$ is the total reasoning steps) will be fed into a classifier.
Actually, the target node's related feature response maps have embedded both multisource and multi-granularity information, and they will be used when formulating the pseudofixations.


\vspace{-0.2cm}
\subsubsection{Fixation Refine Layer}
\label{sub:frl}
In most cases, the C-GCN-based reasoning process can appropriately fuse both multisource and multi-granularity information.
However, there still exists one issue regarding the reasoning process to be handled.
Although the reasoning steps can implicitly incorporate the multi-granularity information, the addition operation-based updating process (\emph{e.g.}, Eq.~\ref{eq:updateTA}) could lead the target node's feature response map
being redundant, resulting in pseudofixations that differ from the real fixations.
Thus, for each reasoning stage, we have additionally assigned one \textbf{f}ixation \textbf{r}efine \textbf{l}ayer (FRL, Fig.~\ref{fig:CGCNpipe}) to eliminate redundant feature responses.
We shall provide the technical details of the proposed FRL as follows.

First, we establish a binary matrix ${\rm rMASK}\in\{0,1\}^{32\times32}$ to indicate which elements in the fused feature tensor have relatively large feature responses, and these qualified elements are more likely to belong to the most discriminative frame regions.
The computation of ${\rm rMASK}$ can be formulated as follows:
\begin{equation}
\label{eq:rmask}
{\rm rMASK} = \Big[max\Big({\rm cMean}\big({h}_{i}^{t}\big)\Big)\times\mathcal{T}_{d} - {\rm cMean}\big({h}_{i}^{t}\big)\Big]_+,
\end{equation}
where $\times$ is the standard multiplication, $\rm cMean(\cdot)$ is a channelwise average operation, which converts a tensor to a matrix, and $max(\cdot)$ is a typical max operation returning the largest value in its input matrix; ${h}_{i}^{t}$ is a tensor feature, which can be obtained by either Eq.~\ref{eq:updateTA} or Eq.~\ref{eq:updateMG}; $[\cdot]_+$ converts all its negative elements to 0 and converts the remaining positive elements to 1; and $\mathcal{T}_{d}$ is a predefined hard threshold.

Next, we use ${\rm rMASK}$ to filter redundant information in ${h}_{i}^{t}$, and this process can be formulated as:
\begin{equation}
\begin{aligned}
\label{eq:updatehh}
{h}_{i}^{t'}&=\frac{1}{2}\times\Big[{h}_{i}^{t} \odot\Big(\mathcal{T}_{r}\times\sigma\left({\rm cMean}({h}_{i}^{t})\right) \odot {\rm rMASK} \\
&+ \sigma\left({\rm cMean}({h}_{i}^{t})\right) \odot \left(1-{\rm rMASK}\right)\Big) +{h}_{i}^{t} \Big],
\end{aligned}
\end{equation}
where $\odot$ is the elementwise multiplicative operation, and $\mathcal{T}_{r}$ is the refinement rate assigned by a predefined value.
The rationale of Eq.~\ref{eq:updatehh} can be explained as follows: since the rMASK can indicate those spatial regions with large feature responses, we use it as an attention filter ($\mathcal{T}_{r}$ controls the compression rate) for those less trustworthy (\textit{i.e.}, with relatively low responses) regions --- those regions are more likely to be redundant.

Specifically, the proposed FRL can further alleviate the feature redundancy problem, making the obtained pseudofixations more consistent with the real fixations. The exact choices of $\mathcal{T}_{d}$ and $\mathcal{T}_{r}$ will be discussed in Sec.~\ref{subsubsec:THFRL}.

\vspace{-0.2cm}
\section{Generic Fixation Prediction Networks}
\label{sec:FPN}
By using the abovementioned SCAM+ (Eq.~\ref{eq:SCAM+}), we can obtain a large number of pseudo visual-audio fixation maps.
Although these maps are usually consistent with the real one, there still exists one critical limitation, \textit{i.e.}, the computation of SCAM+ requires human-provided video tags, leading to a very limited application scope.
Therefore, we shall treat the SCAM+-based pseudo-visual-audio fixation maps as learning objectives to guide knowledge distillation, and thus, we can obtain end-to-end fixation prediction models without using any human-provided video tag.
Clearly, this strategy can make the proposed method a generic visual-audio fixation prediction tool.

\begin{figure}[!t]
  \centering
  \includegraphics[width=0.9\linewidth]{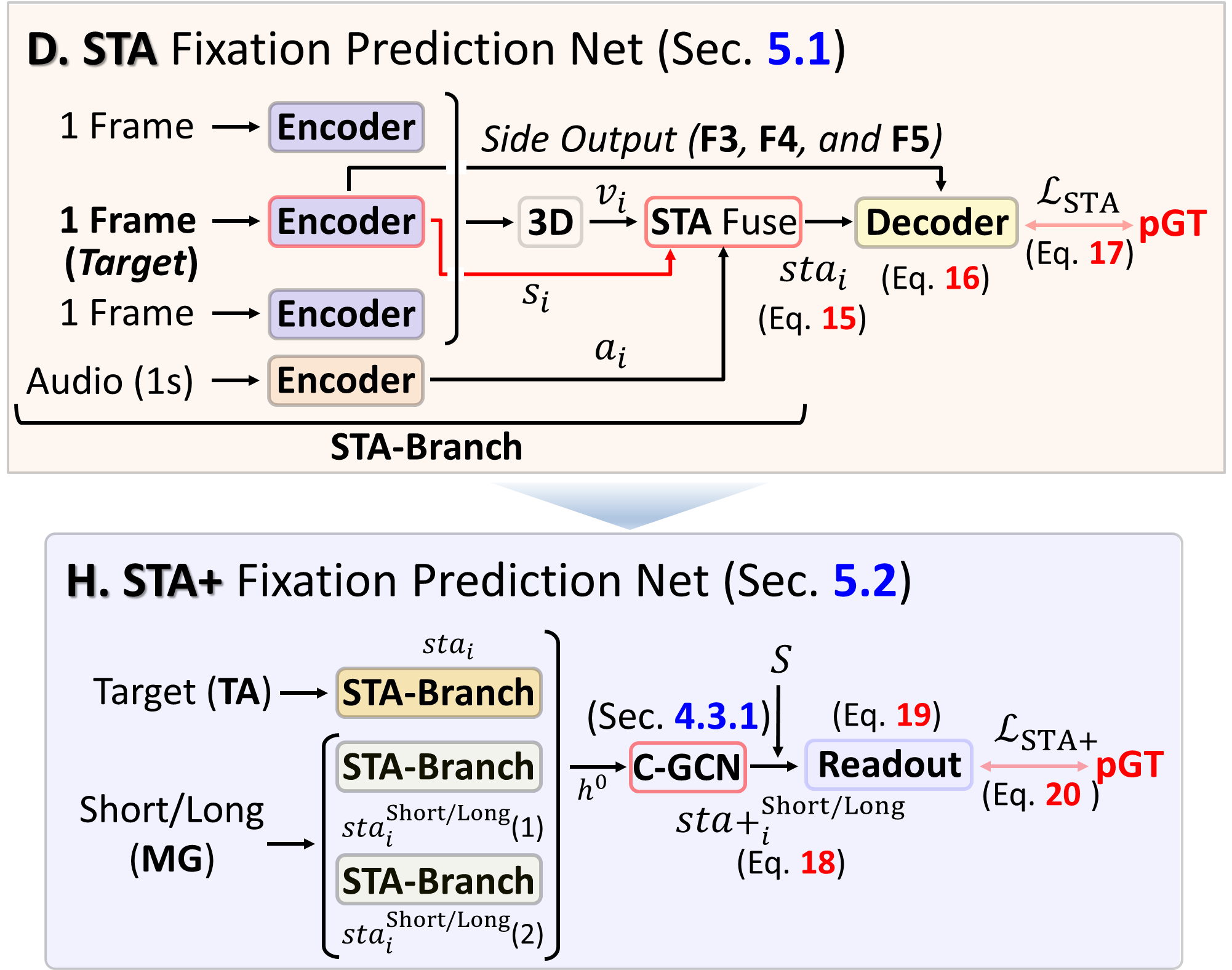}
  \vspace{-0.3cm}
  \caption{Architectures of full automatic visual-audio fixation prediction networks, which take the pseudofixations as training objective, achieving a generic fixation prediction without using video tags. Subfigure H has equipped subfigure D with the proposed multi-granularity perception. Details of STA fusion module can be found in Fig.~\ref{fig:FusionM}.
  \vspace{-0.4cm}}
  \label{fig:VFPNetworks}
\end{figure}

\vspace{-0.2cm}
\subsection{Sourcewise Fixation Prediction Network}
\label{sec:swfpn}
As seen in Fig.~\ref{fig:VFPNetworks}-D, the implementation of \textbf{s}patial-\textbf{t}emporal-\textbf{a}udio (STA) fixation prediction network is very intuitive, where the spatial features ($s_i$) are fused with
either temporal features ($v_i$) or audio features ($a_i$) in advance and later
combined via the simplest feature concatenation operation.
Then, a typical decoder with 3 deconvolutional layers is used to convert
the feature maps derived from the `STA Fuse' module to fixation maps.
The data flow in the `STA Fuse' module can be formulated as
Eq.~\ref{eq:STAFlow}.
\begin{equation}
\label{eq:STAFlow}
\begin{aligned}
&{sta_i}=\\
&Relu\bigg[Cov\bigg(\Big(\sigma\big(\phi({a_i})\big)\odot {s_i} + {s_i}\Big)\otimes \Big(\sigma\big({v_i}\big)\odot {s_i} + { s_i}\Big)\bigg)\bigg],
\end{aligned}
\end{equation}
where $\otimes$ is the typical concatenation operation; $\phi(\cdot)$ is the proposed audio switch (Sec.~\ref{sec:FMAIS}); $Cov(\cdot)$ denotes $1\times1$ convolution; all other symbols are identical to those in Eq.~\ref{eq:SAFlow}.

The $sta_i$ will be fed into the decoder, whose output ($\widehat{sta}_i$) can be formulated as:
\begin{equation}
  \label{eq:decoder}
  \begin{aligned}
    \widehat{sta}_i=
    &\uparrow(Ref(Ref({\rm F3})\otimes\uparrow(Ref(Ref({\rm F4})\otimes\\
    &\ \ \ \ \ \ \ \ \uparrow(Ref(Ref({\rm F5})\otimes sta_i)))))),
    \end{aligned}
  \end{equation}
where $\uparrow(\cdot)$ is the upsampling operation, $Ref(\cdot)$ refines all its input to a fixed channel number (32), $\otimes$ is the feature concatenation operation, and F3, F4, and F5 are the side outputs of the encoder correlated to the target frame.
A more powerful decoder equipped with multiscale connections and
channelwise attention may give rise to some additional performance
gain, but full justification \textit{w.r.t.} this issue is beyond the main
topic of this paper, and we shall leave it for future research
investigation.

For the training process, we choose 2 typical loss functions, \textit{i.e.}, the binary cross-entropy loss ($\mathcal{L}_{\rm BCE}$)~\cite{pan2017salgan} and the Kullback-Leibler divergence loss ($\mathcal{L}_{\mathrm{\rm KL}}$) ~\cite{borji2019saliency}, and thus the overall loss function ($\mathcal{L}_{\rm STA}$) can be formulated simply as:
\begin{equation}
\mathcal{L}_{\rm STA} = \mathcal{L}_{\rm BCE}(\widehat{sta}_i, {\rm pGT})+\mathcal{L}_{\rm KL}(\widehat{sta}_i, {\rm pGT}).
\end{equation}

Clearly, the training process of the proposed STA fixation prediction
network relies on pseudofixations (pGTs) only; thus, it is able to perform end-to-end
fixation prediction for unseen visual-audio sequences without any category
tag.

\vspace{-0.1cm}
\subsection{Multigranularity Fixation Prediction Network}
\label{sec:mgfpn}
As shown in Fig.~\ref{fig:VFPNetworks}-H, the STA fixation prediction network can also be upgraded by combining it with the multi-granularity perception mechanism.
We name the upgraded version STA+, whose input also consists of 3 parts, which are the output of the STA net with different input data.

In the STA+ net, we shall only additionally consider both short-term and long-term information, where the cross-term information has been omitted to ensure its versatility.
Similar to the classification nets mentioned before, we continue using the proposed C-GCN as the blender.
Thus, we independently train 2 STA+ nets, where the input data (3 parts) are \{1 target + 2 short\} or \{1 target + 2 long\}, and the dataflow of the fusion part in these 2 STA+ nets (see Fig.~\ref{fig:VFPNetworks}-H) can be expressed uniformly as the following equation:
\begin{equation}
\label{eq:staadd}
\begin{aligned}
&sta+_i^{\rm Short/Long}=\Big\{h_{\rm TA}^{m},h_{\rm MG_1}^{m},h_{\rm MG_2}^{m}\Big\}\\[-0.5ex]
&\ \ \ \ \ \ \ \ \ \ \ \ \ =\Omega\Big[sta_i,\ sta_i^{\rm Short}(1),\ sta_i^{\rm Short}(2)\Big]\\[-0.5ex]
&\ \ \ \ \ \ \ \ \ \ \ \ \ \ \ \ \ \ \ \ \ \ \ \ or \ \ \Omega\Big[sta_i,\ sta_i^{\rm Long}(1),\ sta_i^{\rm Long}(2)\Big],
\end{aligned}
\end{equation}
where $sta+_i^{\rm Short/Long}$ is the output of the $m$ step reasoning, which includes 3 tensors (\textit{i.e.}, $h_{\rm TA}^{m},h_{\rm MG_1}^{m},h_{\rm MG_2}^{\rm m}$) with size $28\times 32\times 32$, and these tensors correlate to 3 different nodes (\textit{i.e.}, 1 TA node and 2 MG nodes, respectively, see Eq.~\ref{eq:updateTA} and Eq.~\ref{eq:updateMG}); $\Omega[\cdot]$ denotes the proposed C-GCN mentioned in Sec.~\ref{sub:NMP}; $sta_i^{\rm Short}(1)$ and $sta_i^{\rm Short}(2)$ denote 2 individual outputs of the `STA-Branch' using different short-term fragments as input, and, similarly, $sta_i^{\rm Long}(1)$ and $sta_i^{\rm Long}(2)$ are 2 outputs correlated to 2 different long-term input fragments.

The `ReadOut' module takes $sta+_i^{\rm Short/Long}$ as input individually and then outputs 9 fixation maps with size $356\times 356$.
Taking the target node for instance, we abbreviate $sta+_i^{\rm Short/Long}$ as $h_{\rm TA}^{m}$, where $m$ is the total reasoning steps, and the output of the `ReadOut' can be formulated as:
\begin{equation}
\begin{aligned}
\label{eq:hatsta}
\hat{h}_{\rm TA}=RO(h_{\rm TA}^{m})= Ref_1\Big(\uparrow \big(Ref_{32}(h_{\rm TA}^{m}\otimes S)\big)\Big),
\end{aligned}
\end{equation}
where $S$ denotes the spatial feature correlated to the middle frame of the target fragment, \emph{e.g.}, the $s_i$ in Fig.~\ref{fig:VFPNetworks}-D; $RO$ represents the 'ReadOut' operation in Fig.~\ref{fig:VFPNetworks}-H; $Ref(\cdot)_p$ is a refinement operation which uses $1\times 1$ convolution to reduce its input's channel number to $p$; $\uparrow(\cdot)$ upsample its input to $356\times 356$.
Similarly, the output could become $\hat{h}_{\rm MG_1}$ and $\hat{h}_{\rm MG_2}$ by alternately feeding $h_{\rm MG_1}^{m}$ and $h_{\rm MG_2}^{m}$ to the `ReadOut' module of STA+.

We assign an individual loss function for each fixation prediction net, and the overall loss function adopted by these 2 fixation prediction nets (${\rm STA+^{Short}}$, and ${\rm STA+^{\rm Long}}$, \ref{eq:staadd}) can be uniformly expressed as:
\begin{equation}
\label{eq:STA+}
\begin{aligned}
&\mathcal{L}_{\rm STA+} = \mathcal{L}_{\rm BCE}(\hat{h}_{\rm TA}, {\rm pGT}_{\rm TA})+\mathcal{L}_{\rm KL}(\hat{h}_{\rm TA}, {\rm pGT}_{\rm TA})\\
&\ \ +\mathcal{L}_{\rm BCE}(\hat{h}_{\rm MG_1}, {\rm pGT}_{\rm MG_1})+\mathcal{L}_{\rm KL}(\hat{h}_{\rm MG_1}, {\rm pGT}_{\rm MG_1})\\
&\ \ +\mathcal{L}_{\rm BCE}(\hat{h}_{\rm MG_2}, {\rm pGT}_{\rm MG_2})+\mathcal{L}_{\rm KL}(\hat{h}_{\rm MG_2}, {\rm pGT}_{\rm MG_2}),
\end{aligned}
\end{equation}
where ${\rm pGT}_{\rm TA/MG_1/MG_2}$ denotes the correlated pseudofixation maps of different nodes; the input $\hat{h}_{\rm TA}$ can be obtained via Eq.~\ref{eq:hatsta} directly, and the computation of $\hat{h}_{\rm MG_1/MG_2}$ shall replace Eq.~\ref{eq:hatsta}'s input to $h_{\rm MG_1}^{m}$ and $h_{\rm MG_2}^{m}$ accordingly.


\vspace{-0.2cm}
\subsection{Predicted Visual-Audio Fixations}
\label{sec:PVFFina}
Thus far, we can obtain 3 individual fixation prediction networks mentioned in Sec.~\ref{sec:swfpn} and Sec.~\ref{sec:mgfpn}, \textit{i.e.}, from the sourcewise perspective, we can obtain an STA net, and from the multi-granularity aspect, we can additionally obtain 2 STA+ nets, which correlate to short-term or long-term versions\footnote{We have omitted the cross-term version to ensure a good versatility during the testing phase.}, respectively (Eq.~\ref{eq:staadd}).
These 3 fixation prediction nets complement each other in essence, and we shall combine their predictions to derive the final fixation maps that could outperform each of them.

We denote the predicted fixation map of the 3 prediction nets (${\rm STA}$, ${\rm STA+^{Short}}$, and ${\rm STA+^{\rm Long}}$) as ${\rm PF}_{sta}$, ${\rm PF}_{sta+}^{\rm Short}$, and ${\rm PF}_{sta+}^{\rm Long}$.
The final predicted visual-audio fixation map (${\rm PF}_{final}$) can be formulated as follows:
\begin{equation}
\begin{aligned}
\label{eq:finalfp}
{\rm PF}_{final} =\ &\frac{1}{2}\times\mathcal{Z}\Big(\mathcal{Z}({\rm PF}_{sta})\odot \mathcal{Z}({\rm PF}_{sta+}^{\rm Short})\odot \mathcal{Z}({\rm PF}_{sta+}^{\rm Long})\Big)\\[-0.5ex]
&+\frac{1}{2}\times\mathcal{Z}\Big({\rm PF}_{sta}+{\rm PF}_{sta+}^{\rm Short}+{\rm PF}_{sta+}^{\rm Long}\Big),
\end{aligned}
\end{equation}
where $\mathcal{Z}(\cdot)$ is the typical \emph{min-max} normalization function,
and $\times$ represents the standard multiplicative operation.
and $\odot$ denotes the element multiplicative operation.
Eq.~\ref{eq:finalfp} mainly consists of 2 parts, where the left part obtains the common consistency, while the right part serves as the complementary fixation map to ensure good robustness.

Compared with the conventional plain fusion schemes (\emph{e.g.}, the addition/average/maxium-based fusion), the proposed fusion scheme could make the predicted fixation map (\textit{i.e.}, ${\rm PF}_{final}$) more consistent with the real human fixations.

\begin{table*}[!t]
  \centering
  \caption{Quantitative evidence towards the component studies for STA and STA+,
  in order to facilitate the display, we use different colors to distinguish different components.
  This experiment is conducted on the AVAD set~\cite{min2016fixation}.
    \vspace{-0.4cm}}
    \begin{tabular}{c}
    \begin{minipage}{1\textwidth}
          \hspace{-0.3cm}\includegraphics[width=\linewidth]{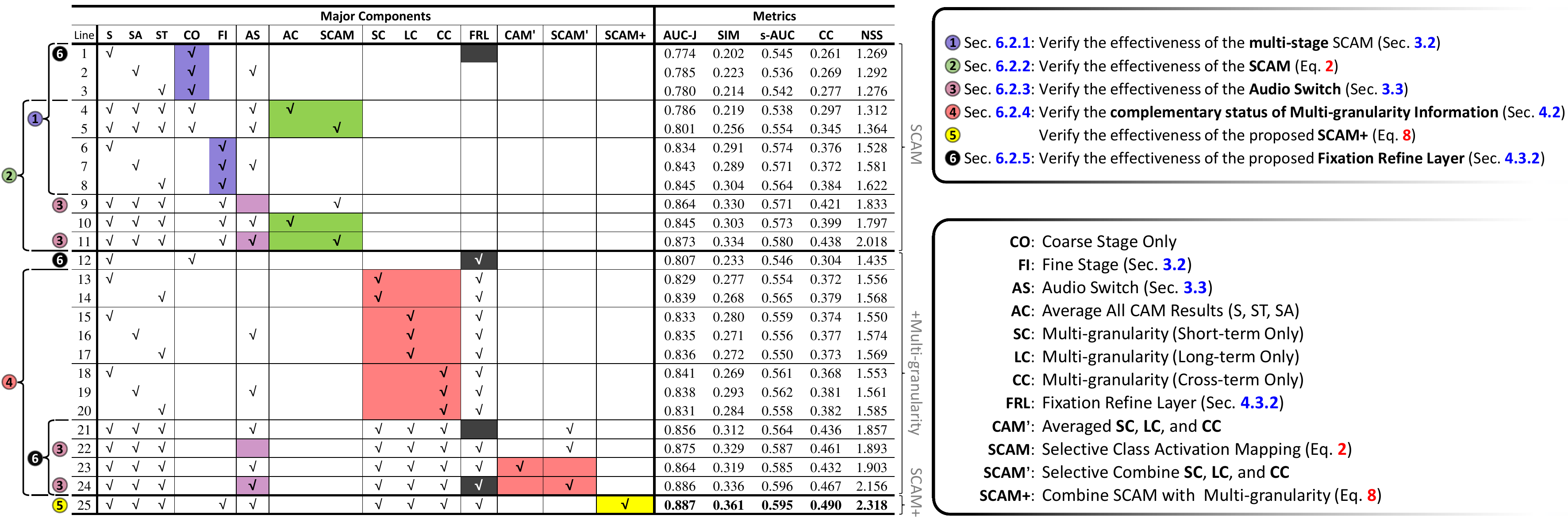}
    \end{minipage}
    \end{tabular}
\label{tab:SCAM+}
\vspace{-0.5cm}
\end{table*}

\vspace{-0.3cm}
\section{Experiments and Validations}
\label{sec:EXP}
\subsection{Implementation Details}
\noindent\textbf{Training Dataset}.
Recently, Google released the Audioset~\cite{gemmeke2017audio},
the largest visual-audio set thus far,
and we use its subset \textbf{a}udio \textbf{v}isual \textbf{e}vent (AVE)
~\cite{tian2018audio}\footnote{ https://sites.google.com/view/audiovisualresearch} location dataset,
which contains 4,143 sequences covering 28 semantic categories,
as the classification training set for the S, ST, SA, S+, ST+ and SA+ classification nets (Sec.~\ref{sec:SCAM} and Sec.~\ref{sec:SCAM+}).

\vspace{0.2cm}
\noindent\textbf{Training Processes}.
For the S, ST, SA, S+, ST+ and SA+ training,
owing to the limitation of computer memory,
we train these branches of the network separately
and use SCAM and SCAM+ to fuse them into the final fake GTs.
For SCAM training,
we follow the widely used multistage training scheme.
In the coarse stage,
where the batch size equals 20 and all video frames are resized to 256$\times$256.
Taking the cropped video patches as input,
three completely new classification nets in the fine stage will be trained,
where the batch size equals 3 and all video patches are resized to 356$\times$356.
For SCAM+ training,
we set the batch size to 16, and all video frames are resized to 256$\times$256, and the S+, ST+ and SA+ multi-granularity reasoning networks used the same experimental setup.
All the classification nets are trained on the AVE dataset.
The STA and STA+ fixation prediction network take the pseudofixations as GTs,
where video frames are resized to 356$\times$356;
thus, the batch size is set to 3.
The STA and STA+ training processes adopted the \textbf{s}tochastic \textbf{g}radient \textbf{d}escent (SGD) ~\cite{bottou2008advan}
optimizer with a learning rate of 0.00005.

\vspace{0.2cm}
\noindent\textbf{Testing Data Sets}.
To test the performance of the proposed approach, we adopted 6 testing datasets, including \textbf{AVAD} ~\cite{min2016fixation}, \textbf{Coutrot1}~\cite{coutrot2014saliency}, \textbf{Coutrot2} ~\cite{coutrot2016multimodal}, \textbf{DIEM} ~\cite{mital2011clustering}, \textbf{SumMe}~\cite{gygli2014creating}, and \textbf{ETMD}~\cite{koutras2015perceptually}. All these sets (241 sequences) are furnished with pixelwise real fixations
that are collected in the visual-audio circumstance.

\vspace{0.2cm}
\noindent\textbf{Quantitative Metrics}.
Following the previous studies on saliency metrics~\cite{wang2018revisiting,borji2012state,borji2019saliency,bylinskii2018different},
we have adopted 5 commonly used evaluation metrics to measure the agreement
between the model-predicted saliency (PS) map and the real human eye movements (CF, FL),
the ranges of predicted saliency map ${\rm PS}\in[0,1]$, continuous fixation map ${\rm CF}\in[0,1]$ and the map of fixation locations ${\rm FL}\in\{0, 1\}$.
It mainly includes location-based metrics \textbf{AUC-J}, \textbf{s-AUC}, \textbf{NSS}, and continuous distribution-based metrics \textbf{SIM}, \textbf{CC}.

\vspace{-0.3cm}
\subsection{Effectiveness Evaluation on Different Components}
\label{subsec:COM}
\subsubsection{Effectiveness of the proposed Multistage Rationale}
\label{subsubsec:COFI}
In Sec.~\ref{sub:MStage}, we adopted the coarse-to-fine rationale to perform SCAM twice, where the primary objective is to mimic the real human attention mechanism --- a multistage mechanism.
To verify the effectiveness of the proposed coarse-to-fine strategy, we tested all CAM results derived from different sources (\textit{i.e.}, S, SA, ST) in different stages, \textit{i.e.}, COARSE vs. and FINE.

The corresponding quantitative evaluation can be found in Table~\ref{tab:SCAM+}, denoted by mark {\large{\textcircled{\small{1}}}}. We have measured the consistency degree between CAM results using different sources and real human-eye fixation data, where the difference between using the FINE stage and without using it is significant.
In cases of using a single source (S, SA, or ST, see lines 1-3 and lines 6-8), the CAM result can get boosted for about 40\%, \emph{e.g.}, the CC metric value has been increased from 2.61 to 3.76 in the case using the spatial source (S) solely.
A similar tendency can also be observed in cases of using multiple sources, \emph{e.g.}, by comparing line 4 with line 10, all metrics can obtain significant improvements, showing that the simply fused multisource CAM results also improve via the proposed multistage strategy.
Furthermore, from the perspective of the proposed SCAM, we can also notice that the multistage strategy can bring solid performance gain, \emph{e.g.}, line 5 vs. line 11, which indicates its generic nature.

\subsubsection{Effectiveness of the Proposed Selective Fusion}
\label{subsubsec:SCAM}
To selectively fuse the multisource CAM results (\textit{i.e.}, $\rm \Phi_S$, $\rm \Phi_{SA}$, and $\rm \Phi_{ST}$, Eq.~\ref{eq:FU}),
we have adopted the classification confidences as the fusion weights ($\rm C_{\rm S}$, $\rm C_{\rm SA}$, and $\rm C_{\rm ST}$).
Actually, the effectiveness of this implementation is based on the precondition
that the classification confidences are indeed positively related to the consistency level between CAM results and real fixations.
To verify this issue, we have compared the proposed selective fusion (\textit{i.e.}, the proposed SCAM, Eq.~\ref{eq:FU}) and the `conventional fusion scheme' --- we average all sourcewise CAM results, which can be formulated as the following equation.
\begin{equation}
{\rm AC} = \frac{1}{3}\big(\Phi_{\rm S} + \Phi_{\rm SA} + \Phi_{\rm ST}\big),
\end{equation}
where all symbols have the same definitions as that of Eq.~\ref{eq:FU}, and we show the corresponding quantitative result of this scheme in the `AC' column of Table~\ref{tab:SCAM+}.

The mark {\large{\textcircled{\small{2}}}} in Table~\ref{tab:SCAM+} has highlighted the advantage of the proposed SCAM against the conventional scheme, \textit{i.e.}, SCAM vs. AC, where the overall quantitative performance can obtain persistent improvement for all considered metrics.
Meanwhile, compared with other single source CAM results (\textit{i.e.}, S, ST, and SA demonstrated in Table~\ref{tab:SCAM+}), we can easily observe that considering all sources simultaneously
cannot take full advantage of the complementary nature between them; thus, the performance improvement achieved by AC is marginal, further showing the performance superiority and effectiveness of the proposed SCAM.

In addition, to further verify the effectiveness of the divide-and-conquer scheme adopted by the proposed selective fusion, we compare the SCAM with the CAM result directly obtained from a three-stream spatial-temporal-audio classification net.
We use `$\rm CAM_{\rm sta}$' to represent the CAM result obtained from this three-stream network, where we simply adopt a three-stream network adopting an identical fusion unit to the proposed STA fixation prediction network (Fig.~\ref{fig:VFPNetworks}-D), while the decoder part has been replaced by a multiclass classifier that is identical to other classification nets.
The quantitative comparison result can be seen in Table~\ref{tab:SCAM}, where, for both coarse and fine stages, the proposed SCAM can outperform $\rm CAM_{\rm sta}$.

\begin{table}[!t]
  \centering
  \caption{Quantitative evidence towards the effectiveness of the proposed SCAM.
  This experiment is conducted on the AVAD set~\cite{min2016fixation}.}
  \vspace{-0.3cm}
  \resizebox{0.9\linewidth}{!}{
    \begin{tabular}{c|l||ccccc}
    \specialrule{1.0pt}{0pt}{0pt}
    & \multicolumn{1}{l||}{Module} & AUC-J $\uparrow$ & SIM$\uparrow$   & s-AUC$\uparrow$ & CC$\uparrow$    & NSS$\uparrow$   \\
    \hline
    \hline
    \multirow{2}[2]{*}{COARSE}
          & CAM$_{\rm sta}$ & 0.793 & 0.227 & 0.551 & 0.293 & 1.273 \\
          & SCAM  & 0.801 & 0.256 & 0.554 & 0.345 & 1.364 \\
    \specialrule{1.0pt}{0pt}{0pt}
    \multirow{2}[2]{*}{FINE}
          & CAM$_{\rm sta}$ & 0.856 & 0.296 & 0.579 & 0.415 & 1.803 \\
          & SCAM  & \textbf{0.873} & \textbf{0.334} & \textbf{0.580} & \textbf{0.438} & \textbf{2.018} \\
     \specialrule{1.0pt}{0pt}{0pt}
    \end{tabular}%
    }
  \label{tab:SCAM}%
  \vspace{-0.5cm}
\end{table}

\subsubsection{The Effectiveness of the Proposed Audio Switch}
\label{subsubsec:AS}

In Table~\ref{tab:SCAM+}, we have reported the performance of the proposed model without using the proposed \textbf{a}udio \textbf{s}witch (Sec.~\ref{sec:FMAIS}), where the corresponding quantitative results have been highlighted by mark {\large{\textcircled{\small{3}}}}.

By comparing line 9 with line 11, and line 22 with line 24, we can easily observe that the proposed audio switch is capable of improving the overall performance by approximately 1.5\% on average.
The main reason is that it can filter meaningless background audio,
alleviating the learning ambiguity when fusing unsynchronized spatial and audio information.

In addition, since the proposed audio switch is a major component of the proposed `SA Fuse' module (Sec.~\ref{sec:FMAIS}), we shall additionally verify whether the SA Fused module can outperform the conventional fusion scheme.
Thus, we have conducted an additional quantitative evaluation, where, in Table~\ref{tab:SAFuseQuan}, we have compared the proposed SA Fuse with the existing plain ones, including addition, multiplicative, concatenation, and bilinear.
As expected, the proposed SA Fuse module significantly outperforms all other approaches.

\subsubsection{Effectiveness of Multi-granularity Selective Fusion}
\label{subsubsec:SLC}
To study the influence of the proposed multi-granularity information on the proposed SCAM,
we tested all CAMs derived from different sources (\textit{i.e.}, S, SA, ST) coupled with multi-granularity information, \textit{i.e.}, short-term (SC), long-term (LC) and cross-term (CC) information, respectively.

\begin{table}[!t]
  \centering
  \caption{Quantitative evidence towards the effectiveness of the proposed `SA Fuse' module,
  `Addition, Multiplication, Concatenation, Bilinear' is the visual and audio info fused mode,
  `SA Fuse' is our proposed module of visual-audio fusion.
  This experiment is conducted on the AVAD set~\cite{min2016fixation}.}
  \vspace{-0.3cm}
  \resizebox{0.85\linewidth}{!}{
    \begin{tabular}{l||ccccc}
    \specialrule{1.0pt}{0pt}{0pt}
          & AUC-J$\uparrow$ & SIM$\uparrow$   & s-AUC$\uparrow$ & CC$\uparrow$    & NSS$\uparrow$ \\
    \hline
    \hline
    Addition   & 0.857  & 0.325  & 0.581  & 0.438  & 1.880  \\
    Multiplicative   & 0.860  & 0.327  & 0.575  & 0.442  & 1.915  \\
    Concatenation   & 0.873  & 0.331  & 0.585  & 0.459  & 1.944  \\
    Bilinear & 0.864  & 0.329  & 0.577  & 0.428  & 1.825  \\
    \hline
    SA~Fuse & \textbf{0.886}  & \textbf{0.336}  & \textbf{0.596}  & \textbf{0.467}  & \textbf{2.156}  \\
    \specialrule{1.0pt}{0pt}{0pt}
    \end{tabular}%
    }
  \label{tab:SAFuseQuan}%
  \vspace{-0.5cm}
\end{table}%

As seen in Table~\ref{tab:SCAM+}, marked by {\large{\textcircled{\small{4}}}}, we found the performance gain when using each type of multi-granularity information with the sourcewise CAM results.
Meanwhile, we have also tested using all types of multi-granularity information simultaneously, where we use CAM' and SCAM' to denote the averaged and selectively fused SC, LC, and CC, respectively.

Clearly, compared with the CAM result obtained via a single source (line 12), the multi-granularity information (line 13) could bring solid performance gain.
However, simply combining the single-source-based CAM result with one type of multi-granularity information might still be inferior to the SCAM, which can be confirmed by comparing line 11 and line 13.
The main reason is that, compared with the sourcewise information, the multi-granularity information could be less important to the classification task, motivating us to use it to serve the SCAM as the subordinates.

As expected, since CAM' (line 23) has adopted multiple multi-granularity information, the overall performance could be improved compared with the single multi-granularity information-based ones (\emph{e.g.}, lines 13-20).
In addition, benefiting from selective fusion, the SCAM' (lines 21, 22, and 24) could further improve the overall performance.
After combining full sourcewise information and full multi-granularity information, the proposed SCAM+, marked by {\large{\textcircled{\small{5}}}}, achieves the best performance.

\subsubsection{Effectiveness of Fixation Refine Layer}
\label{subsubsec:FRL}
To verify the effectiveness of the proposed \textbf{f}ixation \textbf{r}efine \textbf{l}ayer (FRL), we conducted a series of evaluations in Table~\ref{tab:SCAM+}, and the corresponding results are marked by {\large{\textcircled{\small{6}}}}.

First, as seen in line 12, we applied the FRL to the spatial source-based coarse stage feature, where the FRL is directly applied to ${\rm S}_i$ (Fig.~\ref{fig:Networks}-A), then generated the FRL-based CAM result. Comparing line 12 with line 1, a clear performance improvement can be observed easily.

Second, we have also tried to remove the FRL from the SCAM' to verify the performance gap.
By comparing line 21 with line 24, we can easily notice a clear performance decrease, \emph{e.g.}, the AUC-J metric has dropped from 0.886 to 0.856, where the only difference between these two results only depends on whether the FRL is used.

\vspace{-0.3cm}
\subsection{Ablation Study}
\subsubsection{Thresholds Used in FRL}
\label{subsubsec:THFRL}


\begin{table}[!t]\large
  \centering
  \caption{Quantitative evidence towards the effectiveness of the FRL (frame refine layer), the $\mathcal{T}_{d}$ is the threshold for choosing the refine operation or not, the $\mathcal{T}_{r}$ is a suppressor for refining.\vspace{-0.3cm}}
  \resizebox{1\linewidth}{!}{
  \begin{tabular}{r|c||ccc|ccc|ccc}
  \specialrule{1.5pt}{0pt}{0pt}
  \multicolumn{2}{c}{Dataset} & \multicolumn{3}{c|}{AVAD~\cite{min2016fixation}} & \multicolumn{3}{c|}{DIEM~\cite{mital2011clustering}} & \multicolumn{3}{c}{SumMe~\cite{gygli2014creating}} \\
  \hline
  $\mathcal{T}_{d}$     & $\mathcal{T}_{r}$     & AUC-J$\uparrow$ & CC$\uparrow$     & NSS$\uparrow$   & AUC-J$\uparrow$ & CC$\uparrow$     & NSS$\uparrow$   & AUC-J$\uparrow$ & CC$\uparrow$     & NSS$\uparrow$ \\
  \hline
  \hline
  \multicolumn{2}{c||}{\emph{w/o.} FRL} & 0.856  & 0.436  & 1.857  & 0.835  & 0.482  & 1.799  & 0.844  & 0.330  & 1.598  \\
  \hline
  0.8   & 0.4 & 0.860  & 0.426  & 1.671  & 0.859  & 0.504  & 1.832  & 0.849  & 0.335  & 1.602  \\
  0.8   & 0.8  & 0.843  & 0.393  & 1.802  & 0.844  & 0.511  & 1.815  & 0.853  & 0.341  & 1.609  \\
  \hline
  0.9   & 0.6 & 0.868  & 0.421  & 1.790  & 0.865  & 0.498  & 1.856  & 0.848  & 0.332  & 1.611  \\
  0.7   & 0.6  & 0.849  & 0.425  & 1.699  & 0.853  & 0.487  & 1.862  & 0.841  & 0.340  & 1.600  \\
  \hline
  \textbf{0.8} & \textbf{0.6} & \textbf{0.886} & \textbf{0.467} & \textbf{2.156} & \textbf{0.877} & \textbf{0.518} & \textbf{1.955} & \textbf{0.860} & \textbf{0.367} & \textbf{1.633} \\
  \hline
  \multicolumn{2}{c}{Dataset} & \multicolumn{3}{c|}{ETMD~\cite{koutras2015perceptually}} & \multicolumn{3}{c|}{Coutrotl~\cite{coutrot2014saliency} } & \multicolumn{3}{c}{Coutrot2~\cite{coutrot2016multimodal}} \\
  \hline
  $\mathcal{T}_{d}$     & $\mathcal{T}_{r}$     & AUC-J$\uparrow$ & CC$\uparrow$     & NSS$\uparrow$   & AUC-J$\uparrow$ & CC$\uparrow$     & NSS$\uparrow$   & AUC-J$\uparrow$ & CC$\uparrow$     & NSS$\uparrow$ \\
  \hline
  \hline
  \multicolumn{2}{c||}{\emph{w/o.} FRL} & 0.877  & 0.402  & 1.891  & 0.815  & 0.301  & 1.302  & 0.832  & 0.343  & 1.988  \\
  \hline
  0.8   & 0.4 & 0.882  & 0.409  & 1.915  & 0.812  & 0.303  & 1.309  & 0.829  & 0.329  & 1.990  \\
  0.8   & 0.8  & 0.881  & 0.411  & 1.923  & 0.824  & 0.308  & 1.311  & 0.826  & 0.340  & 1.984  \\
  \hline
  0.9   & 0.6 & 0.875  & 0.399  & 1.921  & 0.829  & 0.309  & 1.309  & 0.834  & 0.337  & 1.988  \\
  0.7   & 0.6  & 0.901  & 0.420  & 1.952  & 0.825  & 0.312  & 1.315  & 0.840  & 0.343  & 1.992  \\
  \hline
  \textbf{0.8} & \textbf{0.6} & \textbf{0.903} & \textbf{0.423} & \textbf{2.056} & \textbf{0.833} & \textbf{0.330} & \textbf{1.326} & \textbf{0.858} & \textbf{0.371} & \textbf{2.105} \\
  \specialrule{1.5pt}{0pt}{0pt}
\end{tabular}%
}
\label{tab:FRL}%
\vspace{-0.5cm}
\end{table}%

There exist 2 predefined thresholds in the proposed FRL, \textit{i.e.}, the $\mathcal{T}_{d}$ and $\mathcal{T}_{r}$ in Eq.~\ref{eq:rmask} and Eq.~\ref{eq:updatehh}, which could directly influence the overall performance, and we will conduct an ablation study on the exact choices.
We retrain the STA+ net multiple times using different $\{\mathcal{T}_{d}, \mathcal{T}_{r}\}$ combinations, and the corresponding results can be found in Table ~\ref{tab:FRL}, where the 1st row denotes the result without using the FRL.

Since 2 thresholds should be determined, we shall adopt the one-fixed-another-solved strategy, in which we empirically assign $\mathcal{T}_{d}=0.8$ and $\mathcal{T}_{r}=0.6$ as our initial choice, and we have tested several other choices.
Coincidentally, the initial choice turns to be the best, and the main reason could be the fact that the technical details of other major components are all implemented based on this initial choice, which makes it passively to outperform all other choices.

\begin{figure}[!b]
  \centering
  \vspace{-0.7cm}
  \includegraphics[width=1\linewidth]{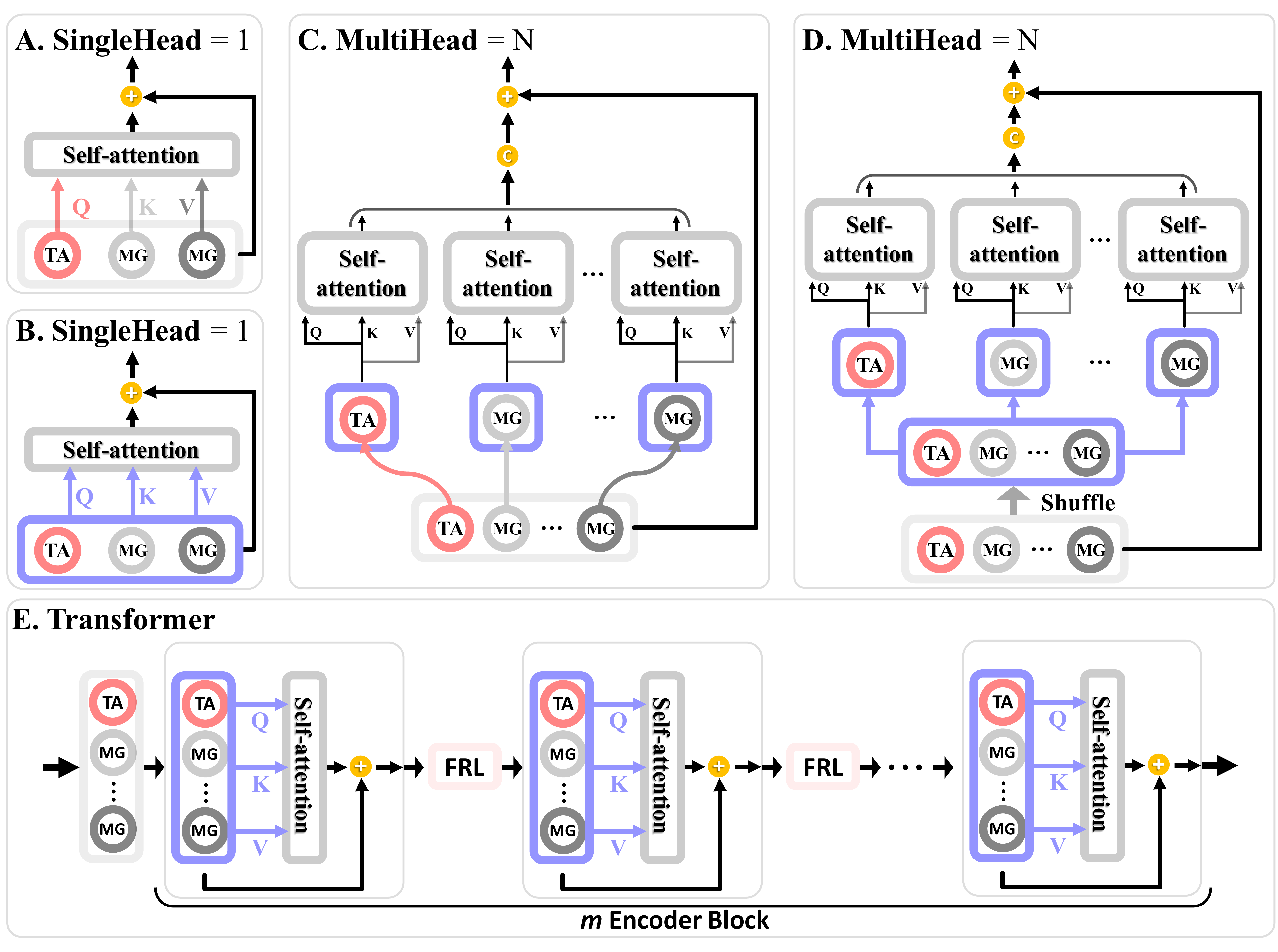}
  \vspace{-0.8cm}
  \caption{The detailed data flow process of transformer adopts SingleHead=1 and MultiHead=N as input for relation modeling.}
  \label{fig:MultiHead}
\end{figure}

\vspace{-0.1cm}
\subsubsection{Reasoning Steps and Node Number}

To combine the multi-granularity information with the proposed SCAM, we have devised the reasoning unit (Sec.~\ref{sub:NMP}), where there exist 2 key parameters that could influence the overall performance, \textit{i.e.}, the exact reasoning steps (\emph{m}) and the \textbf{m}ulti\textbf{g}ranularity (MG) node number ($n$).

Similar to the previous experimental settings, we test various versions of STA+ trained on data using SCAM+ as the pseudo GTs, where we set the ranges of $m$ and $n$ as $\{2,3,4,5\}$ and $\{1,2,3,4\}$,
respectively. The corresponding quantitative results can be found in Table~\ref{tab:VAGNN}.

As seen in this table, we can found that `$w/o$ Reasoning' indicates combining both the target node and other MG nodes simply via feature concatenation. Since these concatenated features are basically not aligned well, this plain implementation has demonstrated the worst result.
Then, by performing the reasoning 3 times, our approach has reached its best performance, and if we continue the reasoning steps more than 3 times, the overall performance could degenerate significantly.
The main reason is that the reasoning step number usually determines the network's learning capacity; with a given training set with limited size, there shall be a balance between capacity and data size, and $m=3$ is the break point.

\textit{w.r.t.} the exact MG node number, we can easily notice that the performance increases with more MG nodes (1 $\rightarrow$2),
while the overall performance could decrease if we use too many MG nodes (3 $\rightarrow$4).
Actually, with the increasing number of MG nodes, the feature space of the nodes' relationships could become more complex, making the learning task quite difficult.
Though the MG nodes could make the process of computing SCAM+ more consistent with the real human attention mechanism, we shall stay conservative towards the exact choice of $n$, and thus we choose $n=2$ as the optimal choice.

\begin{table}[!t]\Large
  \centering
  \caption{Ablation study on semantic reasoning stage number ($m$, Sec.~\ref{sub:NMP}) and MG node number ($n$, Sec.~\ref{sub:NMP}) on the AVAD set~\cite{min2016fixation}. \emph{w/o} Rea.: without Reasoning. We tested multiple variants whose GCN have been replaced by either RNNs or Transformers.}
  \vspace{-0.3cm}
  \resizebox{1\linewidth}{!}{
    \begin{tabular}{c|cc||ccccc}
    \specialrule{1.5pt}{0pt}{0pt}
    \multicolumn{3}{l||}{} & AUC-J$\uparrow$ & SIM$\uparrow$   & s-AUC$\uparrow$ & CC$\uparrow$    & NSS$\uparrow$ \\
    \hline
    \hline
    \multirow{9}[6]{*}{GCN}
    &\multicolumn{2}{c||}{\emph{w/o} Rea.} & 0.825  & 0.253  & 0.556  & 0.338  & 1.519  \\
    \cline{2-8}
    & \multicolumn{1}{c|}
    {\multirow{4}[2]{*}{Reasoning Stage=m}} & $m$=2     & 0.873  & 0.316  & 0.575  & 0.442  & 1.980  \\
    & \multicolumn{1}{c|}{} & $m$=3 &  \textbf{0.886}  & \textbf{0.336}  & \textbf{0.596}  & \textbf{0.467}  & \textbf{2.156}  \\
    & \multicolumn{1}{c|}{} & $m$=4     & 0.872  & 0.324  & 0.580  & 0.451  & 1.981  \\
    & \multicolumn{1}{c|}{} & $m$=5     & 0.870  & 0.309  & 0.576  & 0.439  & 1.967  \\
    \cline{2-8}
    & \multicolumn{1}{c|}
    {\multirow{4}[2]{*}{TA=1, MG=$n$}} & $n$=1     & 0.871  & 0.319  & 0.580  & 0.432  & 1.965  \\
    & \multicolumn{1}{c|}{} & $n$=2 & \textbf{0.886}  & \textbf{0.336}  & \textbf{0.596}  & \textbf{0.467}  & \textbf{2.156}  \\
    & \multicolumn{1}{c|}{} & $n$=3     & 0.867  & 0.306  & 0.578  & 0.424  & 1.946  \\
    & \multicolumn{1}{c|}{} & $n$=4     & 0.852  & 0.289  & 0.565  & 0.415  & 1.863  \\
    \hline
    \multicolumn{3}{r||}{RNNs}  & 0.875  & 0.319  & 0.562  & 0.441  & 1.983  \\
    \hline
    \multirow{4}[2]{*}{Transformer}
    & \multicolumn{2}{l||}{\textbf{A}-SingleHead Realization} & 0.848 & 0.306 & 0.548 & 0.425 & 1.951 \\
    & \multicolumn{2}{l||}{\textbf{B}-{\tiny\ }SingleHead Realization} & \textbf{0.882} & \textbf{0.339} & \textbf{0.599} & \textbf{0.464} & \textbf{2.157} \\
    & \multicolumn{2}{l||}{\textbf{C}-{\tiny\ \ \ }MultiHead Realization} & 0.872  & 0.328  & 0.573  & 0.440  & 1.975  \\
    & \multicolumn{2}{l||}{\textbf{D}-{\tiny\ \ \ }MultiHead Realization} & 0.861  & 0.320  & 0.559  & 0.438  & 1.973  \\
    \hline
    \multirow{4}[2]{*}{Version-\textbf{B}}
    & \multicolumn{2}{l||}{\emph{with} 2 Encoder Blocks} & 0.875  & 0.335  & 0.581  & 0.460  & 2.155  \\
    & \multicolumn{2}{l||}{\emph{with} 3 Encoder Blocks} & \textbf{0.882} & \textbf{0.339} & \textbf{0.599} & \textbf{0.464} & \textbf{2.157} \\
    & \multicolumn{2}{l||}{\emph{with} 4 Encoder Blocks} & 0.876  & 0.330  & 0.584  & 0.453  & 2.164  \\
    & \multicolumn{2}{l||}{\emph{with} 5 Encoder Blocks} & 0.874  & 0.327  & 0.576  & 0.439  & 1.996  \\
   \hline\hline
    \multicolumn{3}{r||}{Modules}   & \multicolumn{3}{r}{FLOPs$\downarrow$} & \multicolumn{2}{r}{Params$\downarrow$}\\
    \hline
    \hline
    \multicolumn{3}{r||}{RNNs}      & \multicolumn{3}{r}{347,504,640.0}  & \multicolumn{2}{r}{113,120.0}\\
    \hline
    \multicolumn{3}{r||}{(\textbf{B})SingleHead Realization 2} & \multicolumn{3}{r}{1,071,673,344.0}  & \multicolumn{2}{r}{1,043,376.0}\\
    \hline
    \multicolumn{3}{r||}{\textbf{C-GCN~(TA=1, MGNodes=2)}} & \multicolumn{3}{r}{189,923,328.0}  & \multicolumn{2}{r}{19,768.0} \\
    \specialrule{1.5pt}{0pt}{0pt}
    \end{tabular}%
    }
  \label{tab:VAGNN}%
  \vspace{-0.5cm}
\end{table}%

\begin{table}[!t]\large
  \centering
  \caption{Quantitative comparison of different fusion schemes for generating generic fixation maps on 6 testing datasets.}
    \vspace{-0.3cm}
  \resizebox{1\linewidth}{!}{
  \begin{tabular}{l||ccc|ccc|ccc}
  \specialrule{1.5pt}{0pt}{0pt}
  & \multicolumn{3}{c|}{AVAD~\cite{min2016fixation}}              & \multicolumn{3}{c|}{DIEM~\cite{mital2011clustering}}      & \multicolumn{3}{c}{SumMe~\cite{gygli2014creating}} \\
  \hline
  & {AUC-J$\uparrow$} & {CC$\uparrow$} & {NSS$\uparrow$} & {AUC-J$\uparrow$} & {CC$\uparrow$} & {NSS$\uparrow$} & {AUC-J$\uparrow$} & {CC$\uparrow$} & {NSS$\uparrow$}\\
  \hline
    PF$_{sta}$   & 0.873  & 0.438  & 2.018  & 0.861  & 0.469  & 1.716  & 0.854  & 0.368  & 1.647 \\
    PF$_{sta+}^{\rm Short}$ & 0.881  & 0.465  & 2.151  & 0.872  & 0.486  & 1.843  & 0.855  & 0.331  & 1.466 \\
    PF$_{sta+}^{\rm Long}$ & 0.886  & 0.469  & 2.159  & 0.874  & 0.501  & 1.896  & 0.860  & 0.383  & 1.650\\
    \hline
    PF$_{agg}$ & 0.886  & \textbf{0.507} &\textbf{2.500} & 0.862  & 0.500  & 1.920  & 0.844  & 0.359  & 1.593\\
    PF$_{final}$ & \textbf{0.887} & 0.490  & 2.318  & \textbf{0.884} &\textbf{0.544} &\textbf{2.042} & \textbf{0.866} &\textbf{0.402} &\textbf{1.761}\\
    \hline
    & \multicolumn{3}{c|}{ETMD~\cite{koutras2015perceptually}} & \multicolumn{3}{c|}{Coutrotl~\cite{coutrot2014saliency}}   & \multicolumn{3}{c}{Coutrot2~\cite{coutrot2016multimodal}} \\
    \hline
     & {AUC-J$\uparrow$} & {CC$\uparrow$} & {NSS$\uparrow$} & {AUC-J$\uparrow$} & {CC$\uparrow$} & {NSS$\uparrow$} & {AUC-J$\uparrow$} & {CC$\uparrow$} & {NSS$\uparrow$} \\
    \hline
    PF$_{sta}$     & 0.908  & 0.448  & 2.176  & 0.829  & 0.339  & 1.376  & 0.850  & 0.273  & 1.475  \\
    PF$_{sta+}^{\rm Short}$   & 0.895  & 0.381  & 1.850  & 0.829  & 0.305  & 1.232  & 0.847  & 0.272  & 1.883  \\
    PF$_{sta+}^{\rm Long}$   & 0.904  & 0.437  & 2.101  & 0.832  & 0.327  & 1.355  & 0.854  & 0.332  & 1.920  \\
    \hline
    PF$_{agg}$   & 0.898  & 0.434  & 2.192  & 0.814  & 0.314  & 1.270  & 0.803  & 0.341  & \textbf{1.952} \\
    PF$_{final}$  &\textbf{0.910} &\textbf{0.452} &\textbf{2.297} & \textbf{0.840} &\textbf{0.354}  & \textbf{1.429} &\textbf{0.862} &\textbf{0.349}  & \textbf{1.952}  \\
    \specialrule{1.5pt}{0pt}{0pt}
  \end{tabular}%
  \label{tab:sta+}%
  }\vspace{-0.6cm}
\end{table}%

\begin{table*}[!t]\Huge
  \centering
    \caption{
    Quantitative comparisons between our method with other fully-/weakly-/un-supervised methods on 6 datasets. \textbf{Bold} means the best result, \textbf{OUR(STANet+$\ddagger$)} denotes the proposed STANet+ trained on the VggSound dataset~\cite{chen2020vggsound}. In addition, we have also provided some representative `\textbf{Qualitative Comparisons}' between our approach and the SOTA works, which can be found in the submitted `\textbf{Supplementary Material}'. Here max J, S, A, C, and N indicate maximal AUC-J, SIM, s-AUC, CC, and NSS respectively.}
    \vspace{-0.3cm}
    \resizebox{1\textwidth}{!}{
      \hspace{-0.2cm}\begin{tabular}{r||ccccc|ccccc|ccccc|ccccc|ccccc|ccccc}
      \specialrule{3.5pt}{0pt}{0pt}
      Dataset & \multicolumn{5}{c|}{AVAD~\cite{min2016fixation}}    & \multicolumn{5}{c|}{DIEM~\cite{mital2011clustering}}       & \multicolumn{5}{c|}{SumMe~\cite{gygli2014creating}} & \multicolumn{5}{c|}{ETMD~\cite{koutras2015perceptually}}    & \multicolumn{5}{c|}{Coutrotl~\cite{coutrot2014saliency}} & \multicolumn{5}{c}{Coutrot2~\cite{coutrot2016multimodal}} \\
      \hline
      Methods & J$\uparrow$ & S$\uparrow$ & A$\uparrow$ & C$\uparrow$ & N$\uparrow$ & J$\uparrow$ & S$\uparrow$ & A$\uparrow$ & C$\uparrow$ & N$\uparrow$ & J$\uparrow$ & S$\uparrow$ & A$\uparrow$ & C$\uparrow$ & N$\uparrow$ & J$\uparrow$ & S$\uparrow$ & A$\uparrow$ & C$\uparrow$ & N$\uparrow$ & J$\uparrow$ & S$\uparrow$ & A$\uparrow$ & C$\uparrow$ & N$\uparrow$ & J$\uparrow$ & S$\uparrow$ & A$\uparrow$ & C$\uparrow$ & N \\
      \hline
      \hline
      ITTI~\cite{itti1998model}*  & .688  & .170  & .533  & .131  & 0.61  & .663  & .217  & .583  & .137  & 0.56  & .666  & .151  & .559  & .097  & 0.44  & \textbf{.856} & .226  & .613  & \textbf{.299} & \textbf{1.40}  & \textbf{.798} & \textbf{.253} & .526  & \textbf{.272} & \textbf{1.06}  & .819  & \textbf{.189} & .577  & \textbf{.183} & 1.07  \\
      GBVS~\cite{harel2007graph}*  & \textbf{.854} & .247  & .572  & \textbf{.337} & \textbf{1.56}  & \textbf{.830} & \textbf{.318} & .605  & \textbf{.356} & \textbf{1.28}  & \textbf{.808} & .221  & .567  & \textbf{.272} & \textbf{1.13}  & \textbf{.856} & .226  & .613  & \textbf{.299} & \textbf{1.40}  & \textbf{.798} & \textbf{.253} & .526  & \textbf{.272} & \textbf{1.06}  & .819  & \textbf{.189} & .577  & \textbf{.183} & 1.07  \\
      SCLI~\cite{rudoy2013learning}*  & .747  & .210  & .535  & .170  & 0.79  & .739  & .267  & .590  & .207  & 0.78  & .746  & .209  & .577  & .184  & 0.80  & .761  & .165  & .570  & .129  & 0.62  & .754  & .216  & .536  & .239  & 0.88  & .669  & .137  & .510  & .014  & 0.09  \\
      SBF~\cite{zhang2017supervision}*   & .833  & \textbf{.272} & .576  & .308  & 1.49  & .759  & .292  & .608  & .301  & 1.08  & .783  & \textbf{.228} & .590  & .230  & 1.02  & .805  & \textbf{.232} & .641  & .262  & 1.30  & .726  & .187  & .530  & .215  & 0.79  & \textbf{.827} & .152  & .583  & .131  & \textbf{1.10}  \\
      AWS-D~\cite{leboran2016dynamic}* & .825  & .221  & \textbf{.589} & .304  & 1.38  & .733  & .250  & \textbf{.612} & .301  & 1.13  & .747  & .192  & \textbf{.603} & .186  & 0.85  & .754  & .161  & \textbf{.664} & .181  & 0.91  & .729  & .214  & .581  & .207  & 0.87  & .783  & .170  & \textbf{.590} & .146  & 0.84  \\
      \specialrule{1.0pt}{0pt}{0pt}
      CAM~\cite{zhou2016learning}\#& .743  & .195  & .542  & .217  & 0.93  & .730  & .248  & .589  & .284  & 0.84  & .744  & .184  & .570  & .201  & 0.80  & .715  & .145  & .556  & .140  & 0.63  & .680  & .196  & .508  & .151  & 0.55  & .482  & .126  & .425  & .029  & 0.16  \\
      GCAM~\cite{selvaraju2017grad}\#& .743  & .196  & .542  & .217  & 0.93  & .730  & .249  & .589  & .236  & 0.84  & .744  & .185  & .570  & .201  & 0.80  & .715  & .146  & .555  & .141  & 0.63  & .680  & .197  & .508  & .151  & 0.55  & .482  & .126  & .425  & .030  & 0.16  \\
      GCAMpp~\cite{chattopadhay2018grad}\#& .777  & .273  & .559  & .255  & 1.22  & .732  & .216  & .583  & .271  & 0.78  & .774  & .217  & .593  & .225  & 0.92  & .575  & .124  & .157  & .576  & 0.74  & .704  & .137  & .537  & .210  & 0.51  & .733  & .114  & .567  & .168  & 0.63  \\
      SGCAMpp~\cite{omeiza2019smooth}\#& .809  & .206  & .550  & .275  & 1.18  & .802  & .271  & .620  & .319  & 1.12  & .786  & .191  & .591  & .234  & 0.94  & .791  & .162  & .599  & .212  & 0.95  & .750  & .217  & .523  & .220  & 0.84  & .618  & .155  & .449  & .061  & 0.26  \\
      xGCAM~\cite{fu2020axiom}\#& .743  & .196  & .543  & .217  & 0.93  & .730  & .249  & .589  & .236  & 0.84  & .743  & .184  & .570  & .199  & 0.79  & .715  & .146  & .555  & .141  & 0.63  & .680  & .197  & .508  & .151  & 0.55  & .482  & .126  & .425  & .030  & 0.16  \\
      SSCAM~\cite{wang2020ss}\#& .777  & .186  & .531  & .228  & 0.97  & .750  & .242  & .604  & .248  & 0.87  & .763  & .179  & .591  & .206  & 0.83  & .730  & .144  & .595  & .149  & 0.68  & .686  & .194  & .521  & .154  & 0.59  & .502  & .141  & .417  & .003  & 0.02  \\
      ScoCAM~\cite{wang2020score}\#& .772  & .196  & .548  & .237  & 1.02  & .770  & .257  & .614  & .279  & 0.98  & .753  & .182  & .577  & .202  & 0.80  & .737  & .148  & .581  & .157  & 0.70  & .708  & .203  & .518  & .176  & 0.67  & .538  & .143  & .423  & .018  & 0.07  \\
      LCAM~\cite{jiang2021layercam}\#& .776  & .199  & .542  & .241  & 1.03  & .773  & .259  & .616  & .285  & 1.01  & .778  & .189  & .593  & .228  & 0.91  & .749  & .151  & .581  & .168  & 0.75  & .699  & .201  & .516  & .168  & 0.62  & .511  & .141  & .400  & .003  & 0.03  \\
      ISCAM~\cite{naidu2020cam}\#& .774  & .195  & .545  & .240  & 1.03  & .774  & .256  & .619  & .282  & 0.99  & .761  & .183  & .582  & .208  & 0.90  & .738  & .147  & .585  & .157  & 0.71  & .704  & .201  & .520  & .171  & 0.65  & .480  & .135  & .409  & .018  & 0.12  \\
      ACAM~\cite{ramaswamy2020ablation}\#& .759  & .198  & .539  & .231  & 0.98  & .735  & .247  & .607  & .245  & 0.88  & .756  & .183  & .584  & .209  & 0.84  & .722  & .145  & .571  & .147  & 0.66  & .677  & .198  & .516  & .154  & 0.55  & .484  & .130  & .465  & .030  & 0.17  \\
      EGCAM~\cite{muhammad2020eigen}\#& .737  & .222  & .533  & .212  & 0.91  & .758  & .310  & .618  & .308  & 1.10  & .741  & .220  & .583  & .215  & 0.87  & .687  & .156  & .570  & .124  & 0.58  & .640  & .193  & .509  & .114  & 0.41  & .575  & .107  & .425  & .031  & 0.18  \\
      ECAM~\cite{muhammad2020eigen}\#& .725  & .219  & .526  & .205  & 0.88  & .740  & .294  & .605  & .273  & 0.97  & .727  & .211  & .570  & .198  & 0.79  & .683  & .151  & .555  & .116  & 0.54  & .647  & .200  & .508  & .130  & 0.46  & .610  & .104  & .415  & .037  & 0.22  \\
      SPG~\cite{zhang2018self}\#& .662  & .176  & .506  & .165  & 0.73  & .713  & .238  & .579  & .233  & 0.86  & .714  & .182  & .561  & .209  & 0.91  & .695  & .138  & .550  & .144  & 0.69  & .650  & .187  & .505  & .142  & 0.53  & .511  & .123  & .464  & .017  & 0.07  \\
      VUNP~\cite{Li_2021_WACV}\#& .574  & .067  & .500  & .142  & 0.29  & .558  & .047  & .515  & .172  & 0.19  & .555  & .013  & .507  & .114  & 0.05  & .505  & .030  & .103  & .132  & 0.59  & .589  & .063  & .514  & .152  & 0.30  & .661  & .101  & .536  & .162  & 0.49  \\
      WSS~\cite{wang2017learning}\#  & .858  & .292  & .592  & .347  & 1.66  & .803  & .333  & .620  & .344  & 1.29  & .812  & .245  & .589  & .279  & 1.10  & .854  & .277  & .661  & .334  & 1.65  & .772  & .247  & .547  & .233  & 0.98  & .835  & .208  & .578  & .192  & 1.18  \\
      MWS~\cite{zeng2019multi}\#  & .834  & .272  & .573  & .309  & 1.48  & .806  & .336  & .628  & .350  & 1.31  & .808  & .237  & .607  & .258  & 1.16  & .833  & .237  & .649  & .293  & 1.43  & .743  & .231  & .528  & .201  & 0.80  & .839  & .188  & .581  & .168  & 1.20  \\
      WSSA~\cite{zhang2020weakly}\#& .807  & .261  & .574  & .285  & 1.34  & .767  & .305  & .608  & .311  & 1.18  & .755  & .225  & .585  & .231  & 1.06  & .793  & .201  & .622  & .222  & 1.08  & .701  & .180  & .535  & .169  & 0.78  & .797  & .185  & .571  & .180  & 1.26  \\
      STANet~\cite{Wang_2021_CVPR}\#& .873  & .334  & .580  & .438  & 2.02  & .861  & .391  & .658  & .469  & 1.72  & .854  & .294  & .627  & .368  & 1.65  & .908  & .318  & .682  & .448  & 2.18  & .829  & .306  & .542  & .339  & 1.38  & .850  & .247  & .597  & .273  & 1.48  \\
      \textbf{STANet+}\#& .887  & .361  & .595  & .490  & 2.32  & .884  & \textbf{.436} & .679  & .544  & 2.04  & .866  & \textbf{.323} & .634  & .402  & 1.76  & .910  & .328  & .683  & .452  & 2.30  & .840  & .315  & .552  & .354  & 1.43  & .862  & .267  & \textbf{.612} & \textbf{.349} & 1.95  \\
      \textbf{STANet+$\ddagger$}\# & \textbf{.892} & \textbf{.366} & \textbf{.603} & \textbf{.508} & \textbf{2.43}  & \textbf{.887} & .433  & \textbf{.687} & \textbf{.558} & \textbf{2.13}  & \textbf{.868} & .319  & \textbf{.642} & \textbf{.417} & \textbf{1.86}  & \textbf{.915} & \textbf{.345} & \textbf{.699} & \textbf{.493} & \textbf{2.48}  & \textbf{.843} & \textbf{.315} & \textbf{.555} & \textbf{.370} & \textbf{1.54}  & \textbf{.871} & \textbf{.269} & .609  & .348  & \textbf{1.97}  \\
      \specialrule{1.0pt}{0pt}{0pt}
      DeepNet~\cite{pan2016shallow}$\dagger$ & .869  & .256  & .561  & .383  & 1.85  & .832  & .318  & .622  & .407  & 1.52  & .848  & .227  & .645  & .332  & 1.55  & .889  & .225  & .699  & .387  & 1.90  & .824  & .273  & .559  & .346  & 1.41  & .896  & .201  & .600  & .301  & 1.82  \\
      SalGAN~\cite{pan2017salgan}$\dagger$ & .886  & .360  & .579  & .491  & 2.55  & .857  & .393  & \textbf{.660} & .486  & 1.89  & \textbf{.875} & .289  & \textbf{.688} & \textbf{.397} & \textbf{1.97}  & .903  & .311  & \textbf{.746} & .476  & 2.46  & \textbf{.853} & .332  & \textbf{.579} & .416  & 1.85  & \textbf{.933} & .290  & .618  & .439  & 2.96  \\
      DeepVS~\cite{jiang2018deepvs}$\dagger$ & .896  & .391  & \textbf{.585} & .528  & 3.01  & .840  & .392  & .625  & .452  & 1.86  & .842  & .262  & .612  & .317  & 1.62  & .904  & \textbf{.349} & .686  & .461  & \textbf{2.48}  & .830  & .317  & .561  & .359  & 1.77  & .925  & .259  & \textbf{.646} & \textbf{.449} & \textbf{3.79}  \\
      ACLNet~\cite{wang2018revisiting}$\dagger$ & \textbf{.905} & \textbf{.446} & .560  & \textbf{.580} & \textbf{3.17}  & \textbf{.869} & \textbf{.427} & .622  & \textbf{.522} & \textbf{2.02}  & .868  & \textbf{.296} & .609  & .379  & 1.79  & \textbf{.915} & .329  & .675  & \textbf{.477} & 2.36  & .850  & \textbf{.361} & .542  & \textbf{.425} & \textbf{1.92}  & .926  & \textbf{.322} & .594  & .448  & 3.16  \\
      \specialrule{3.5pt}{0pt}{0pt}
      \multicolumn{31}{l}{Compared with other fully supervised methods, our method has adopted a different training protocol, \emph{i.e.}, SOTA fully supervised methods were trained on the widely-}\\
      \multicolumn{31}{l}{used training set (2,857 videos~\cite{wang2018revisiting, rodriguez2008action, marszalek2009actions}) with real human-eye fixations, yet our method was solely trained on the AVE set, a sub-set of the AudioSet~\cite{gemmeke2017audio}, contai-}\\
      \multicolumn{31}{l}{ning 4,143 clips equipped with semantic tags. \textbf{Notice}: */\#/$\dagger$ denotes Un-/Weakly-/Fully-supervised model.}\\
      \specialrule{3.5pt}{0pt}{0pt}
    \end{tabular}%
    \label{tab:Quantitative}%
      }\vspace{-0.5cm}
  \end{table*}%
{
We tested multiple variants whose GCN have been replaced by either RNNs or Transformers. As reported in Table~\ref{tab:VAGNN}, the one with the best performance among the Transformer-based variants places at the same performance level as our C-GCN-based version, and the RNNs-based version is clearly inferior to our C-GCN-based version.Since we have tested multiple Transformer-based variants, we detail their implementations/architectures in Fig.~\ref{fig:MultiHead}.
In our implementation, we have respectively tested both singlehead and multihead Transformers, each of which have two different realizations, \emph{i.e.}, \{\textbf{A},~\textbf{C}\} \emph{v.s.} \{\textbf{B},~\textbf{D}\}, where \textbf{C} is the MultiHead version of \textbf{A}, and \textbf{D} is the MultiHead version of \textbf{B}.
An ablation study for verrifiying the number of Encoder Block (Fig.~\ref{fig:MultiHead}-\textbf{E}) is introduced.
As reported in Table~\ref{tab:VAGNN}, the performance the singlehead Transformer-based variant with three encoder blocks is comparable to our C-GCN-based model with three step reasoning.
The Transformer might require additional learning parameters, \emph{i.e.}, almost 52 times model size and 5.6 times FLOPs than C-GCN.
}

\subsubsection{Fusion Schemes of the Final Fixation Prediction Nets}
\label{subsubsec:SCAM+}


As we have stated in Sec.~\ref{sec:PVFFina}, we have proposed a new fusion scheme to combine 3 fixation maps predicted by 3 different networks, \textit{i.e.}, the sourcewise ${\rm PF}_{sta}$, and the multi-granularity perception-based ${\rm PF}_{sta+}^{\rm Short}$ and ${\rm PF}_{sta+}^{\rm Long}$.
The rationale of the proposed fusion scheme is simply based on the complementary facts between these 3 fixation maps, thus, we verify the advantage of this fusion scheme towards conventional plain schemes from a quantitative perspective.

The comprehensive quantitative evaluation results can be found in Table~\ref{tab:sta+}.
Compared with the 3 basic fixation maps (\textit{i.e.}, the top-3 rows), the plain fusion schemes cannot take full advantage of their complementary attribute, and thus, the performance gains tend to shift between different sets and are truly marginal.

Specifically, since the proposed ${\rm PF}_{final}$ is based on performing an addition operation over its two parts (Eq.~\ref{eq:finalfp}), we have additionally tested the multiplicative-based alternative, denoted as ${\rm PF}_{agg}$, which can be formulated as:
\begin{equation}
\begin{aligned}
\label{eq:finalfp2}
{\rm PF}_{agg} =\ &\frac{1}{2}\times\mathcal{Z}\Big(\mathcal{Z}({\rm PF}_{sta})\odot \mathcal{Z}({\rm PF}_{sta+}^{\rm Short})\odot \mathcal{Z}({\rm PF}_{sta+}^{\rm Long})\Big)\\[-0.5ex]
&\times \frac{1}{2}\times\mathcal{Z}\Big({\rm PF}_{sta}+{\rm PF}_{sta+}^{\rm Short}+{\rm PF}_{sta+}^{\rm Long}\Big),
\end{aligned}
\end{equation}
where all symbols and operations are completely identical to those of Eq.~\ref{eq:finalfp}.

As seen from the Table~\ref{tab:sta+}, we find that the performance of ${\rm PF}_{agg}$ outperforms ${\rm PF}_{final}$ in several cases, the numeric gaps are marginal. In sharp contrast, ${\rm PF}_{final}$ can outperform ${\rm PF}_{agg}$ significantly in many cases, \emph{e.g.}, the CC metric in the SunMe dataset, $0.359\rightarrow 0.402$.

\vspace{-0.3cm}
\subsection{Quantitative Comparisons with SOTA Works}

We have compared our model, \textit{i.e.}, the STANet+, trained using pseudofixations only, with 27 other SOTA methods, including 5 unsupervised methods, 18 weakly supervised methods, and 4 fully supervised methods on 6 publicly available testing sets.

\vspace{0.2cm}
\noindent\textbf{Quantitative Comparisons with Unsupervised and Weakly Supervised Methods.}
As shown in Table~\ref{tab:Quantitative}, our method (STANet+) significantly outperforms all unsupervised methods, and it also outperforms the most recent weakly supervised competitors (\emph{e.g.}, MWS~\cite{zeng2019multi} and WSSA~\cite{zhang2020weakly}).
The most straightforward way to encourage the weakly supervised model to output small regions close to fixation is to use thresholding. The result can be seen in Table~\ref{tab:shink3}.
And the fact is that simply shrinking salient regions via naive thresholding is incapable of obtaining high-quality fixation maps. The main reason is that the saliency maps generated by weakly supervised methods are \textbf{\emph{theoretically different}} with the real fixations.

Different from the conventional video-based CAM approaches~\cite{chattopadhay2018grad,Li_2021_WACV} which tend to highlight the single salient object persistently at the object level, the frame regional level information highlighted by our approach --- the most discriminative ones, might vary from frame to frame in video, because, in the visual-audio circumstance, either spatial, temporal, or audio could alternatively contribute most to the classification task.
This attribute is very consistent with the real human attention mechanism because humans never pay attention to a fixed location for a long period of time in visual-audio circumstances.

\begin{table}[!t]\large
  \centering
  \caption{The influence of different ``Thresholds" on the saliency detection results of MWS~\cite{zeng2019multi}, WSSA~\cite{zhang2020weakly}, WSS~\cite{wang2017learning}, ScoCAM~\cite{wang2020score}, LCAM~\cite{jiang2021layercam}, and SGradCAMpp~\cite{omeiza2019smooth}, ``\textbf{0.0}" means no threshold is applied on the saliency result, and ``0.3" means that the saliency value less than 0.3 is set to 0.}
  \vspace{-0.3cm}
  \resizebox{1\linewidth}{!}{
    \begin{tabular}{c||ccc|ccc|ccc}
      \specialrule{1.5pt}{0pt}{0pt}
  Methods & \multicolumn{3}{c|}{WSSA~\cite{zhang2020weakly}}   & \multicolumn{3}{c|}{MWS~\cite{zeng2019multi}}  & \multicolumn{3}{c}{WSS~\cite{wang2017learning}}  \\
    \hline
    Metrics & AUC-J$\uparrow$ &   CC$\uparrow$    & NSS$\uparrow$  & AUC-J$\uparrow$ &   CC$\uparrow$    & NSS$\uparrow$  &  AUC-J$\uparrow$   & CC$\uparrow$    & NSS$\uparrow$ \\
    \hline
    \hline
      0.0   & \textbf{0.807} & \textbf{0.285} & 1.339  & \textbf{0.834} & \textbf{0.309} & \textbf{1.477} & \textbf{0.858} & \textbf{0.347} & \textbf{1.655} \\
      0.3   & 0.799  & 0.285  & 1.339  & 0.801  & 0.296  & 1.444  & 0.827  & 0.337  & 1.623  \\
      0.5   & 0.801  & 0.285  & 1.344  & 0.772  & 0.262  & 1.419  & 0.813  & 0.333  & 1.617  \\
      0.7   & 0.800  & 0.285  & \textbf{1.347}  & 0.735  & 0.254  & 1.279  & 0.788  & 0.320  & 1.574  \\
    \hline
  Methods & \multicolumn{3}{c}{ScoCAM~\cite{wang2020score}} & \multicolumn{3}{c|}{LCAM~\cite{jiang2021layercam}} & \multicolumn{3}{c}{SGradCAMpp~\cite{omeiza2019smooth}} \\
    \hline
    Metrics & AUC-J$\uparrow$ &   CC$\uparrow$    & NSS$\uparrow$  & AUC-J$\uparrow$ &   CC$\uparrow$    & NSS$\uparrow$  &  AUC-J$\uparrow$   & CC$\uparrow$    & NSS$\uparrow$ \\
    \hline
    \hline
    0.0   & \textbf{0.772} & \textbf{0.237} & \textbf{1.016} & \textbf{0.776} & \textbf{0.241} & \textbf{1.033} & \textbf{0.809} & \textbf{0.275} & \textbf{1.181} \\
    0.3   & 0.763  & 0.231  & 0.989  & 0.764  & 0.233  & 0.992   & 0.806  & 0.269  & 1.151  \\
    0.5   & 0.735  & 0.224  & 0.969  & 0.737  & 0.227  & 0.974   & 0.726  & 0.181  & 0.789  \\
    0.7   & 0.687  & 0.207  & 0.916  & 0.698  & 0.221  & 0.978   & 0.630  & 0.122  & 0.501  \\
    \specialrule{1.5pt}{0pt}{0pt}
  \end{tabular}%
  \label{tab:shink3}%
  }\vspace{-0.5cm}
\end{table}%

\vspace{0.2cm}
\noindent\textbf{Quantitative Comparisons with Fully Supervised Methods.}
As shown in Table~\ref{tab:Quantitative}, our method achieves comparable results to the fully supervised methods and even outperforms some of them, \emph{e.g.}, DeepNet~\cite{pan2016shallow} in all testing sets except Coutrot2. The main reason is that the semantic contents of the Coutrot2 set are quite different from those of the AVE set, while our model is weakly supervised by the category tags of the AVE set. Note that the performance of our approach can be boosted further by including more tagged audiovisual sequences.

For fair comparison, both our method and the other three representative fully supervised SOTA methods~\cite{wang2018revisiting,pan2017salgan,pan2016shallow} are trained on an identical training set, \emph{i.e.}, the widely-used 168 video clips (70\% of the total 241 sequences) obtained from the six visual-audio sets~\cite{min2016fixation,coutrot2014saliency,coutrot2016multimodal,mital2011clustering,gygli2014creating,koutras2015perceptually}, all these clips have been equipped with real fixations.
In the training of our approach, we have omitted the use of real fixations, and, instead, we have assigned each clip with an appropriate video tag.
The results reported in Table~\ref{tab:TrainData} suggest that our approach cannot perform well on such a small-scale set, and other SOTA methods also get degenerate.

As shown in Table~\ref{tab:Quantitative}, we can achieve some performance gain by using the VggSound dataset~\cite{chen2020vggsound} (STANet+$\ddagger$), but this gain is at a relatively large expense on data and computation. The main reason is that a large portion of videos in VggSound is entirely irrelevant to the adopted testing sets.

\noindent\textbf{Limitation}
In our implementation, we have only considered one single semantic tag for each visual-audio sequence, while, in practice, a video sequence could be assigned with multiple tags.
Thus, our method might not be able to perform well for videos with massive out-of-scope semantic contents.
This problem can be alleviated by including more data with multiple tags, calling for future research.
\begin{table}[!t]
  \centering
  \caption{
    {Qualitative comparison between our method and the other three representative fully supervised SOTA models. All methods were trained on an identical set (168 sequences), where the real fixations are used when training fully supervised SOTA, while, in our training, only the assigned video tags are used. The numeric results are obtained by testing these models on the AVAD set~\cite{min2016fixation}}.
  }
  \vspace{-0.3cm}
  \resizebox{0.9\linewidth}{!}{
    \begin{tabular}{r||ccccc}
    \specialrule{1.0pt}{0pt}{0pt}
     Methods  & AUC-J$\uparrow$ & SIM$\uparrow$   & S-AUC$\uparrow$ & CC$\uparrow$    & NSS$\uparrow$ \\
    \hline
    \hline
    STANet+   & 0.887  & 0.361 & 0.595  & 0.490 & 2.318  \\
    STANet+~(trained on 168)   & 0.856  & 0.264 & 0.609  & 0.378 & 1.711  \\
    ACLNet~\cite{wang2018revisiting}   & 0.905  & 0.446  & 0.560  & 0.580  & 3.170 \\
    ACLNet~(re-trained on 168)   & 0.838  & 0.329  & 0.579  & 0.393  & 1.843  \\
    SalGAN~\cite{pan2017salgan}  & 0.886  & 0.360  &  0.579   & 0.491   &  2.550 \\
    SalGAN~(re-trained on 168) & 0.845  & 0.327  & 0.556  & 0.382  & 1.854  \\
    DeepNet~\cite{pan2016shallow}  &  0.869  &  0.256  &  0.561  &  0.383   &  1.850 \\
    DeepNet~(re-trained on 168)  & 0.852  & 0.247  & 0.582  & 0.410  & 1.750  \\
    \specialrule{1.0pt}{0pt}{0pt}
    \end{tabular}%
  \label{tab:TrainData}%
  }\vspace{-0.4cm}
\end{table}%

\vspace{-0.2cm}
\section{Conclusion}
In this paper, we have detailed a novel scheme for converting video-audio semantic category tags to pseudofixations.
Compared with the widely used CAM mechanism, the proposed SCAM and SCAM+ are able to produce pseudofixations that are more consistent with the real human fixations.
The key technical innovations include multisource-based selective fusion and its multistage methodology.
In addition, multi-granularity information has also been considered, where a novel way to combine it with the sourcewise feature has been devised, eventually enabling multi-granularity perception.
With the help of multi-granularity perception, our approach could be more consistent with the real human attention mechanism to correctly locate the most discriminative spatial regions in multimodality circumstances.
Moreover, to enable a generic application in cases without any human-provided video tags, we devised multiple spatial-temporal-audio fixation prediction networks (STA and STA+), which are trained on pseudofixations produced by the proposed SCAM and SCAM+, enabling end-to-end fixation prediction.
We have also compared our model --- the STA and STA+ fixation prediction network trained using our pseudofixations, with other SOTA methods.
The results favor our new method over unsupervised and weakly supervised methods. Besides, they also show that our method is even better than some fully supervised methods.
\vspace{-0.4cm}

\bibliographystyle{IEEEtran}
\bibliography{papern}{}

\begin{thebibliography}{100}
\providecommand{\url}[1]{#1}
\csname url@samestyle\endcsname
\providecommand{\newblock}{\relax}
\providecommand{\bibinfo}[2]{#2}
\providecommand{\BIBentrySTDinterwordspacing}{\spaceskip=0pt\relax}
\providecommand{\BIBentryALTinterwordstretchfactor}{4}
\providecommand{\BIBentryALTinterwordspacing}{\spaceskip=\fontdimen2\font plus
\BIBentryALTinterwordstretchfactor\fontdimen3\font minus
  \fontdimen4\font\relax}
\providecommand{\BIBforeignlanguage}[2]{{%
\expandafter\ifx\csname l@#1\endcsname\relax
\typeout{** WARNING: IEEEtran.bst: No hyphenation pattern has been}%
\typeout{** loaded for the language `#1'. Using the pattern for}%
\typeout{** the default language instead.}%
\else
\language=\csname l@#1\endcsname
\fi
#2}}
\providecommand{\BIBdecl}{\relax}
\BIBdecl

\bibitem{Wang_2021_CVPR}
G.~Wang, C.~Chen, D.-P. Fan, A.~Hao, and H.~Qin, ``From semantic categories to
  fixations: A novel weakly-supervised visual-auditory saliency detection
  approach,'' in \emph{CVPR}, 2021.

\bibitem{wang2019revisiting}
W.~Wang, J.~Shen, J.~Xie, M.-M. Cheng, H.~Ling, and A.~Borji, ``Revisiting
  video saliency prediction in the deep learning era,'' \emph{TPAMI}, vol.~43,
  no.~1, pp. 220--237, 2021.

\bibitem{linardos2019simple}
P.~Linardos, E.~Mohedano, J.~Nieto, N.~OConnor, X.~GiroiNieto, and
  K.~McGuinness, ``Simple vs complex temporal recurrences for video saliency
  prediction,'' \emph{BMVC}, 2019.

\bibitem{lai2019video}
Q.~Lai, W.~Wang, H.~Sun, and J.~Shen, ``Video saliency prediction using
  spatiotemporal residual attentive networks,'' \emph{TIP}, vol.~29, pp.
  1113--1126, 2019.

\bibitem{fang2018deep3dsaliency}
Y.~Fang, G.~Ding, J.~Li, and Z.~Fang, ``Deep3dsaliency: Deep stereoscopic video
  saliency detection model by 3d convolutional networks,'' \emph{TIP}, vol.~28,
  no.~5, pp. 2305--2318, 2018.

\bibitem{newvideosal20aaai}
X.~Wu, Z.~Wu, J.~Zhang, L.~Ju, and S.~Wang, ``Salsac: A video saliency
  prediction model with shuffled attentions and correlation-based convlstm,''
  in \emph{AAAI}, 2020.

\bibitem{xu2018personalized}
Y.~Xu, S.~Gao, J.~Wu, N.~Li, and J.~Yu, ``Personalized saliency and its
  prediction,'' \emph{TPAMI}, vol.~41, no.~12, pp. 2975--2989, 2018.

\bibitem{gu2020pyramid}
Y.~Gu, L.~Wang, Z.~Wang, Y.~Liu, M.-M. Cheng, and S.~Lu, ``Pyramid constrained
  self-attention network for fast video salient object detection,'' in
  \emph{AAAI}, 2020.

\bibitem{wang2020learning}
B.~Wang, W.~Liu, G.~Han, and S.~He, ``Learning long-term structural
  dependencies for video salient object detection,'' \emph{TIP}, vol.~29, pp.
  9017--9031, 2020.

\bibitem{ren2020tenet}
S.~Ren, C.~Han, X.~Yang, G.~Han, and S.~He, ``Tenet: Triple excitation network
  for video salient object detection,'' in \emph{ECCV}, 2020.

\bibitem{MGA}
H.~Li, G.~Chen, G.~Li, and Y.~Yu, ``Motion guided attention for video salient
  object detection,'' in \emph{ICCV}, 2019.

\bibitem{SSAV}
D.-P. Fan, W.~Wang, M.-M. Cheng, and J.~Shen, ``Shifting more attention to
  video salient object detection,'' in \emph{CVPR}, 2019.

\bibitem{CC2017TIP}
C.~Chen, S.~Li, Y.~Wang, A.~Hao, and H.~Qin, ``Video saliency detection via
  spatial-temporal fusion and low-rank coherency diffusion,'' \emph{TIP},
  vol.~26, no.~7, pp. 3156--3170, 2017.

\bibitem{CC2020TIPWGT}
C.~Chen, G.~Wang, C.~Peng, X.~Zhang, and H.~Qin, ``Improved robust video
  saliency detection based on long-term spatial-temporal information,''
  \emph{TIP}, vol.~29, no.~1, pp. 1090--1100, 2020.

\bibitem{cong2019video}
R.~Cong, J.~Lei, H.~Fu, F.~Porikli, Q.~Huang, and C.~Hou, ``Video saliency
  detection via sparsity-based reconstruction and propagation,'' \emph{TIP},
  vol.~28, no.~10, pp. 4819--4831, 2019.

\bibitem{li2019supervae}
B.~Li, Z.~Sun, and Y.~Guo, ``Supervae: Superpixelwise variational autoencoder
  for salient object detection,'' in \emph{AAAI}, 2019.

\bibitem{min2019tased}
K.~Min and J.~Corso, ``Tased-net: Temporally-aggregating spatial
  encoder-decoder network for video saliency detection,'' in \emph{ICCV}, 2019.

\bibitem{wu2020salsac}
X.~Wu, Z.~Wu, J.~Zhang, L.~Ju, and S.~Wang, ``Salsac: A video saliency
  prediction model with shuffled attentions and correlation-based convlstm,''
  in \emph{AAAI}, 2020.

\bibitem{droste2020unified}
R.~Droste, J.~Jiao, and J.~Noble, ``Unified image and video saliency
  modeling,'' \emph{ECCV}, 2020.

\bibitem{bak2017spatio}
C.~Bak, A.~Kocak, E.~Erdem, and A.~Erdem, ``Spatio-temporal saliency networks
  for dynamic saliency prediction,'' \emph{TMM}, vol.~20, no.~7, pp.
  1688--1698, 2017.

\bibitem{sun2018sg}
M.~Sun, Z.~Zhou, Q.~Hu, Z.~Wang, and J.~Jiang, ``Sg-fcn: A motion and
  memory-based deep learning model for video saliency detection,'' \emph{TCYB},
  vol.~49, no.~8, pp. 2900--2911, 2018.

\bibitem{gorji2018going}
S.~Gorji and J.~Clark, ``Going from image to video saliency: Augmenting image
  salience with dynamic attentional push,'' in \emph{CVPR}, 2018.

\bibitem{jiang2015salicon}
M.~Jiang, S.~Huang, J.~Duan, and Q.~Zhao, ``Salicon: Saliency in context,'' in
  \emph{CVPR}, 2015.

\bibitem{pan2016shallow}
J.~Pan, E.~Sayrol, X.~GiroiNieto, K.~McGuinness, and N.~OConnor, ``Shallow and
  deep convolutional networks for saliency prediction,'' in \emph{CVPR}, 2016.

\bibitem{wang2017deep}
W.~Wang and J.~Shen, ``Deep visual attention prediction,'' \emph{TIP}, vol.~27,
  no.~5, pp. 2368--2378, 2017.

\bibitem{shewchenko2005heading}
ShewchenkoN, WithnallC, KeownM, GittensR, and DvorakJ, ``Heading in football.
  part 1: development of biomechanical methods to investigate head response,''
  \emph{BJSM}, vol.~39, no.~1, pp. 10--25, 2005.

\bibitem{shigeoka2013tumor}
M.~Shigeoka, N.~Urakawa, T.~Nakamura, M.~Nishio, T.~Watajima, D.~Kuroda,
  T.~Komori, Y.~Kakeji, S.~Semba, and H.~Yokozaki, ``Tumor associated
  macrophage expressing cd 204 is associated with tumor aggressiveness of
  esophageal squamous cell carcinoma,'' \emph{Cancer science}, vol. 104, no.~8,
  pp. 1112--1119, 2013.

\bibitem{mandal2016towards}
B.~Mandal, L.~Li, G.~S. Wang, and J.~Lin, ``Towards detection of bus driver
  fatigue based on robust visual analysis of eye state,'' \emph{TITS}, vol.~18,
  no.~3, pp. 545--557, 2016.

\bibitem{podlesny1977physiological}
J.~Podlesny and D.~Raskin, ``Physiological measures and the detection of
  deception.'' \emph{Psychological bulletin}, vol.~84, no.~4, p. 782, 1977.

\bibitem{kok2017before}
E.~Kok and H.~Jarodzka, ``Before your very eyes: The value and limitations of
  eye tracking in medical education,'' \emph{Medical education}, vol.~51,
  no.~1, pp. 114--122, 2017.

\bibitem{zhu2018prediction}
Y.~Zhu, G.~Zhai, and X.~Min, ``The prediction of head and eye movement for 360
  degree images,'' \emph{SP:IC}, vol.~69, pp. 15--25, 2018.

\bibitem{zhu2021viewing}
Y.~Zhu, G.~Zhai, Y.~Yang, H.~Duan, X.~Min, and X.~Yang, ``Viewing behavior
  supported visual saliency predictor for 360 degree videos,'' \emph{TCSVT},
  2021.

\bibitem{tsiami2020stavis}
A.~Tsiami, P.~Koutras, and P.~Maragos, ``Stavis: Spatio-temporal audiovisual
  saliency network,'' in \emph{CVPR}, 2020.

\bibitem{tavakoli2019dave}
H.~Tavakoli, A.~Borji, E.~Rahtu, and J.~Kannala, ``Dave: A deep audio-visual
  embedding for dynamic saliency prediction,'' \emph{arXiv preprint
  arXiv:1905.10693}, 2019.

\bibitem{tian2018audio}
Y.~Tian, J.~Shi, B.~Li, Z.~Duan, and C.~Xu, ``Audio-visual event localization
  in unconstrained videos,'' in \emph{ECCV}, 2018.

\bibitem{Zhou_2016_CVPR}
B.~Zhou, A.~Khosla, A.~Lapedriza, A.~Oliva, and A.~Torralba, ``Learning deep
  features for discriminative localization,'' in \emph{CVPR}, 2016.

\bibitem{zeng2019multi}
Y.~Zeng, Y.~Zhuge, H.~Lu, L.~Zhang, M.~Qian, and Y.~Yu, ``Multi-source weak
  supervision for saliency detection,'' in \emph{CVPR}, 2019.

\bibitem{wang2017learning}
L.~Wang, H.~Lu, Y.~Wang, M.~Feng, D.~Wang, B.~Yin, and X.~Ruan, ``Learning to
  detect salient objects with image-level supervision,'' in \emph{CVPR}, 2017.

\bibitem{saito2019strong}
K.~Saito, Y.~Ushiku, T.~Harada, and K.~Saenko, ``Strong-weak distribution
  alignment for adaptive object detection,'' in \emph{CVPR}, 2019.

\bibitem{bargal2018excitation}
S.~Bargal, A.~Zunino, D.~Kim, J.~Zhang, V.~Murino, and S.~Sclaroff,
  ``Excitation backprop for rnns,'' in \emph{CVPR}, 2018.

\bibitem{chattopadhay2018grad}
A.~Chattopadhay, A.~Sarkar, P.~Howlader, and V.~Balasubramanian, ``Grad-cam++:
  Generalized gradient-based visual explanations for deep convolutional
  networks,'' in \emph{WACV}, 2018.

\bibitem{Li_2021_WACV}
Z.~Li, W.~Wang, Z.~Li, Y.~Huang, and Y.~Sato, ``Towards visually explaining
  video understanding networks with perturbation,'' in \emph{WACV}, 2021.

\bibitem{Shipami}
Z.~Shi, Y.~Yang, T.~M. Hospedales, and T.~Xiang, ``Weakly-supervised image
  annotation and segmentation with objects and attributes,'' \emph{TPAMI},
  vol.~39, no.~12, pp. 2525--2538, 2017.

\bibitem{Sanginetopami}
E.~Sangineto, M.~Nabi, D.~Culibrk, and N.~Sebe, ``Self paced deep learning for
  weakly supervised object detection,'' \emph{TPAMI}, vol.~41, no.~3, pp.
  712--725, 2019.

\bibitem{lu2016learning}
Z.~Lu, Z.~Fu, T.~Xiang, P.~Han, L.~Wang, and X.~Gao, ``Learning from weak and
  noisy labels for semantic segmentation,'' \emph{TPAMI}, vol.~39, no.~3, pp.
  486--500, 2016.

\bibitem{qian2019weakly}
R.~Qian, Y.~Wei, H.~Shi, J.~Li, J.~Liu, and T.~Huang, ``Weakly supervised scene
  parsing with point-based distance metric learning,'' in \emph{AAAI}, 2019.

\bibitem{zhang2020weakly}
J.~Zhang, X.~Yu, A.~Li, P.~Song, B.~Liu, and Y.~Dai, ``Weakly-supervised
  salient object detection via scribble annotations,'' in \emph{CVPR}, 2020.

\bibitem{dai2015boxsup}
J.~Dai, K.~He, and J.~Sun, ``Boxsup: Exploiting bounding boxes to supervise
  convolutional networks for semantic segmentation,'' in \emph{ICCV}, 2015.

\bibitem{li2019progressive}
D.~Li, J.~Huang, Y.~Li, S.~Wang, and M.~Yang, ``Progressive representation
  adaptation for weakly supervised object localization,'' \emph{TPAMI},
  vol.~42, no.~6, pp. 1424--1438, 2019.

\bibitem{Wanpami}
F.~Wan, P.~Wei, Z.~Han, J.~Jiao, and Q.~Ye, ``Min-entropy latent model for
  weakly supervised object detection,'' \emph{TPAMI}, vol.~41, no.~10, pp.
  2395--2409, 2019.

\bibitem{zeng2019wsod2}
Z.~Zeng, B.~Liu, J.~Fu, H.~Chao, and L.~Zhang, ``Wsod2: Learning bottom-up and
  top-down objectness distillation for weakly-supervised object detection,'' in
  \emph{ICCV}, 2019.

\bibitem{jiang2021online}
P.~Jiang, L.~Han, Q.~Hou, M.-M. Cheng, and Y.~Wei, ``Online attention
  accumulation for weakly supervised semantic segmentation,'' \emph{TPAMI},
  2021.

\bibitem{choe2020attention}
J.~Choe, S.~Lee, and H.~Shim, ``Attention-based dropout layer for weakly
  supervised single object localization and semantic segmentation,''
  \emph{TPAMI}, vol.~43, no.~12, pp. 4256--4271, 2021.

\bibitem{li2019guided}
K.~Li, Z.~Wu, K.~Peng, J.~Ernst, and Y.~Fu, ``Guided attention inference
  network,'' \emph{TPAMI}, vol.~42, no.~12, pp. 2996--3010, 2019.

\bibitem{2020Reliability}
B.~Zhang, J.~Xiao, Y.~Wei, M.~Sun, and K.~Huang, ``Reliability does matter: An
  end-to-end weakly supervised semantic segmentation approach,'' in
  \emph{AAAI}, 2020.

\bibitem{Salehpami}
F.~S. Saleh, M.~S. Aliakbarian, M.~Salzmann, L.~Petersson, J.~Alvarez, and
  S.~Gould, ``Incorporating network built-in priors in weakly-supervised
  semantic segmentation,'' \emph{TPAMI}, vol.~40, no.~6, pp. 1382--1396, 2018.

\bibitem{CAMYangyi}
X.~Zhang, Y.~Wei, J.~Feng, Y.~Yang, and T.~Huang, ``Adversarial complementary
  learning for weakly supervised object localization,'' in \emph{CVPR}, 2018.

\bibitem{gao2019c}
Y.~Gao, B.~Liu, N.~Guo, X.~Ye, F.~Wan, H.~You, and D.~Fan, ``C-midn: Coupled
  multiple instance detection network with segmentation guidance for weakly
  supervised object detection,'' in \emph{ICCV}, 2019.

\bibitem{wan2019c}
F.~Wan, C.~Liu, W.~Ke, X.~Ji, J.~Jiao, and Q.~Ye, ``C-mil: Continuation
  multiple instance learning for weakly supervised object detection,'' in
  \emph{CVPR}, 2019.

\bibitem{arun2019dissimilarity}
A.~Arun, C.~Jawahar, and M.~Kumar, ``Dissimilarity coefficient based weakly
  supervised object detection,'' in \emph{CVPR}, 2019.

\bibitem{chen2020slv}
Z.~Chen, Z.~Fu, R.~Jiang, Y.~Chen, and X.~Hua, ``Slv: Spatial likelihood voting
  for weakly supervised object detection,'' in \emph{CVPR}, 2020.

\bibitem{ye2019cap2det}
K.~Ye, M.~Zhang, A.~Kovashka, W.~Li, D.~Qin, and J.~Berent, ``Cap2det: Learning
  to amplify weak caption supervision for object detection,'' in \emph{ICCV},
  2019.

\bibitem{yang2019activity}
Z.~Yang, D.~Mahajan, D.~Ghadiyaram, R.~Nevatia, and V.~Ramanathan, ``Activity
  driven weakly supervised object detection,'' in \emph{CVPR}, 2019.

\bibitem{inoue2018cross}
N.~Inoue, R.~Furuta, T.~Yamasaki, and K.~Aizawa, ``Cross-domain
  weakly-supervised object detection through progressive domain adaptation,''
  in \emph{CVPR}, 2018.

\bibitem{lu2018revealing}
Y.~Lu, J.~Yin, Z.~Chen, H.~Gong, Y.~Liu, L.~Qian, X.~Li, R.~Liu, I.~Andolina,
  and W.~Wang, ``Revealing detail along the visual hierarchy: neural clustering
  preserves acuity from v1 to v4,'' \emph{Neuron}, vol.~98, no.~2, pp.
  417--428, 2018.

\bibitem{chen2020vggsound}
H.~Chen, W.~Xie, A.~Vedaldi, and A.~Zisserman, ``Vggsound: A large-scale
  audio-visual dataset,'' in \emph{ICASSP}, 2020.

\bibitem{Karen_2015_iclr}
K.~Simonyan and A.~Zisserman, ``Very deep convolutional networks for
  large-scale image recognition,'' in \emph{ICLR}, 2015.

\bibitem{wang2019zero}
W.~Wang, X.~Lu, J.~Shen, D.~J. Crandall, and L.~Shao, ``Zero-shot video object
  segmentation via attentive graph neural networks,'' in \emph{ICCV}, 2019.

\bibitem{li2020group}
X.~Li, T.~Zhou, J.~Li, Y.~Zhou, and Z.~Zhang, ``Group-wise semantic mining for
  weakly supervised semantic segmentation,'' \emph{AAAI}, 2021.

\bibitem{sun2020mining}
G.~Sun, W.~Wang, J.~Dai, and L.~VanGool, ``Mining cross-image semantics for
  weakly supervised semantic segmentation,'' in \emph{ECCV}, 2020.

\bibitem{Xia2021evalu}
C.~Xia, J.~Han, and D.~Zhang, ``Evaluation of saccadic scanpath prediction:
  Subjective assessment database and recurrent neural network based metric,''
  \emph{TPAMI}, vol.~43, no.~12, pp. 4378--4395, 2021.

\bibitem{yang2019visual}
T.~Yang and A.~Chan, ``Visual tracking via dynamic memory networks,''
  \emph{TPAMI}, vol.~43, no.~1, pp. 360--374, 2019.

\bibitem{sun2019visual}
W.~Sun, Z.~Chen, and F.~Wu, ``Visual scanpath prediction using ior-roi
  recurrent mixture density network,'' \emph{TPAMI}, vol.~43, no.~6, pp.
  2101--2118, 2019.

\bibitem{zhang2020neur}
B.~Zhang, D.~Xiong, J.~Xie, and J.~Su, ``Neural machine translation with
  gru-gated attention model,'' \emph{TNNLS}, vol.~31, no.~11, pp. 4688--4698,
  2020.

\bibitem{Kipf_2017_iclr}
T.~Kipf and M.~Welling, ``Semi-supervised classification with graph
  convolutional networks,'' in \emph{ICLR}, 2017.

\bibitem{pan2017salgan}
J.~Pan, C.~Ferrer, K.~McGuinness, N.~OConnor, J.~Torres, E.~Sayrol, and
  X.~GiroiNieto, ``Salgan: Visual saliency prediction with generative
  adversarial networks,'' \emph{CVPR SUNw}, 2017.

\bibitem{borji2019saliency}
A.~Borji, ``Saliency prediction in the deep learning era: Successes and
  limitations,'' \emph{TPAMI}, 2019.

\bibitem{min2016fixation}
X.~Min, G.~Zhai, K.~Gu, and X.~Yang, ``Fixation prediction through multimodal
  analysis,'' \emph{ACM TOMM}, 2016.

\bibitem{gemmeke2017audio}
J.~Gemmeke, D.~Ellis, D.~Freedman, A.~Jansen, W.~Lawrence, R.~Moore, M.~Plakal,
  and M.~Ritter, ``Audio set: An ontology and human-labeled dataset for audio
  events,'' in \emph{ICASSP}, 2017.

\bibitem{bottou2008advan}
L.~Bottou and O.~Bousquet, ``The tradeoffs of large scale learning,'' in
  \emph{NeurIPS}, 2008.

\bibitem{coutrot2014saliency}
A.~Coutrot and N.~Guyader, ``How saliency, faces, and sound influence gaze in
  dynamic social scenes,'' \emph{JoV}, vol.~14, no.~8, pp. 5--5, 2014.

\bibitem{coutrot2016multimodal}
{A. Coutrot and N. Guyader}, ``Multimodal saliency models for videos,'' in
  \emph{From Human Attention to Computational Attention}, 2016, pp. 291--304.

\bibitem{mital2011clustering}
P.~Mital, T.~Smith, R.~Hill, and J.~Henderson, ``Clustering of gaze during
  dynamic scene viewing is predicted by motion,'' \emph{COGN COMPUT}, vol.~3,
  no.~1, pp. 5--24, 2011.

\bibitem{gygli2014creating}
M.~Gygli, H.~Grabner, H.~Riemenschneider, and L.~VanGool, ``Creating summaries
  from user videos,'' in \emph{ECCV}, 2014.

\bibitem{koutras2015perceptually}
P.~Koutras and P.~Maragos, ``A perceptually based spatio-temporal computational
  framework for visual saliency estimation,'' \emph{SP:IC}, vol.~38, pp.
  15--31, 2015.

\bibitem{wang2018revisiting}
W.~Wang, J.~Shen, F.~Guo, M.-M. Cheng, and A.~Borji, ``Revisiting video
  saliency: A large-scale benchmark and a new model,'' in \emph{CVPR}, 2018.

\bibitem{borji2012state}
A.~Borji and L.~Itti, ``State-of-the-art in visual attention modeling,''
  \emph{TPAMI}, vol.~35, no.~1, pp. 185--207, 2012.

\bibitem{bylinskii2018different}
Z.~Bylinskii, T.~Judd, A.~Oliva, A.~Torralba, and F.~Durand, ``What do
  different evaluation metrics tell us about saliency models?'' \emph{TPAMI},
  vol.~41, no.~3, pp. 740--757, 2018.

\bibitem{itti1998model}
L.~Itti, C.~Koch, and E.~Niebur, ``A model of saliency-based visual attention
  for rapid scene analysis,'' \emph{TPAMI}, vol.~20, no.~11, pp. 1254--1259,
  1998.

\bibitem{harel2007graph}
J.~Harel, C.~Koch, and P.~Perona, ``Graph-based visual saliency,'' in
  \emph{NeurIPS}, 2007.

\bibitem{rudoy2013learning}
D.~Rudoy, D.~Goldman, E.~Shechtman, and L.~ZelnikManor, ``Learning video
  saliency from human gaze using candidate selection,'' in \emph{CVPR}, 2013.

\bibitem{zhang2017supervision}
D.~Zhang, J.~Han, and Y.~Zhang, ``Supervision by fusion: Towards unsupervised
  learning of deep salient object detector,'' in \emph{ICCV}, 2017.

\bibitem{leboran2016dynamic}
V.~Leboran, A.~GarciaDiaz, X.~FdezVidal, and X.~Pardo, ``Dynamic whitening
  saliency,'' \emph{TPAMI}, vol.~39, no.~5, pp. 893--907, 2016.

\bibitem{zhou2016learning}
B.~Zhou, A.~Khosla, A.~Lapedriza, A.~Oliva, and A.~Torralba, ``Learning deep
  features for discriminative localization,'' in \emph{CVPR}, 2016.

\bibitem{selvaraju2017grad}
R.~Selvaraju, M.~Cogswell, A.~Das, R.~Vedantam, D.~Parikh, and D.~Batra,
  ``Grad-cam: Visual explanations from deep networks via gradient-based
  localization,'' in \emph{ICCV}, 2017.

\bibitem{omeiza2019smooth}
D.~Omeiza, S.~Speakman, C.~Cintas, and K.~Weldermariam, ``Smooth grad-cam++: An
  enhanced inference level visualization technique for deep convolutional
  neural network models,'' \emph{arXiv preprint arXiv:1908.01224}, 2019.

\bibitem{fu2020axiom}
R.~Fu, Q.~Hu, X.~Dong, Y.~Guo, Y.~Gao, and B.~Li, ``Axiom-based grad-cam:
  Towards accurate visualization and explanation of cnns,'' \emph{BMVC}, 2020.

\bibitem{wang2020ss}
H.~Wang, R.~Naidu, J.~Michael, and S.~Kundu, ``Ss-cam: Smoothed score-cam for
  sharper visual feature localization,'' \emph{arXiv preprint
  arXiv:2006.14255}, 2020.

\bibitem{wang2020score}
H.~Wang, Z.~Wang, M.~Du, F.~Yang, Z.~Zhang, S.~Ding, P.~Mardziel, and X.~Hu,
  ``Score-cam: Score-weighted visual explanations for convolutional neural
  networks,'' in \emph{CVPRW}, 2020.

\bibitem{jiang2021layercam}
P.~Jiang, C.~Zhang, Q.~Hou, M.-M. Cheng, and Y.~Wei, ``Layercam: Exploring
  hierarchical class activation maps for localization,'' \emph{TIP}, vol.~30,
  pp. 5875--5888, 2021.

\bibitem{naidu2020cam}
R.~Naidu, A.~Ghosh, Y.~Maurya, S.~Kundu \emph{et~al.}, ``Is-cam: Integrated
  score-cam for axiomatic-based explanations,'' \emph{arXiv preprint
  arXiv:2010.03023}, 2020.

\bibitem{ramaswamy2020ablation}
H.~Ramaswamy \emph{et~al.}, ``Ablation-cam: Visual explanations for deep
  convolutional network via gradient-free localization,'' in \emph{WACV}, 2020.

\bibitem{muhammad2020eigen}
M.~Muhammad and M.~Yeasin, ``Eigen-cam: Class activation map using principal
  components,'' in \emph{IJCNN}, 2020.

\bibitem{zhang2018self}
X.~Zhang, Y.~Wei, G.~Kang, Y.~Yang, and T.~Huang, ``Self-produced guidance for
  weakly-supervised object localization,'' in \emph{ECCV}, 2018.

\bibitem{jiang2018deepvs}
L.~Jiang, M.~Xu, T.~Liu, M.~Qiao, and Z.~Wang, ``Deepvs: A deep learning based
  video saliency prediction approach,'' in \emph{ECCV}, 2018.

\bibitem{rodriguez2008action}
M.~Rodriguez, J.~Ahmed, and M.~Shah, ``Action mach a spatio-temporal maximum
  average correlation height filter for action recognition,'' in \emph{CVPR},
  2008.

\bibitem{marszalek2009actions}
M.~Marszalek, I.~Laptev, and C.~Schmid, ``Actions in context,'' in \emph{CVPR},
  2009.

\end{thebibliography}

%
%
%
%

\begin{figure}[!b]
  \centering
  \vspace{-0.8cm}

  \hspace{0.4cm}\includegraphics[width=1\linewidth]{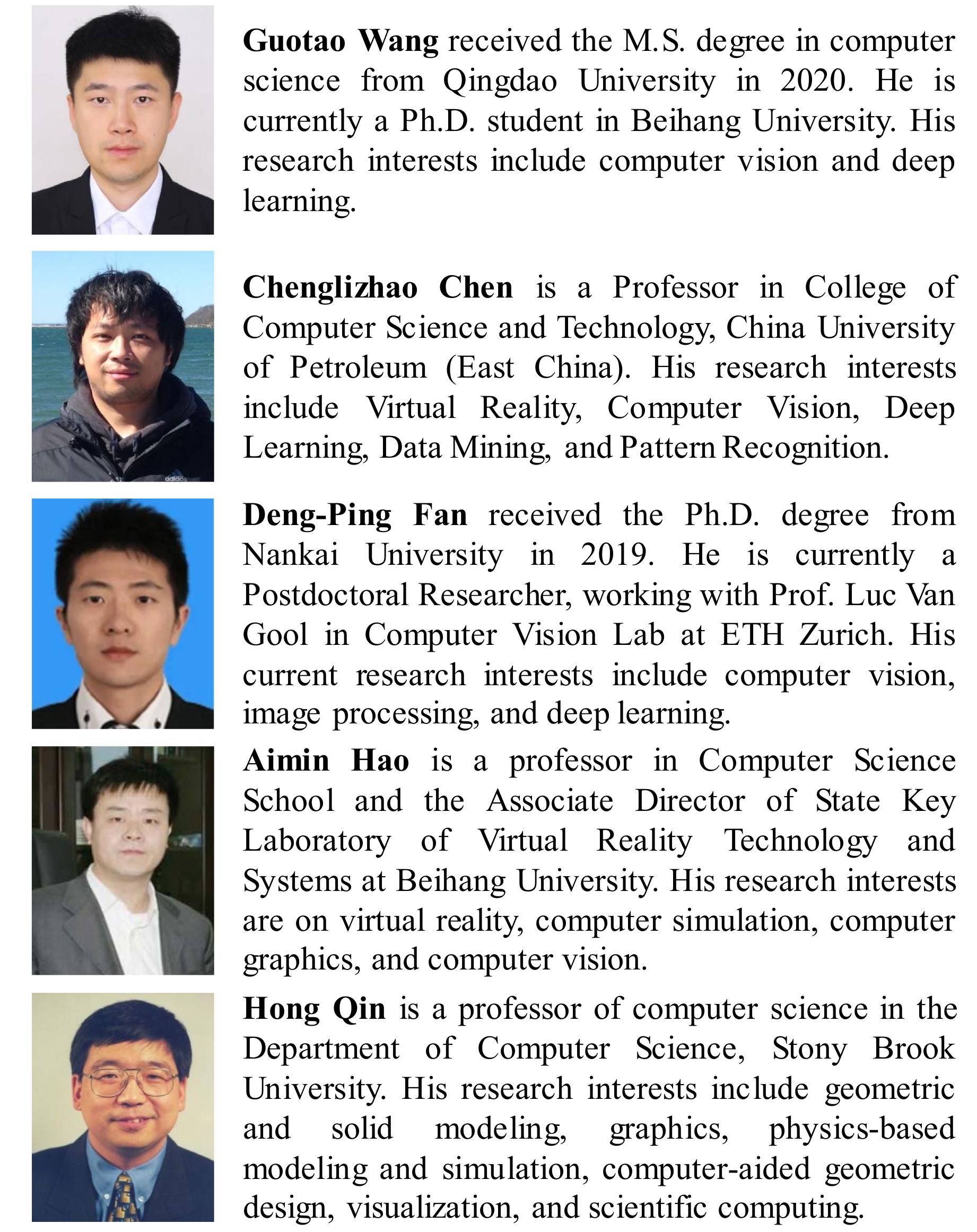}
  \vspace{-0.6cm}
\end{figure}

\end{document}